\RequirePackage[T1]{fontenc}
\documentclass[journal]{IEEEtran}
\usepackage{amsmath,amssymb,graphicx,cite,booktabs,array,multirow}
\usepackage{newtxtext,newtxmath}
\usepackage{algorithm,algorithmic,tikz,url,placeins}
\usetikzlibrary{arrows.meta,positioning}
\usepackage[hidelinks]{hyperref}
\usepackage{pifont}
\usepackage[caption=false,font=footnotesize]{subfig}

\newcommand{\pending}{\ensuremath{\ast}}

\newcommand{\sg}{\operatorname{stopgrad}}
\newcolumntype{P}[1]{>{\raggedright\arraybackslash}p{#1}}

\newcommand{\reconpanel}[3]{\begin{tikzpicture}
		\node[inner sep=0,anchor=south west] (im) at (0,0) {\includegraphics[width=#2,height=#3,trim=0bp 0bp 275bp 0bp,clip]{#1}};
		\begin{scope}[x={(im.south east)},y={(im.north west)}]
			\fill[white] (.965,0) rectangle (1,.916);
\end{scope}\end{tikzpicture}}
\newcommand{\gatherpanel}[2]{\ifnum#2=0\includegraphics[width=.139\textwidth,height=1.17in]{#1}\else\reconpanel{#1}{.139\textwidth}{1.17in}\fi}
\begin{document}
\title{MetaPOCS: Meta-Learned Multi-Exit Unfolding for Budget-Aware Seismic Data Interpolation}
\author{Baisong~Jiang, Chunxia~Zhang, Xiaoli Wei, Anxiang Di, Nanqi~Zhao, Deng Xiong
\thanks{Corresponding author: Chunxia Zhang. E-mail: cxzhang@mail.xjtu.edu.cn.}
\thanks{Baisong Jiang, Chunxia Zhang, Anxiang Di, Nanqi~Zhao are with the School of Mathematics and Statistics, Xi'an Jiaotong University, Xi'an, Shaanxi, 710049, China.}
\thanks{Xiaoli Wei is with the Business School, Beijing Wuzi University, Beijing, 101149, China.}
\thanks{Deng Xiong is with the Geophysical Technology Research and Development Center, BGP, Zhuozhou, Hebei, 072751, China.}
\thanks{The research is supported by the National Natural Science Foundation of China under grant 12371512; and in part by the Key Acquisition, Processing and Interpretation Techniques for Seismic Data Processing of China National Petroleum Corporation (CNPC) under Grant 01-03-2023.}
}
\markboth{IEEE Transactions on Geoscience and Remote Sensing, Working Manuscript}{Jiang \MakeLowercase{\textit{et al.}}: MetaPOCS}
\maketitle
\begin{abstract}
Seismic data interpolation must reconcile reconstruction fidelity with the computational budgets of acquisition quality control and subsequent imaging. Fixed-depth unfolding networks are optimized for a single operating point, whereas their intermediate estimates may be insufficiently trained for early deployment. We present MetaPOCS, a meta-learned multi-exit unfolding framework for budget-aware seismic interpolation. Building on a spatial--spectral POCS reconstruction backbone, MetaPOCS supervises intermediate stages and preserves the acquired traces at every exit. A Meta-GF-based optimization procedure constructs exit-oriented parameter displacements from a common initialization and learns their fusion weights through an equal-exit outer objective. The weights vary across parameter tensors and respect the nested support of the reconstruction stages. We derive the resulting fusion hypergradient to characterize how the outer objective reallocates competing updates, and distinguish this mechanism from scalar loss reweighting. At inference, one checkpoint provides a set of reconstruction budgets through a latency-calibrated exit policy. The evaluation framework couples matched-cost interpolation comparisons with finite-update diagnostics and a fixed reverse-time-migration operator, connecting multi-exit optimization to waveform and imaging fidelity. MetaPOCS thus provides a unified formulation for studying the reconstruction--computation trade-off of seismic unfolding across deployment budgets.
\end{abstract}

\section{Introduction}\label{sec:intro}
\IEEEPARstart{A}{cquiring} complete seismic data supports subsurface imaging and reservoir characterization. Terrain restrictions, receiver failures, and acquisition costs can leave irregularly sampled or missing traces. Interpolation seeks to recover the unobserved wavefield while preserving the information in acquired data. Its practical value depends on event kinematics and amplitude fidelity, as well as on computational resources available before subsequent processing.

Existing approaches can be organized into three classes: physical or mathematical prior-based model-driven methods, constrained data-driven methods, and hybrid methods integrating data and models. This classification separates assumptions about the wavefield from the mechanism used to learn or impose those assumptions.

Physical prior-based methods exploit propagation laws or coherent wavefield structure. Ronen~\cite{1987Ronen} formulated wave-equation trace interpolation, and Canning and Gardner~\cite{1996Canning} addressed regularly sampled data using wave-equation principles. Migration-based formulations incorporate propagation geometry into reconstruction~\cite{2016Verma,2018CHEN}; their usefulness depends on the propagation model and acquisition description. Prediction filtering provides another established route. Spitz~\cite{1991spitz} exploited frequency--space predictability, while nonstationary prediction filters~\cite{2021Chen} and streaming filters with varying smoothness~\cite{Liu2022Streaming} relaxed stationarity or computational requirements. Dynamic matching uses local event alignment to guide interpolation~\cite{2024xu}. These methods encode meaningful seismic structure, but their assumptions become harder to satisfy across wide gaps and rapidly changing events.

Sparse-transform methods formulate reconstruction in a domain where complete data admit a compact representation~\cite{2008candes}. Fourier sparse inversion and antileakage Fourier interpolation address incomplete spatial sampling~\cite{2007zwartjes,2008Xu,xu2010}. Curvelet and seislet representations capture multiscale directional or slope-aligned structure~\cite{2006candes,2010fomel}, and POCS-type iterations combine transform-domain regularization with reinsertion of acquired traces~\cite{2017zhang}. Local Radon and apex-shifted hyperbolic Radon representations supply additional seismic dictionaries~\cite{2010wang,2018gong}. Sampling operators for off-grid data extend this framework beyond simple trace deletion~\cite{2022yu}. Transform choice, geometry, and regularization jointly determine which missing events can be recovered.

Constrained data-driven methods learn nonlinear representations from examples. Residual learning and texture-constrained reconstruction have been explored for seismic interpolation~\cite{2019Wang,2021Fang}, as have conditional adversarial models~\cite{2021Kaur}. Spatial--Fourier losses and wavelet representations provide alternatives to purely pointwise supervision~\cite{2022Park,Liu2022Wavelet}. Transformer and feature-fusion models introduce broader contextual interactions~\cite{2024Gao,2025Mu}. Score-based and diffusion formulations address stochastic reconstruction~\cite{2024Meng,2025Li}, while learned off-grid regularization explicitly incorporates the sampling operator~\cite{2025modeep}. Generalization to unseen acquisition patterns and preservation of measured amplitudes nevertheless require explicit evaluation.

Hybrid deep unfolding networks retain an iterative reconstruction structure while learning parts of the prior. CNN-assisted POCS~\cite{2020Zhang}, POCS-Net~\cite{2023Chen}, and SP-Net~\cite{2024Wu} illustrate this direction. Our related SSI-POCS work~\cite{Jiang2026SSIPOCS} develops a fixed-stage spatial--spectral prior for large-gap interpolation by coupling Restormer and fast Fourier convolution branches through a dual-domain fusion module (DFM). The present work adopts that reconstruction backbone and addresses a complementary problem: optimizing its intermediate reconstructions for deployment at different computational budgets. The spatial--spectral prior, data reinsertion, analysis--synthesis regularization, and frequency-domain supervision build on SSI-POCS. Here, each exit receives explicit time--space amplitude and frequency-domain supervision, aligning the reconstruction objective with the implicit dual-domain representation. The focus of the new formulation is the coordination and use of nested reconstruction exits.

The question considered here is how to train that reconstruction process for several deployment budgets within a single model. A separate fixed-depth model can be trained for each budget, but this increases training and checkpoint storage. Conversely, a network optimized only at its deepest stage need not provide a reliable shallow reconstruction. Deep supervision~\cite{Lee2015DSN} and multi-exit networks~\cite{Huang2018MSDNet} suggest supervising intermediate outputs. However, exits share early stages and can prefer different parameter updates. GradNorm~\cite{Chen2018GradNorm} addresses unequal training rates, while gradient surgery~\cite{Yu2020PCGrad} addresses conflicting directions. Meta-reweighting~\cite{Ren2018Reweight} learns weighting rules through an outer objective. Most directly, Meta-GF~\cite{Sun2022MetaGF} learns how to fuse exit-specific updates in dynamic-depth networks. Meta-learning here concerns the training of reconstruction exits, not few-shot adaptation to a new survey.

For seismic unfolding, exit coordination is coupled to the acquisition operator. Each exit estimates the same amplitude field under the same measured-data constraint, but a shallow exit must complete that estimate with fewer refinement stages. Early-stage parameters consequently serve two roles: producing an immediately usable wavefield and preparing features for later iterations. Their preferred updates can differ even though the reconstruction target is shared. Moreover, the nested computation graph gives different exits access to different parameter tensors. These properties motivate an explicit treatment of parameter support and update interference, while the phase sensitivity of migration motivates evaluation beyond aggregate reconstruction error.

MetaPOCS combines supervised unfolding exits with parameter-tensor-wise fusion of exit-oriented training displacements. Following Meta-GF~\cite{Sun2022MetaGF}, the fusion weights are learned from an outer reconstruction objective. The formulation specializes this update to acquisition-consistent seismic outputs and makes the support of each exit explicit. An equal-exit outer objective retains all reconstruction budgets in the optimization target; inference then selects an available exit according to a calibrated latency budget. This separates the problem of training reliable intermediate wavefields from the choice of a deployment operating point.

The contributions are summarized as follows:
\begin{enumerate}
\item We formulate seismic interpolation as a family of supervised, acquisition-consistent unfolding exits. A single checkpoint supports several deployment budgets through a latency-calibrated prefix selection rule.
\item We adapt Meta-GF to the nested parameter support of the unfolding process and derive its tensor-wise fusion hypergradient. A scalar meta-supervision formulation and a finite-update analysis distinguish loss reweighting from the redistribution of exit-oriented optimization trajectories.
\item We establish an evaluation framework linking budget-dependent reconstruction, optimization interference, and migration-domain error. Matched-backbone and matched-computation controls separate training effects from the spatial--spectral prior inherited from SSI-POCS.
\end{enumerate}
Section~\ref{sec:method} describes the model and training. Section~\ref{sec:protocol} specifies the evaluation protocol. Sections~\ref{sec:results} and~\ref{sec:mechanism} organize reconstruction and mechanism experiments. Section~\ref{sec:rtm} examines downstream imaging, followed by limitations and conclusions.

\section{Budget-Aware Multi-Exit Unfolding}\label{sec:method}
\subsection{Problem Formulation and Classical POCS Reconstruction}
In field seismic exploration, adverse acquisition conditions, economic constraints, and environmental limitations often lead to incomplete spatial sampling. Let $x \in \mathbb{R}^{H \times W}$ denote the latent complete seismic gather (where $H$ and $W$ represent the temporal sampling points and spatial receiver traces, respectively), and let $M \in \{0, 1\}^{H \times W}$ represent the binary sampling mask, where $1$ designates an acquired trace and $0$ indicates an unrecorded position. The forward observation model for noise-free seismic data interpolation can be formulated as:
\begin{equation}
	y = M \odot x,
	\label{eq:obs}
\end{equation}
where $y \in \mathbb{R}^{H \times W}$ represents the observed incomplete seismic record, and $\odot$ denotes the element-wise Hadamard product. Recovering $x$ from $y$ is intrinsically an ill-posed underdetermined inverse problem. To obtain a physically plausible solution, compressive sensing (CS) frameworks exploit the underlying transform-domain sparsity of complete wavefields, casting the reconstruction as a sparsity-constrained regularized optimization problem:
\begin{equation}
	\arg\min_{x} \frac{1}{2} \| y - M \odot x \|_2^2 + \lambda \|\Psi x\|_1,
	\label{eq:cs_opt}
\end{equation}
where the first term represents the data-fidelity constraint in the time-space ($t\text{-}x$) domain, the second term enforces sparsity regularization under a specific transform operator $\Psi$, and $\lambda > 0$ governs the trade-off between measurement fidelity and transform sparsity.

Under the framework of proximal gradient descent, such as the Iterative Shrinkage-Thresholding Algorithm (ISTA), the iterative solution for problem~\eqref{eq:cs_opt} at iteration $k$ is given by:
\begin{align}
	r^k &= x^{k-1} - \tau M^T (M \odot x^{k-1} - y), \label{eq:ista_gd}\\
	x^k &= \operatorname{prox}_{\lambda \tau, \|\Psi \cdot\|_1}(r^k) = \Psi^{-1} \mathcal{S}_{\lambda \tau} (\Psi r^k), \label{eq:ista_prox}
\end{align}
where $\tau$ denotes the gradient descent step size, $\operatorname{prox}(\cdot)$ denotes the proximal operator, and $\mathcal{S}_\lambda(z) = \operatorname{sign}(z) \odot \max(|z| - \lambda, 0)$ is the element-wise soft-thresholding function. Because seismic trace selection satisfies $M^T M = M \odot M = M$, setting the step size to $\tau = 1$ reduces ISTA to the classical Projection Onto Convex Sets (POCS) algorithm. Specifically, the POCS iterative update simplifies to alternating between transform-domain thresholding and spatial data reinsertion:
\begin{align}
	\hat{x}^k &= \Psi^{-1} \mathcal{S}_{\lambda_k} (\Psi x^k), \label{eq:pocs_thresh}\\
	x^{k+1} &= y + (1 - M) \odot \hat{x}^k, \label{eq:pocs_reinsert}
\end{align}
where $x^0 = y$, and $\lambda_k$ denotes the iteration-dependent threshold parameter. 

Although classical POCS algorithms provide rigorous mathematical tractability and enforce exact data consistency on acquired traces, they exhibit notable operational drawbacks:
\begin{enumerate}
	\item \textbf{Rigid Prior Representations:} Predefined analytical transforms $\Psi$ (e.g., Fourier, Radon, or Curvelet transforms) are fundamentally linear and struggle to represent complex subsurface wavefields characterized by severe spatial aliasing, conflicting dips, and large contiguous trace gaps.
	\item \textbf{Heuristic Hyperparameter Tuning:} Manual scheduling of threshold parameters $\lambda_k$ often requires trial-and-error heuristics across different shot gathers, yielding sub-optimal convergence trajectories.
	\item \textbf{High Iteration Latency:} Traditional POCS workflows demand dozens or even hundreds of iterations to achieve acceptable wavefield restoration, severely restricting computational throughput.
\end{enumerate}

To circumvent these limitations, deep unfolding networks (DUNs) map iterative stages into trainable cascade architectures, replacing handcrafted transforms with parameterized neural networks~\cite{2024Wu, Jiang2026SSIPOCS}. However, existing DUNs are typically optimized as static, fixed-depth models~\cite{Jiang2026SSIPOCS}. When deployed under strict computational budgets requiring early truncated execution, they suffer a severe performance cliff because intermediate states act primarily as transitional features rather than reliable wavefield estimates~\cite{Jiang2026SSIPOCS}.

To resolve this conflict, we propose MetaPOCS, a meta-learned multi-exit unfolding framework featuring: (1) Budget-Aware Multi-Exit Backbone: Intermediate stages act as independently supervised exits (Exit $t \in \{1, \dots, T\}$) ensuring strict data consistency; (2) Support-Aware Meta Gradient Fusion: Tensor-wise meta-learning harmonizes conflicting parameter trajectories across shared early layers; (3) Calibrated Dynamic Deployment: A single checkpoint enables budget-driven prefix execution along a quality-latency Pareto frontier.

\subsection{Implicit Dual-Domain Prior Within Each Unrolled Stage}
Classical POCS relies on a fixed, handcrafted transform $\Psi$ in which complete seismic data are assumed to be sparse. In our deep unfolding framework, the rigid analysis and synthesis transforms are replaced by parameterized, learnable neural networks. Let $F_k$ and $G_k$ denote the forward and inverse transform networks at the $k$-th iteration stage, respectively.

\begin{figure}[htp]
	\centering
	\hspace{-2mm}
	\includegraphics[width=1\columnwidth]{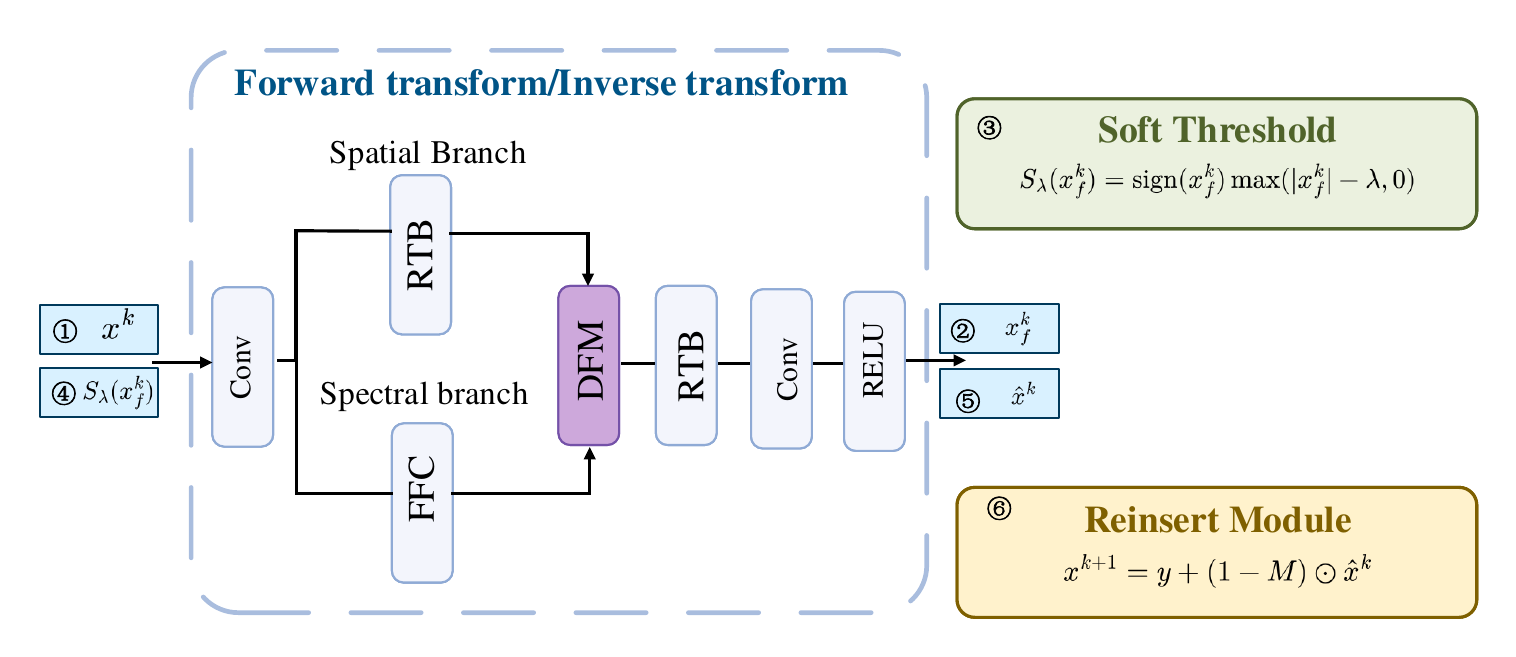}
	\caption{Detailed architecture of a single unrolled stage.}
	\label{fig:models}
\end{figure}

As illustrated in Fig.~\ref{fig:models}, the internal data flow of a single unrolled stage strictly mimics the mathematical logic of the classical POCS algorithm. First, the input wavefield estimate $x^k$ is mapped into a sparse feature representation via the forward transform: $x_f^k = F_k(x^k)$. Subsequently, a soft-thresholding operator enforces sparsity on this learned representation:
\begin{equation}
	S_\lambda(x_f^k) = \operatorname{sign}(x_f^k) \odot \max(|x_f^k| - \lambda, 0),
	\label{eq:soft_threshold}
\end{equation}
where $\lambda \geq 0$ is a learnable stage-specific threshold parameter. The sparsified features are then mapped back to the seismic-amplitude domain via the inverse transform to generate the intermediate reconstruction: $\hat{x}^k = G_k(S_\lambda(x_f^k))$. Finally, the reinsert module explicitly imposes the hard data-fidelity constraint by replacing the reconstructed traces at the acquired positions with the original observations $y$, yielding the updated state for the next iteration:
\begin{equation}
	x^{k+1} = y + (1 - M) \odot \hat{x}^k.
	\label{eq:reinsert}
\end{equation}

While the macroscopic unrolling follows the POCS logic, the core innovation lies in the internal architecture of $F_k$ and $G_k$. Unlike conventional single-domain unfolding methods that struggle to reconstruct large contiguous gaps due to restricted receptive fields, we design a heterogeneous dual-branch architecture. After an initial convolutional projection, the feature processing diverges into a Spatial Branch and a Spectral Branch. The Spatial Branch employs Restormer blocks (RTB) to capture local waveform structures and preserve the kinematic continuity of seismic events along the time-space lattice. Concurrently, the Spectral Branch utilizes Fast Fourier Convolution (FFC) to extract global contextual representations in the frequency domain, effectively bridging extensive missing regions that lack local structural information.

To fully exploit the complementarity between the local spatial details and the global spectral context, the features from both branches are aggregated through a Dual-domain Fusion Module (DFM). Instead of a naive feature concatenation, the DFM dynamically exchanges cross-domain context through bidirectional cross-channel attention and adaptive gated modulation, ensuring that the network suppresses incoherent artifacts while enhancing spectral energy concentration. Detailed operations of the internal cross-domain feature refinement and gated modulation mechanisms are referred to~\cite{Jiang2026SSIPOCS}. The modulated features are subsequently processed by a trailing RTB and convolution layer to produce the final branch output.

Ultimately, this structural design encodes an implicit dual-domain prior into the network weights. It avoids explicitly maintaining and alternately updating dual variables in two separate mathematical spaces; rather, the stage-wise composition of the learned transforms naturally regularizes the reconstruction by fusing multiscale spatial and spectral physical properties.

\subsection{Trained Exits and Budget-Aware Deployment}
Let $D=\{d_1,\ldots,d_T\}$ denote the set of trained exit depths, where $T$ is the total number of supervised exits. Exit $t$ (where $t \in \{1, \dots, T\}$) directly outputs the intermediate wavefield estimate $\hat{x}_t = x^{d_t}$ without requiring any auxiliary reconstruction head. While all $T$ stages are evaluated to compute the multi-exit losses during training, only the required prefix up to the selected depth is executed during deployment.

To accommodate real-world deployment constraints for a given hardware device, numerical precision, and input geometry, we define a maximum allowable latency budget $B$. Let $\tau_t$ represent the profiled inference latency and $q_t$ denote the validation-set quality score for Exit $t$. The optimal deployment exit $t^\star$ is dynamically selected by maximizing the reconstruction quality subject to the latency constraint:
\begin{equation}
	t^\star(B)=\arg\max_{t:\tau_t\le B}q_t. \label{eq:budget}
\end{equation}
The feasible set in Eq.~\eqref{eq:budget} comprises all exits whose profiled inference latency satisfies the budget $B$; an empty set indicates an unsupported budget constraint. This validation-driven selection policy is crucial because reconstruction accuracy in deep unfolding networks does not strictly increase monotonically with depth. By establishing this budget-to-exit mapping offline prior to actual testing, only the selected prefix is executed online. Consequently, the computational cost is strictly limited to the executed layers, providing a guaranteed, budget-aware seismic inference framework tailored for latency-sensitive applications.

\begin{figure}[t]
\centering
\hspace{-2mm}
\includegraphics[width=1\linewidth]{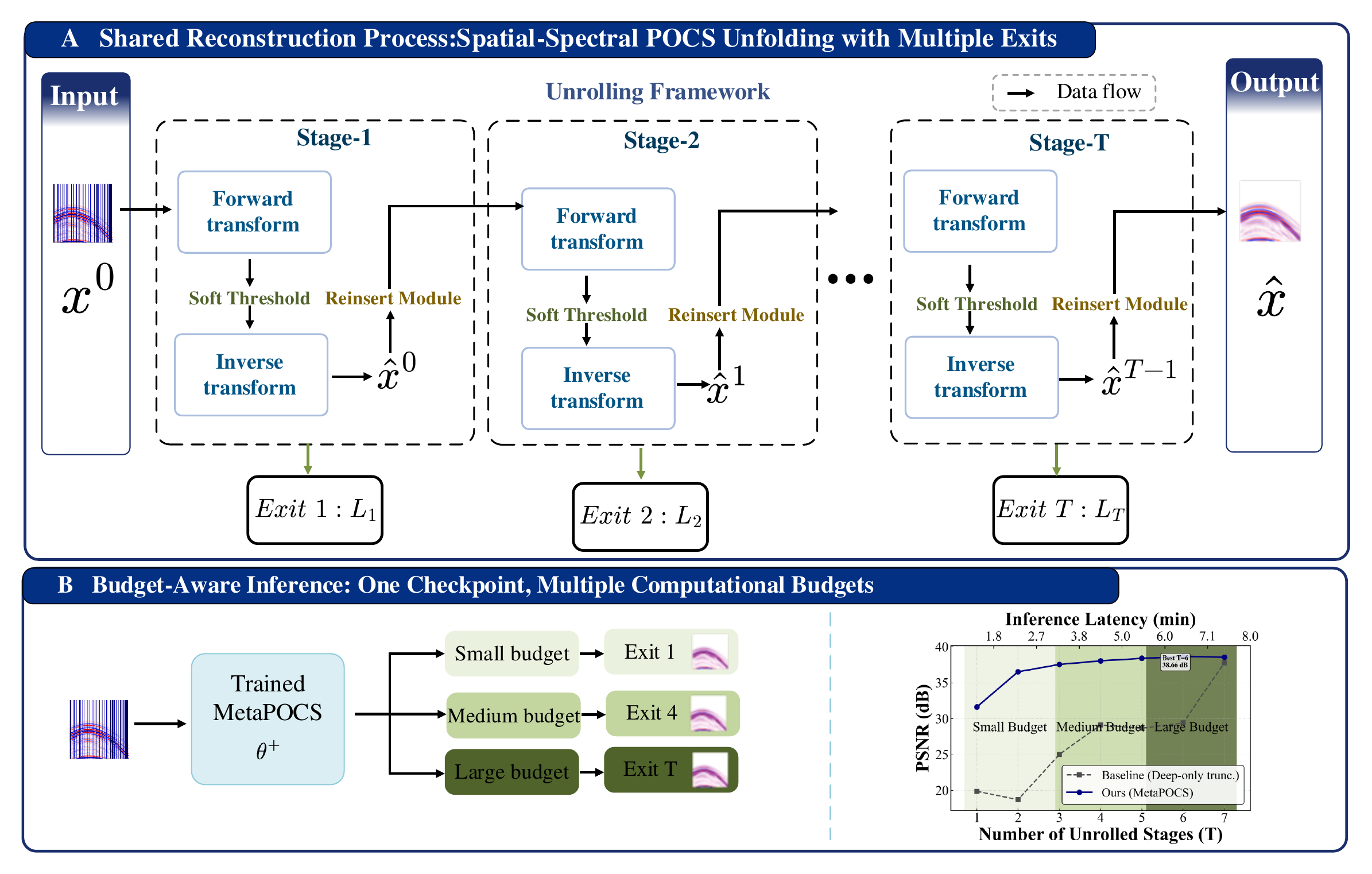}
\caption{MetaPOCS structure. Each stage combines an implicit time--space/Fourier prior with exact data consistency (DC), and each trained exit receives explicit time--space and spectral supervision. Reconstruction stages inherit SSI-POCS; multi-exit coordination and deployment are the focus here. The lower block operates during training. Inference runs a budget-selected prefix, without computing all exits and subsequently discarding their cost.}\label{fig:workflow}
\end{figure}

\begin{figure}[t]
	\centering
	\hspace{-2mm}
	\includegraphics[width=1\linewidth]{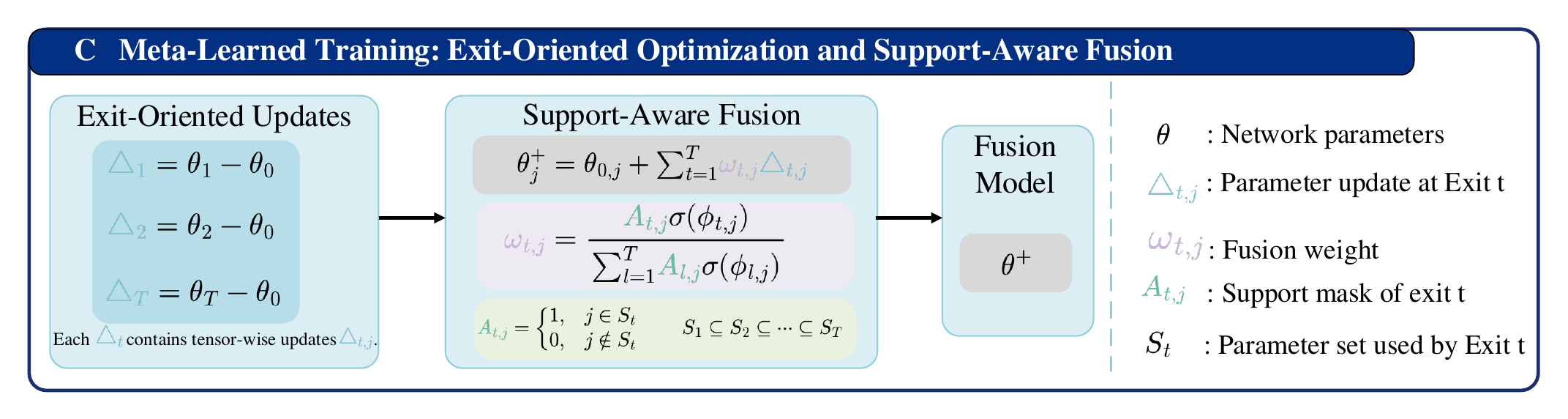}
	\caption{MetaPOCS meta fusion.}\label{fig:metafusion}
\end{figure}

\subsection{Explicit Dual-Domain Exit Supervision and Scalar Meta-Supervision}
The implicit prior controls the representation available to the unfolded stages, whereas the training objective specifies which reconstruction errors that representation must reduce. We supervise every exit explicitly in two complementary views of the same seismic record: samplewise amplitudes on the time--space grid and the complex two-dimensional Fourier spectrum. The common exit objective is
\begin{equation}
\mathcal L_k=\mathcal L_{1,k}+\eta_s\mathcal L_{\mathrm{sym},k}
+\eta_f\mathcal L_{\mathrm{FFL},k}. \label{eq:loss}
\end{equation}
The time--space amplitude term is $\mathcal L_{1,k}=(HW)^{-1}\|\hat x_k-x\|_1$. It directly penalizes samplewise waveform discrepancies and is less dominated by isolated large residuals than a squared error. With noiseless hard reinsertion, its residual is zero on measured samples, while its full-grid normalization retains a dependence on the missing fraction. All exits within a comparison use the same coefficients. Evaluation additionally normalizes errors by the number of missing samples, allowing comparisons across acquisition densities.

The transform term is a prefix mean,
\begin{equation}
\mathcal L_{\mathrm{sym},k}=\frac1{d_k}\sum_{t=1}^{d_k}
\operatorname{MSE}\!\left(G_t(F_t(x^{t-1})),x^{t-1}\right). \label{eq:sym}
\end{equation}
The prefix average makes the strength of transform reg\textbf{}ularization comparable across exits. It encourages the unthresholded analysis--synthesis pair to approximate the identity along the unfolding trajectory. This term regularizes the learned transforms; the explicit dual-domain reconstruction constraints are supplied by the amplitude term and focal frequency loss~\cite{2021jiang}. The latter weights the complex Fourier residual:
\begin{equation}
\mathcal L_{\mathrm{FFL},k}=\frac1{HW}\sum_{u,v}
\omega_{k,uv}|\mathcal F(\hat x_k)_{uv}-\mathcal F(x)_{uv}|^2.
\end{equation}
Writing $r_{k,uv}=\mathcal F(\hat x_k)_{uv}-\mathcal F(x)_{uv}$, the default weighting is
\begin{equation}
\omega_{k,uv}=\sg\!\left(
\frac{|r_{k,uv}|}{\max_{u',v'}|r_{k,u'v'}|}
\right), \label{eq:fflweight}
\end{equation}
with $\omega=0$ for a zero residual spectrum. The Fourier transform is orthonormal, and the spectral weights are held fixed during each loss differentiation. This weighting emphasizes frequencies with large reconstruction discrepancies rather than imposing a fixed high-frequency preference. Because the residual is complex valued, the objective is sensitive to both amplitude and phase discrepancies. For an orthonormal transform, an unweighted squared spectral residual equals its time--space counterpart by Parseval's identity. Here, residual-dependent spectral weights and a time--space $L_1$ term provide distinct error penalties. Thus, the architecture and objective play different roles: Restormer--FFC--DFM supplies an implicit dual-domain hypothesis class, while $\mathcal L_{1,k}$ and $\mathcal L_{\mathrm{FFL},k}$ provide explicit error signals that select a reconstruction within that class. Their empirical complementarity is examined through the crossed prior--objective comparison in Table~\ref{tab:interaction}.

This coupling is especially consequential in a shared multi-exit network. Let $s(j)$ be the stage containing parameter tensor $\theta_j$ and let $\mathcal K_j=\{k:d_k\ge s(j)\}$ collect the exits whose forward paths contain that tensor. Under equal or scalar supervision, its inner gradient is
\begin{equation}
\nabla_{\theta_j}\mathcal L_{\rm in}
=\sum_{k\in\mathcal K_j}\pi_k
\left(\nabla_{\theta_j}\mathcal L_{1,k}
+\eta_f\nabla_{\theta_j}\mathcal L_{\mathrm{FFL},k}
+\eta_s\nabla_{\theta_j}\mathcal L_{\mathrm{sym},k}\right). \label{eq:shareddualgrad}
\end{equation}
Earlier stages receive explicit dual-domain signals from more exits, whereas a stage introduced after depth $d_k$ receives no gradient from exit $k$. The time--space and spectral objectives may agree on some shared tensors and compete on others; different exit depths can introduce an additional source of disagreement. Scalar supervision can change the global contribution of an exit through $\pi_k$, but it scales all parameter tensors and all loss components of that exit together. This limitation motivates the supported, tensor-dependent update fusion developed in the next subsection.

Equal supervision minimizes $K^{-1}\sum_k\mathcal L_k$. A scalar meta-supervision control uses
\begin{align}
\pi_k(\alpha)&=\rho/K+(1-\rho)\operatorname{softmax}(\alpha)_k,\\
\theta^\star(\alpha)&\approx\operatorname{Optimize}_{\theta}
\sum_k\pi_k(\alpha)\mathcal L_k(\mathcal D_{\rm in};\theta),\\
\min_\alpha\quad &\frac1K\sum_k
\mathcal L_k(\mathcal D_{\rm meta};\theta^\star(\alpha)). \label{eq:scalar}
\end{align}
The lower bound $\rho/K$ maintains supervision at every exit; $\rho=1$ recovers equal weights. The dependence of $\theta^\star$ on $\alpha$ is approximated through inner optimization using the multilevel differentiation framework Betty~\cite{2023choebetty}. The outer objective assigns equal importance to all exits, so decreasing a training weight does not directly remove that exit from the meta objective. This scalar formulation changes the relative contribution of each exit uniformly across its parameters.

\subsection{Meta-Learned Fusion of Exit-Oriented Updates}
The principal training variant adapts Meta-GF~\cite{Sun2022MetaGF}. Each exit-oriented trajectory starts from the same parameters $\theta_0$ and optimizer state, using
\begin{equation}
\mathcal J_k=\mathcal L_k+\beta\sum_{\ell\ne k}\mathcal L_\ell,\qquad 0\le\beta\le1. \label{eq:aux}
\end{equation}
After the prescribed inner traversal, $\Delta_{k,j}=\theta_{k,j}-\theta_{0,j}$ is the displacement of parameter tensor $j$. Auxiliary terms train other exits during each trajectory. Displacements are cached and detached for the outer update.

Let $A_{k,j}$ indicate whether tensor $j$ belongs to exit $k$'s inference prefix. Define
\begin{align}
w_{k,j}(\phi)&=\frac{A_{k,j}\sigma(\phi_{k,j})}
{\sum_{\ell=1}^{K}A_{\ell,j}\sigma(\phi_{\ell,j})}, \label{eq:weights}\\
\theta_j^+(\phi)&=\theta_{0,j}+\sum_{k=1}^{K}w_{k,j}(\phi)
\sg(\Delta_{k,j}), \label{eq:fuse}\\
\min_\phi\quad &\frac1K\sum_{k=1}^{K}
\mathcal L_k(\mathcal D_{\rm meta};\theta^+(\phi)). \label{eq:outer}
\end{align}
Every backbone tensor has at least one supported exit. For independent stage parameters, $A_{k,j}=1$ precisely when tensor $j$ belongs to a stage no deeper than $d_k$; for stage-shared parameters, all exits are supported. The weights are nonnegative and sum to one over the supported exits of each tensor. Thus, $\theta_j^+$ lies in the convex hull of the corresponding candidate tensor values, although neither the reconstruction network nor its loss is convex. Auxiliary objectives can move parameters outside the target exit's prefix, but the support mask excludes those displacements from that exit's fusion contribution.

After the outer updates, exponential smoothing gives $\phi\leftarrow\mu\phi_{\rm old}+(1-\mu)\phi$, where $\mu\in[0,1)$. The corresponding fused parameters initialize the next round. Each displacement already includes the inner learning rate and optimizer effects, so Eq.~\eqref{eq:fuse} uses it directly. The outer derivative passes through the fused model and its weights while treating the candidate trajectories as fixed.

To expose the weighting mechanism, define $\bar\Delta_j=\sum_k w_{k,j}\Delta_{k,j}$ and let $h_j$ be the gradient of the equal-exit outer objective with respect to $\theta_j^+$. For a supported exit $\ell$, differentiating the normalized sigmoid weights yields
\begin{align}
\frac{\partial\theta_j^+}{\partial\phi_{\ell,j}}
&=w_{\ell,j}\bigl(1-\sigma(\phi_{\ell,j})\bigr)
(\Delta_{\ell,j}-\bar\Delta_j), \label{eq:fusionjac}\\
\frac{\partial\mathcal L_{\rm out}}{\partial\phi_{\ell,j}}
&=w_{\ell,j}\bigl(1-\sigma(\phi_{\ell,j})\bigr)
\left\langle h_j,\Delta_{\ell,j}-\bar\Delta_j\right\rangle. \label{eq:hypergradient}
\end{align}
The inner product is taken over all entries of tensor $j$; unsupported logits have zero derivative. Equation~\eqref{eq:hypergradient} compares an exit-oriented displacement with the currently fused displacement under the outer loss gradient. A candidate that is locally more favorable than the current mixture receives a negative logit gradient and hence an increased weight under gradient descent. The assessment is tensor dependent, unlike scalar loss reweighting.

This derivative characterizes local weight adaptation, rather than guaranteeing simultaneous descent of every exit. It also identifies a limiting case: when all candidate displacements coincide, the fusion hypergradient vanishes. At $\beta=1$, the task objectives are identical, and identical minibatch orders and stochastic states produce this collapse. Both observations motivate the finite-update diagnostics in Section~\ref{sec:mechanism}.

\begin{algorithm}[t]
\caption{A full MetaPOCS fusion round}\label{alg:meta}
\begin{algorithmic}[1]
\REQUIRE Parameters $\theta_0$, optimizer state $s_0$, logits $\phi$, exits $\mathcal E$, inner/meta data.
\FOR{$k=1,\ldots,K$}
\STATE Restore $(\theta,s)=(\theta_0,s_0)$.
\STATE Optimize $\mathcal J_k$ for the prescribed traversal.
\STATE Store detached displacements $\Delta_k=\theta-\theta_0$.
\ENDFOR
\STATE Restore the base model; save old logits $\phi_{\rm old}$.
\FOR{each permitted meta minibatch}
\STATE Form $\theta^+(\phi)$ by Eqs.~\eqref{eq:weights}--\eqref{eq:fuse}.
\STATE Update only $\phi$ using objective~\eqref{eq:outer}.
\ENDFOR
\STATE Optionally set $\phi\leftarrow\mu\phi_{\rm old}+(1-\mu)\phi$.
\STATE Install fused parameters $\theta\leftarrow\theta^+(\phi)$.
\STATE Record steps, wall time, and per-exit quality.
\end{algorithmic}
\end{algorithm}

\subsection{Optimization Schedule and Computational Cost}
Algorithm~\ref{alg:meta} describes a full fusion round. If an inner traversal contains $N_{\rm in}$ minibatches and the outer update uses $N_{\rm meta}$ minibatches, the round performs $KN_{\rm in}$ task updates and $N_{\rm meta}$ meta updates. Its training cost therefore exceeds that of one equal-supervision traversal. The experimental comparison distinguishes data exposure from elapsed GPU time. Fusion introduces $KJ$ logits for $J$ parameter tensors and storage for candidate displacements; these quantities are training overhead, since the fused reconstruction backbone alone is needed for inference.

A periodic schedule reduces the number of full rounds by refreshing the fusion weights every $m$ outer epochs after a warm-up. Intermediate epochs either minimize the equal-exit mean or combine current task gradients with frozen fusion weights. The latter operates on instantaneous gradients rather than the cached multi-step displacements of Eq.~\eqref{eq:fuse}, so the two schedules define separate optimization variants.

The meta batch distribution is denoted by $\mathcal D_{\rm meta}$. It can reuse the training distribution or be defined by disjoint acquisition units. These choices distinguish training-objective coordination from held-out meta supervision; checkpoint selection uses a separate monitoring role in the latter setting. Across fixed- and multi-exit controls, Eq.~\eqref{eq:sym} defines the common regularization scale. A fixed-depth implementation returning the sum of stage losses therefore uses coefficient $\eta_s/d$. The original SSI-POCS recipe forms a separate reference for the relationship to the preceding work.

\section{Experimental Setup}\label{sec:protocol}
\noindent\textit{Numerical completion note:} In this revision, \pending\ marks an unverified numerical entry, and framed panels reserve space for experimental figures. The following sections define the comparison protocol and analysis; no result is inferred from a placeholder.
\subsection{Data, Splits, and Acquisition Geometry}
The evaluation comprises synthetic SEG C3 data, North Sea field MAVO data, and controlled acoustic models for downstream imaging. SEG C3 represents structurally complex synthetic wavefields, whereas MAVO provides field acquisition characteristics. The corresponding patch dimensions are $128\times128$ and $256\times112$. Table~\ref{tab:dataset_params_en} defines the acquisition-level partitions and physical metadata, including a reserved external-survey test. Field reconstruction accuracy is evaluated by withholding recorded traces so that a reference remains available. Regions that were never recorded support qualitative interpretation but have no direct reconstruction-error reference.

The partitioning unit is a shot gather or a spatial acquisition block rather than an extracted patch. The protocol assigns each unit to one partition before patch extraction; spatial buffers separate strongly correlated neighboring blocks. Training, held-out meta supervision, monitoring, and testing have distinct roles. The source/receiver identifiers, axis order, temporal crop, and physical sampling interval accompany each partition. Consequently, the number of independent acquisition units is distinct from the number of overlapping training patches.

\begin{table*}[t]
	\caption{Datasets Information.}
	\label{tab:dataset_params_en}
	\centering
	\resizebox{\textwidth}{!}{
		\begin{tabular}{l c c c c c c}
			\toprule
			Dataset & Shots & Receivers per Shot & Sample Time (\(N_t\)) & Time Sample Interval (\(\Delta t\)) & Spatial Sampling Interval (\(\Delta s\)) & Train/val/test \\
			\midrule
			SEGC3 & \(45\) & \(201 \times 201\) & \(625\) & \(8 \text{ ms}\) & \(20 \text{ m}\) & 25000/10000/10000 \\
			MAVO & \(1001\) & \(120\) & \(1500\) & \(4 \text{ ms}\) & \(25 \text{ m}\) & 10000/4000/6000 \\
			F3 & 600 & 440  & 951 & 4\text{ ms}  &\(25\text{ m }\)& 8000/3000/6000 \\
			Model94 & \(277\) & \(480\) & \(2000\) & \(4 \text{ ms}\) & \(15 \text{ m}\) & 10000/3000/3500 \\
			\bottomrule
		\end{tabular}
	}
\end{table*}

Three sampling patterns characterize acquisition incompleteness: random trace removal, consecutive gaps, and periodic decimation. The primary missing fractions are $r\in\{0.3,0.5,0.7\}$, with $r=0.9$ defining a severe-undersampling condition. A test mask is defined on the full gather before patch extraction and remains identical across methods. Repeated mask realizations are nested within the same acquisition unit. Consecutive gaps are characterized by both their trace fraction and physical width where coordinates are available. In-distribution comparisons use matching training and test conditions; transfer experiments hold the trained model fixed while changing the mask, density, or survey.

The amplitude protocol uses a common reversible normalization determined from observed traces or a training-only calibration. For a scale $s_y>0$, the normalized input and target are $y/s_y$ and $x/s_y$, and the physical prediction is $s_y\hat x_{\rm norm}$. The scale is independent of unobserved test amplitudes. Waveform and migration metrics operate in the restored amplitude range, using identical preprocessing across methods. Tracewise gain normalization would alter relative amplitudes and is excluded from this comparison.

\subsection{Controls and Implementation Settings}
The reference configuration contains $T=8$ stages, a feature width of 16, and four attention heads. Stage parameters are independent, and $\mathcal E=\{2,3,\ldots,8\}$ gives seven supervised exits. The operating points $d\in\{2,4,6,8\}$ summarize the budget range. The common reconstruction coefficients are $(\eta_s,\eta_f)=(0.1,100)$; the objective ablation removes the spectral term while retaining the same amplitude supervision and transform regularization. The optimization configuration uses Adam, initial inner and outer learning rates of $3\times10^{-4}$, batch size four, $\beta=0.4$, and $\mu=0.9$. The final training budget and hardware-dependent measurements remain represented by \pending. Equal-exposure comparisons count task and meta updates separately, while equal-time comparisons use total measured GPU time.

\begin{table*}[t]
\centering\small
\caption{Training and deployment controls. Matched comparisons share the reconstruction backbone, loss scaling, acquisition partitions, and masks.}\label{tab:controls}
\begin{tabular}{P{30mm}P{43mm}P{79mm}}
\toprule
Control & Training / deployment & Isolated factor\\
\midrule
SSI-POCS original & Original recipe, final exit & Relationship to the preceding reconstruction method\\
Fixed-$d$ matched & Independent model at each $d=2,4,6,8$ & Specialized training for each deployment budget\\
Deep-only truncated & Depth-8 supervision; prefix inference & Intermediate reconstruction without direct supervision\\
Equal multi-exit & Average identical exit losses & Direct supervision at several reconstruction depths\\
Scalar meta & Eq.~\eqref{eq:scalar} & Global redistribution of exit loss weights\\
PCGrad / GradNorm & Common exits and objectives & Gradient-direction or gradient-scale balancing\\
Uniform displacement fusion & Same trajectories; uniform supported weights & Extra inner optimization with unlearned fusion\\
Unrestricted-support fusion & All auxiliary-connected candidates participate & Effect of the prefix support restriction\\
MetaPOCS & Eqs.~\eqref{eq:aux}--\eqref{eq:outer}; one checkpoint & Tensor-dependent fusion of supported updates\\
\bottomrule
\end{tabular}
\end{table*}

The reconstruction comparison includes Curvelet-POCS, POCS-Net~\cite{2023Chen}, SP-Net~\cite{2024Wu}, and a Restormer-based end-to-end reconstruction baseline~\cite{2022Restormer}. The matched controls in Table~\ref{tab:controls} isolate training effects on the SSI-POCS backbone. The original SSI-POCS recipe is separated from the fixed-depth matched control because its loss coefficients differ. Uniform displacement fusion uses the same exit-oriented trajectories as MetaPOCS, replacing the learned weights with $w_{k,j}=A_{k,j}/\sum_\ell A_{\ell,j}$. Unrestricted-support fusion instead permits every auxiliary-connected candidate to contribute to each tensor. MetaPOCS remains a Meta-GF-based method throughout these comparisons.

\subsection{Metrics, Timing, and Statistical Units}
Let $\mathcal U=\{i:M_i=0\}$ and $e=\hat x-x$. Primary metrics are
\begin{align}
\operatorname{MAE}_{\mathcal U}&=\|Q\odot e\|_1/|\mathcal U|,\\
\operatorname{SNR}_{\mathcal U}&=10\log_{10}
\frac{\|Q\odot x\|_F^2}{\|Q\odot e\|_F^2},\\
E_{\rm DC}&=\frac{\|M\odot(\hat x-y)\|_F}{\|M\odot y\|_F+\epsilon}.
\end{align}
Full-gather SNR and SSIM~\cite{wang2004imagequa} complement the missing-region metrics. SSIM is used only as an evaluation metric, with the same window and amplitude range across methods. The consistency error $E_{\rm DC}$ measures preservation of acquired samples, whereas missing-region SNR measures interpolation. Windows with zero reference energy are excluded from SNR and retained in absolute-error summaries. Fixed reflection windows additionally characterize signed amplitude bias, waveform correlation, and arrival-time error. Spectral comparisons use a common orthonormal Fourier transform and physical frequency--wavenumber axes.

The replication protocol specifies three training seeds for the main comparisons, with five for the principal meta-versus-equal contrast. Seed-to-seed dispersion describes optimization variability. Uncertainty over acquisition units is summarized by paired bootstrap confidence intervals, resampling shots or buffered blocks while retaining matched methods, masks, and exits. Thus, repeated masks and overlapping patches remain nested observations rather than additional independent geological samples. The principal effect is the paired change in missing-region SNR at matched inference cost; the deepest-exit change quantifies any accompanying accuracy trade-off.

Inference profiling measures the selected prefix with batch size one, fixed input dimensions, and a common numerical precision. Device warm-up and synchronization precede timed measurements. Median and 95th-percentile latency describe central cost and runtime variability, respectively; peak allocated memory describes the executed prefix. Network time is separated from preprocessing and gather assembly. Operation counts identify the coverage of Fourier and attention kernels. Training GPU hours include all task trajectories, meta updates, and validation, while checkpoint storage measures the cost of retaining deployable models.

\section{Reconstruction and Budget Evaluation}\label{sec:results}
\subsection{Main Interpolation Comparison}
Table~\ref{tab:main} organizes the full-depth comparison by dataset and sampling pattern at 50\% trace removal. Random removal retains distributed observations, whereas a consecutive gap removes the local support of events within an extended interval. Separating these conditions distinguishes interpolation between available traces from reconstruction across a missing region. Comparisons with conventional and end-to-end methods characterize the overall reconstruction method; comparisons with fixed-8 and equal multi-exit training isolate the effect of supervising and coordinating intermediate exits.

\begin{figure*}[t]
	\centering
	\subfloat[Missing]{\includegraphics[width=0.16\linewidth]{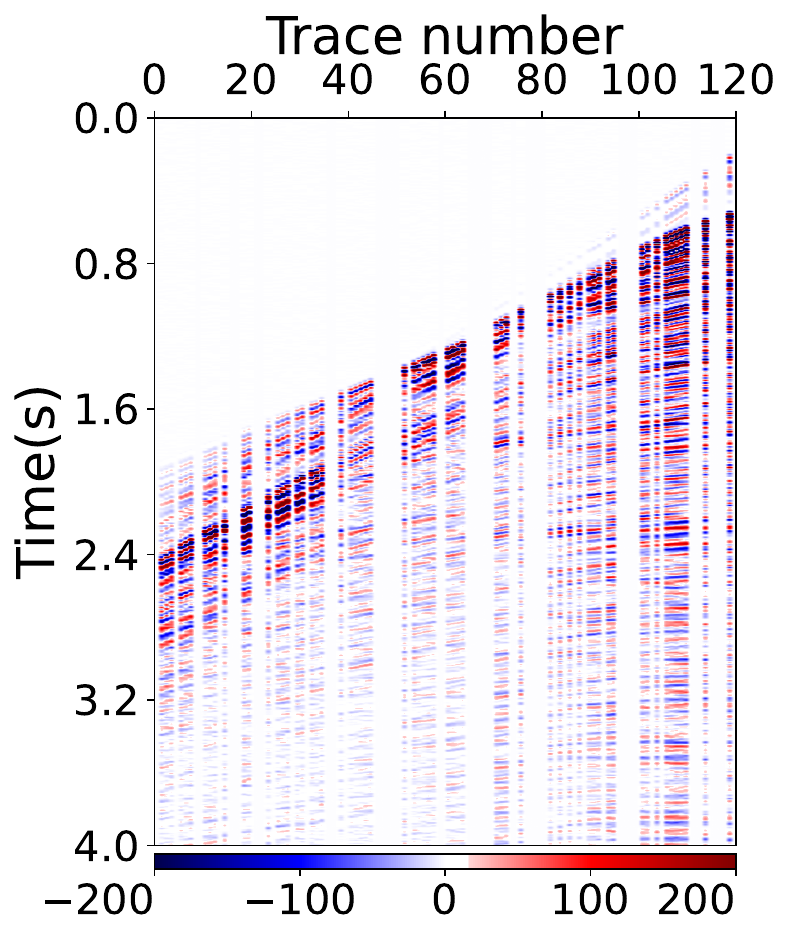}\label{fig:evo_d3_miss}}
	\hfill
	\subfloat[D3UNet (T1)]{\includegraphics[width=0.16\linewidth]{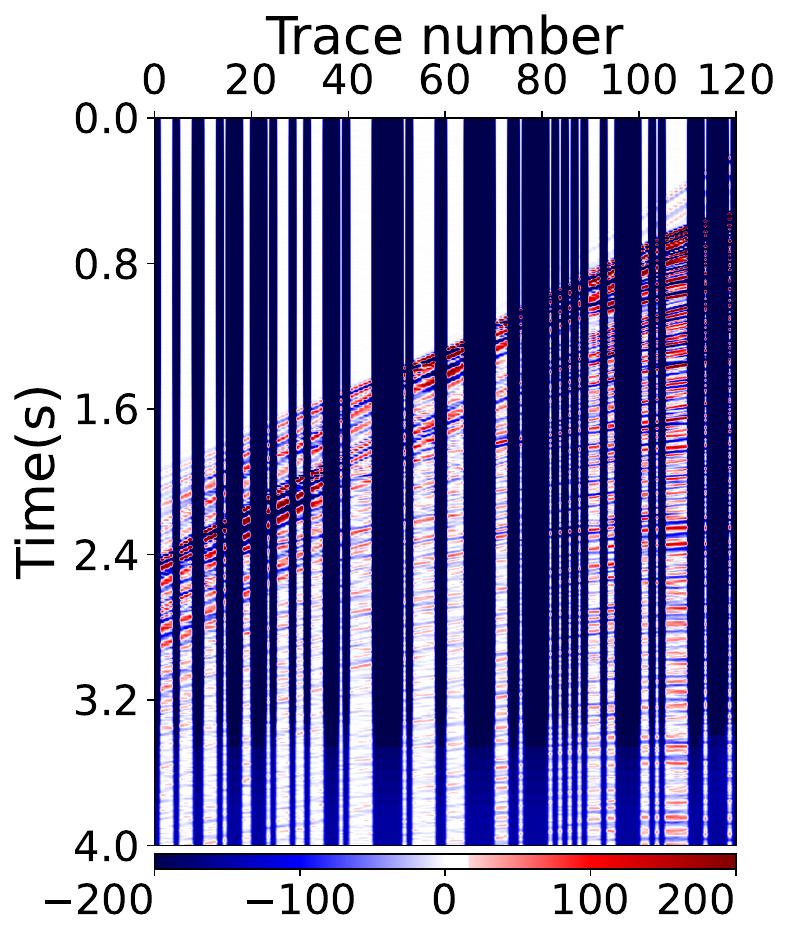}\label{fig:evo_d3_t1}}
	\hfill
	\subfloat[D3UNet (T3)]{\includegraphics[width=0.16\linewidth]{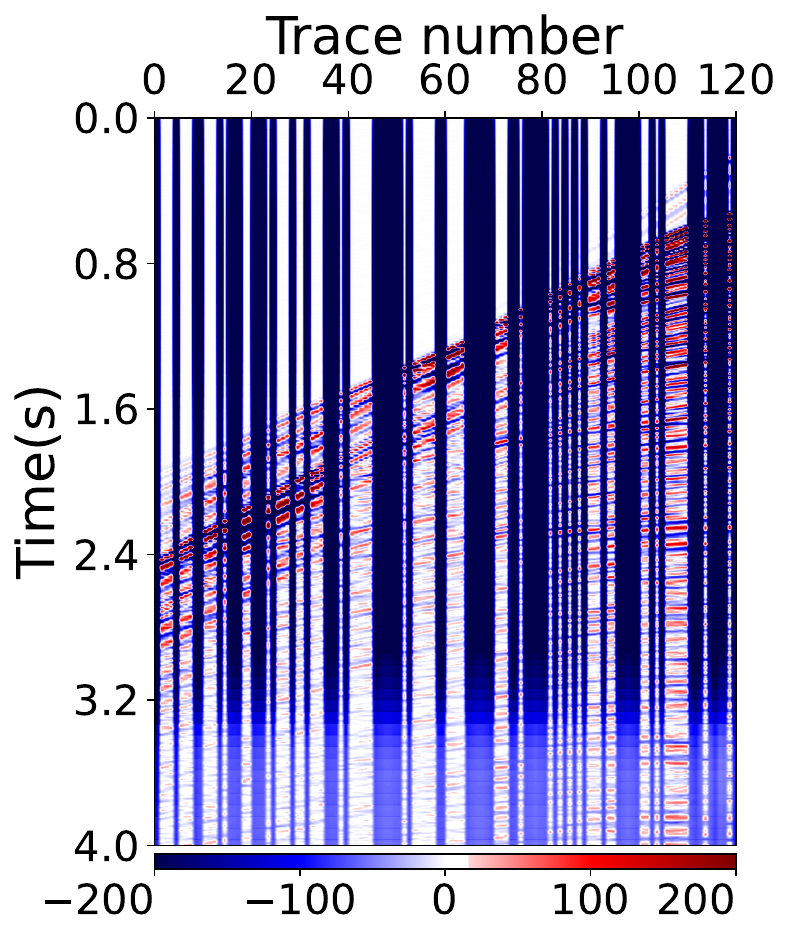}\label{fig:evo_d3_t3}}
	\hfill
	\subfloat[D3UNet (T5)]{\includegraphics[width=0.16\linewidth]{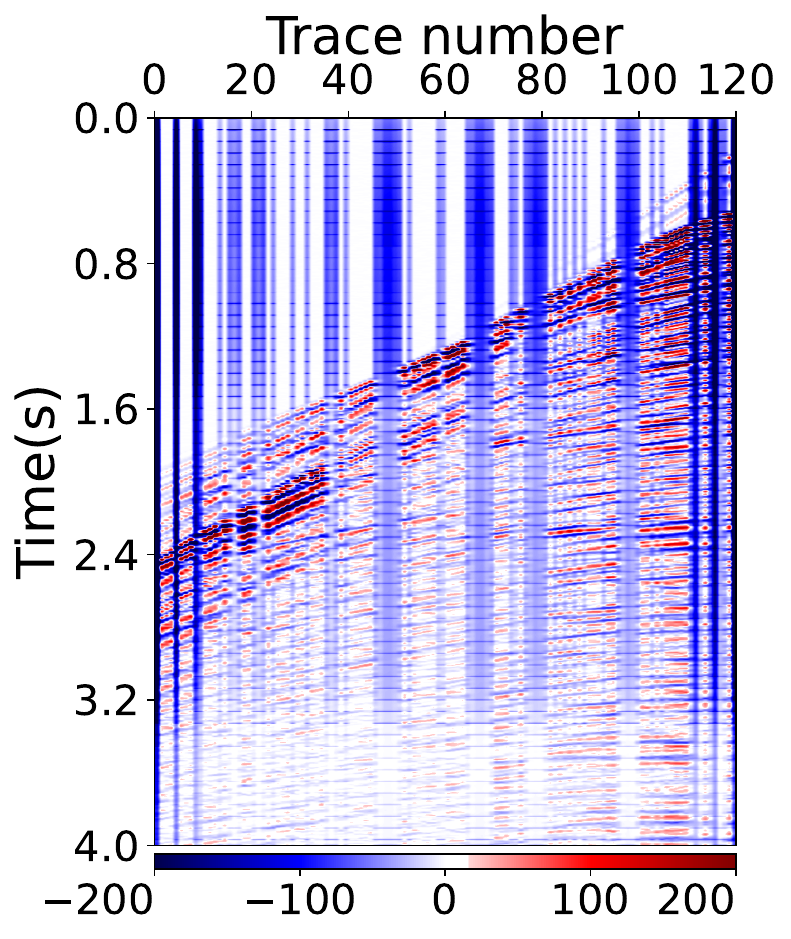}\label{fig:evo_d3_t5}}
	\hfill
	\subfloat[D3UNet (T8 - Final)]{\includegraphics[width=0.16\linewidth]{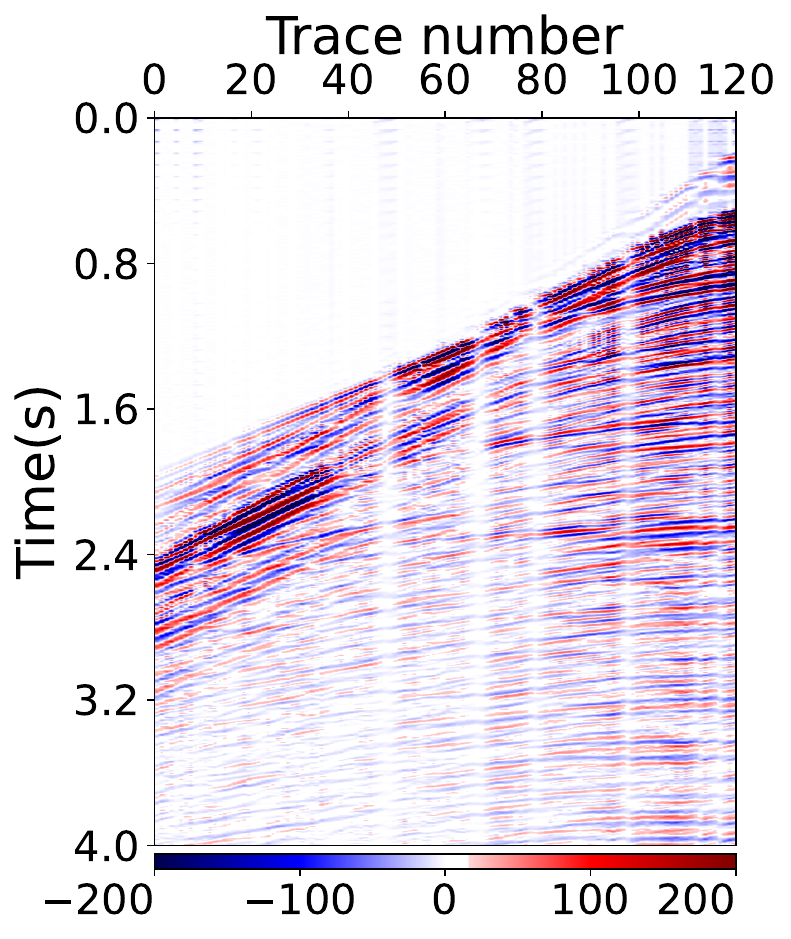}\label{fig:evo_d3_t8}}
	\hfill
	\subfloat[Reference]{\includegraphics[width=0.16\linewidth]{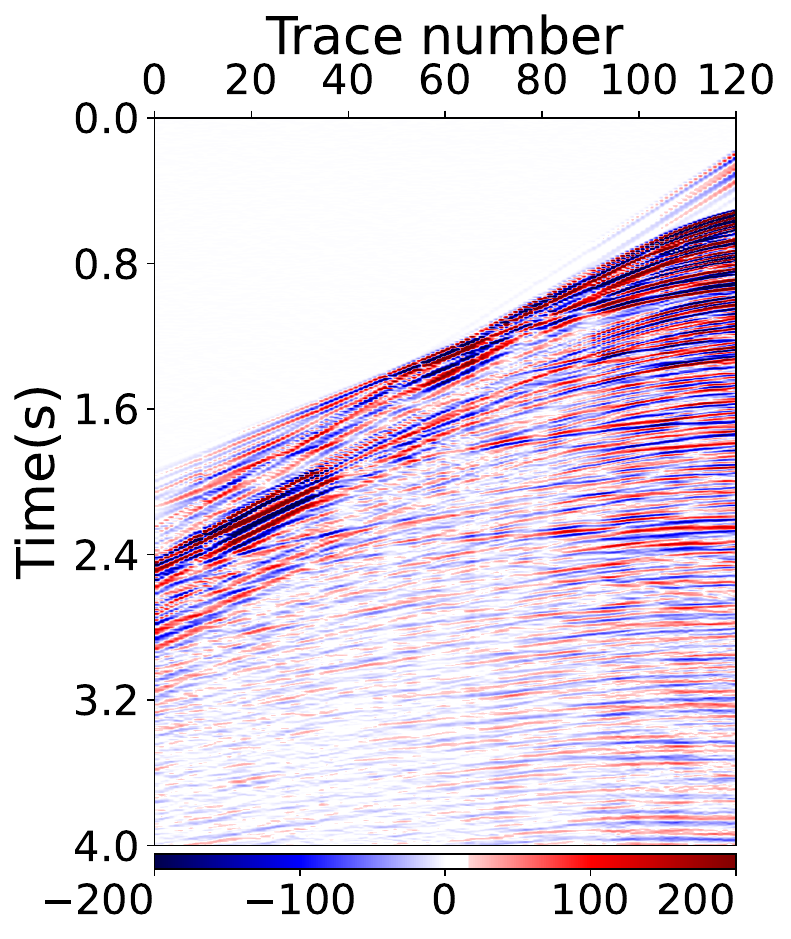}\label{fig:evo_d3_ori}}
	\\
	\vspace{-2mm}
	
	\subfloat[Missing]{\includegraphics[width=0.16\linewidth]{figures/fusion/shot_474_Miss.pdf}\label{fig:evo_fd_miss}}
	\hfill
	\subfloat[Fixed-depth (T1)]{\includegraphics[width=0.16\linewidth]{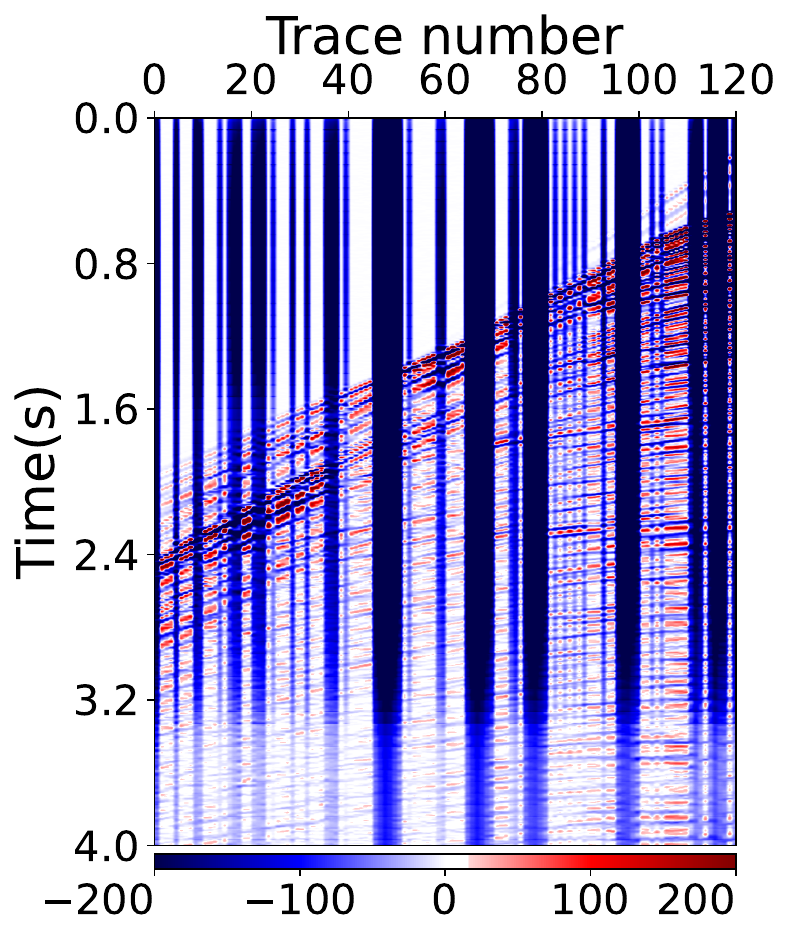}\label{fig:evo_fd_t1}}
	\hfill
	\subfloat[Fixed-depth (T3)]{\includegraphics[width=0.16\linewidth]{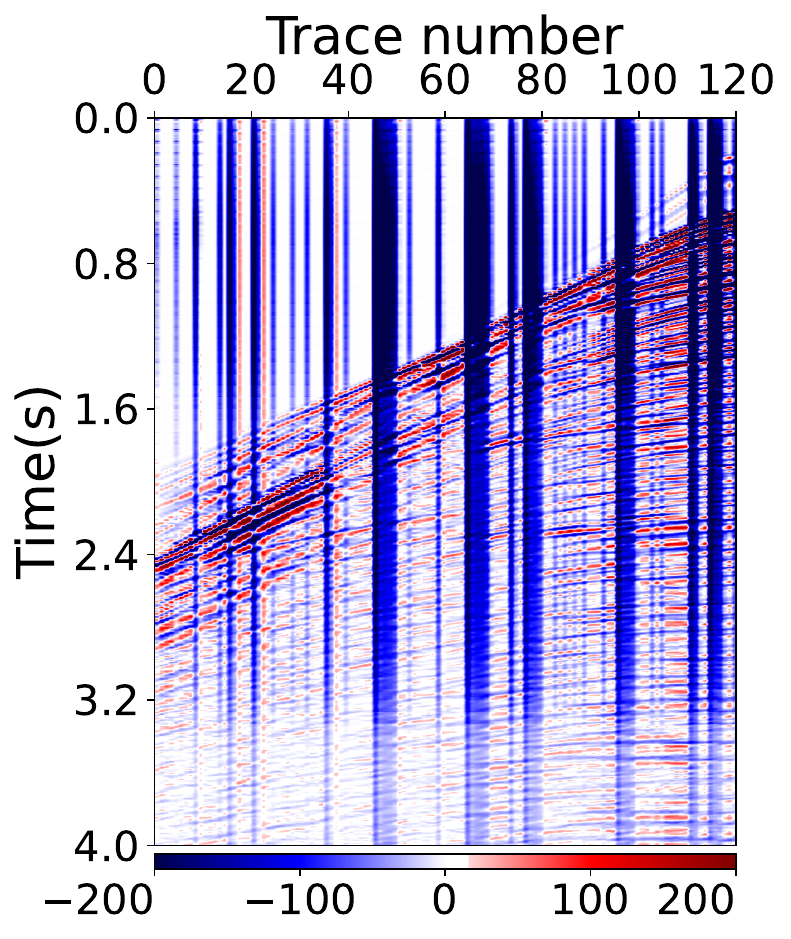}\label{fig:evo_fd_t3}}
	\hfill
	\subfloat[Fixed-depth (T5)]{\includegraphics[width=0.16\linewidth]{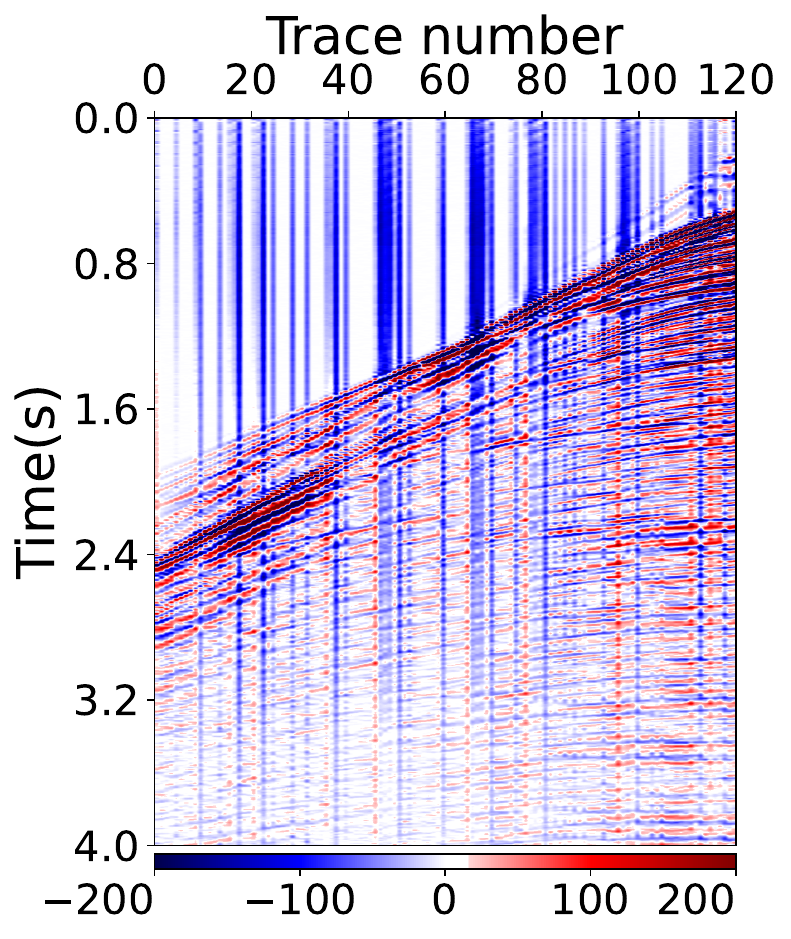}\label{fig:evo_fd_t5}}
	\hfill
	\subfloat[Fixed-depth (T7 - Final)]{\includegraphics[width=0.16\linewidth]{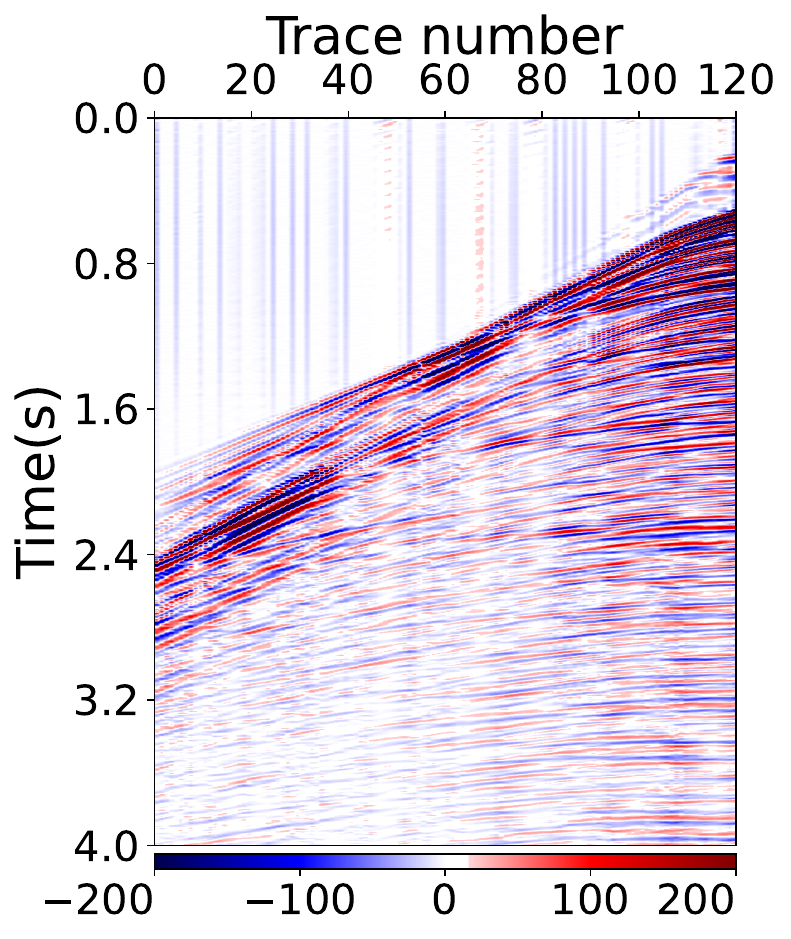}\label{fig:evo_fd_t7}}
	\hfill
	\subfloat[Reference]{\includegraphics[width=0.16\linewidth]{figures/fusion/shot_474_Ori.pdf}\label{fig:evo_fd_ori}}
	\\
	\vspace{-2mm}

	\subfloat[Missing]{\includegraphics[width=0.16\linewidth]{figures/fusion/shot_474_Miss.pdf}\label{fig:evo_meta_miss}}
	\hfill
	\subfloat[MetaPOCS (T1)]{\includegraphics[width=0.16\linewidth]{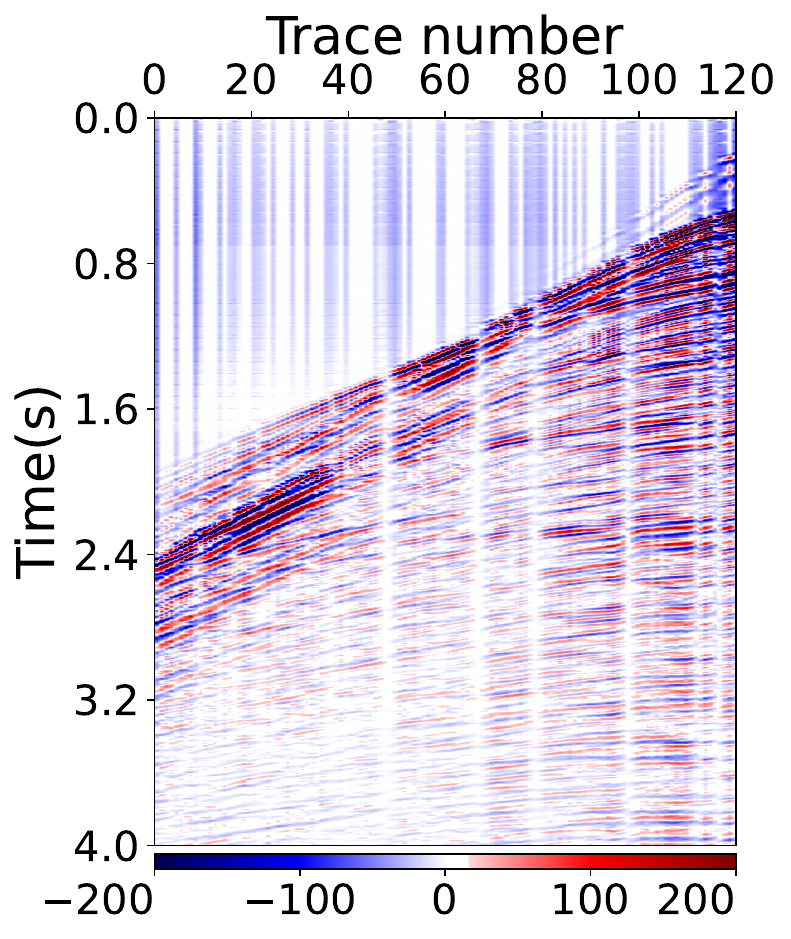}\label{fig:evo_meta_t1}}
	\hfill
	\subfloat[MetaPOCS (T3)]{\includegraphics[width=0.16\linewidth]{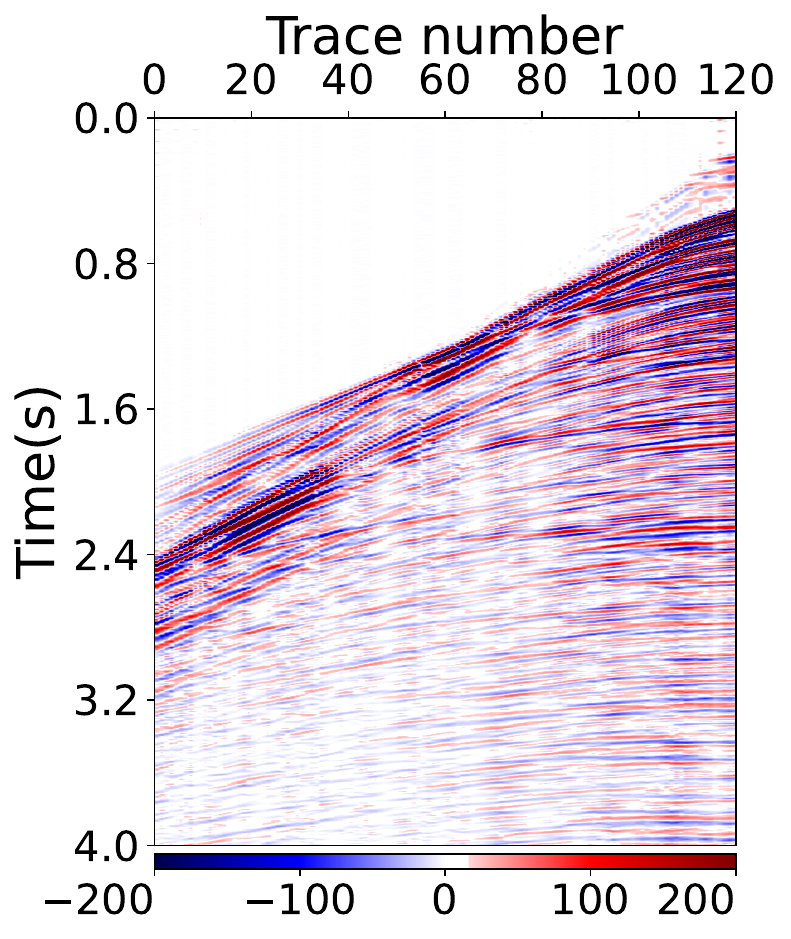}\label{fig:evo_meta_t3}}
	\hfill
	\subfloat[MetaPOCS (T5)]{\includegraphics[width=0.16\linewidth]{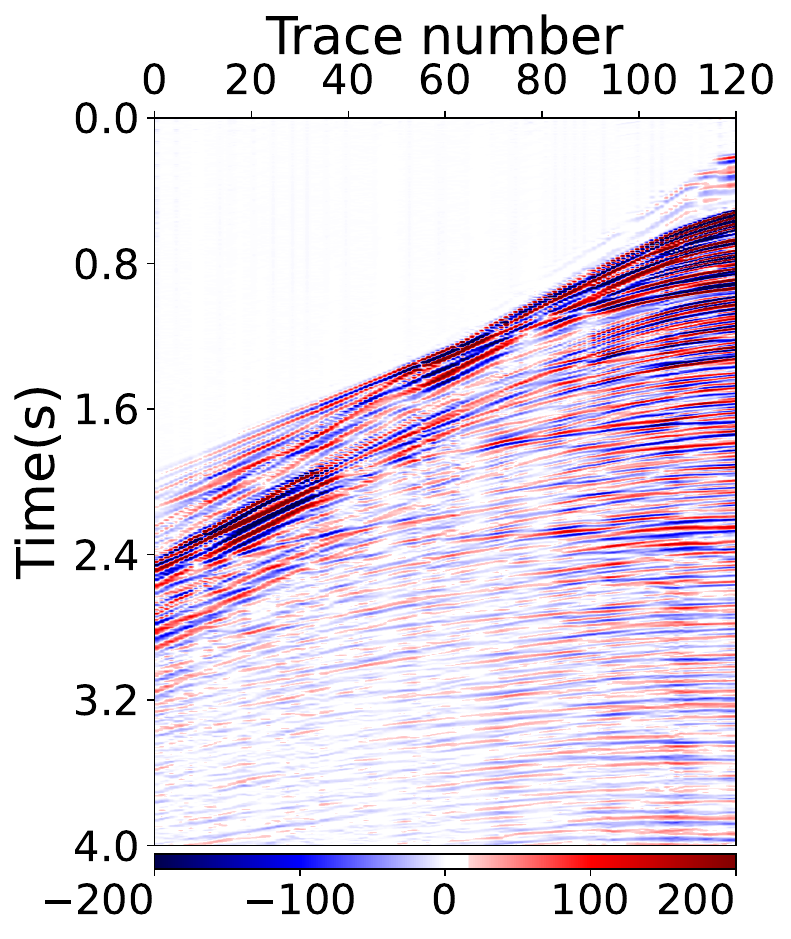}\label{fig:evo_meta_t5}}
	\hfill
	\subfloat[MetaPOCS (T7 - Final)]{\includegraphics[width=0.16\linewidth]{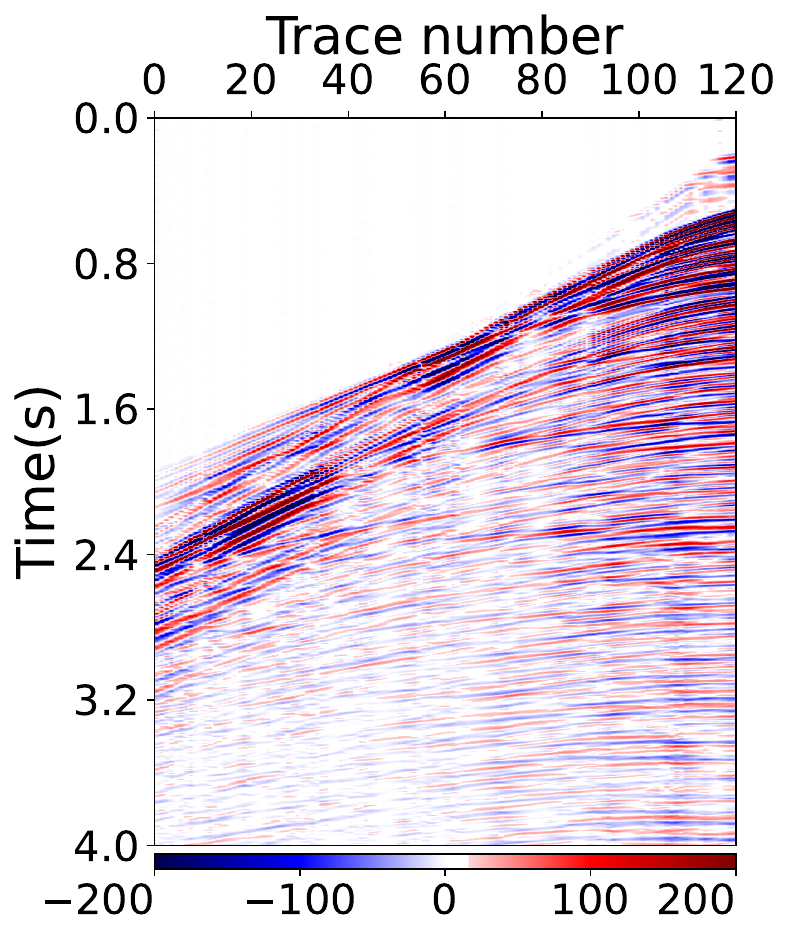}\label{fig:evo_meta_t7}}
	\hfill
	\subfloat[Reference]{\includegraphics[width=0.16\linewidth]{figures/fusion/shot_474_Ori.pdf}\label{fig:evo_meta_ori}}
	\par
	
	\caption{Panoramic view of physical iteration evolution across different unrolling mechanisms and supervision strategies (shot 474). The matrices visualize the step-wise wavefield reconstruction from initial iterations (T1, T3) to the final stage. Notably, traditional fixed-depth schemes (Rows 1-2) exhibit slow convergence, while the `Equal multi-exit` mechanism (Row 3) introduces severe gradient conflicts, blurring the intermediate wavefields (T3, T5). In contrast, the proposed `MetaPOCS` (Row 5), driven by dynamic bi-level meta-supervision, resolves early-stage feature entanglements rapidly, achieving unparalleled amplitude fidelity and structural continuity as early as T3, and yields the optimal final reconstruction.}
	\label{fig:panoramic_evolution}
\end{figure*}

\begin{figure}[t]
	\centering
	\includegraphics[width=\columnwidth]{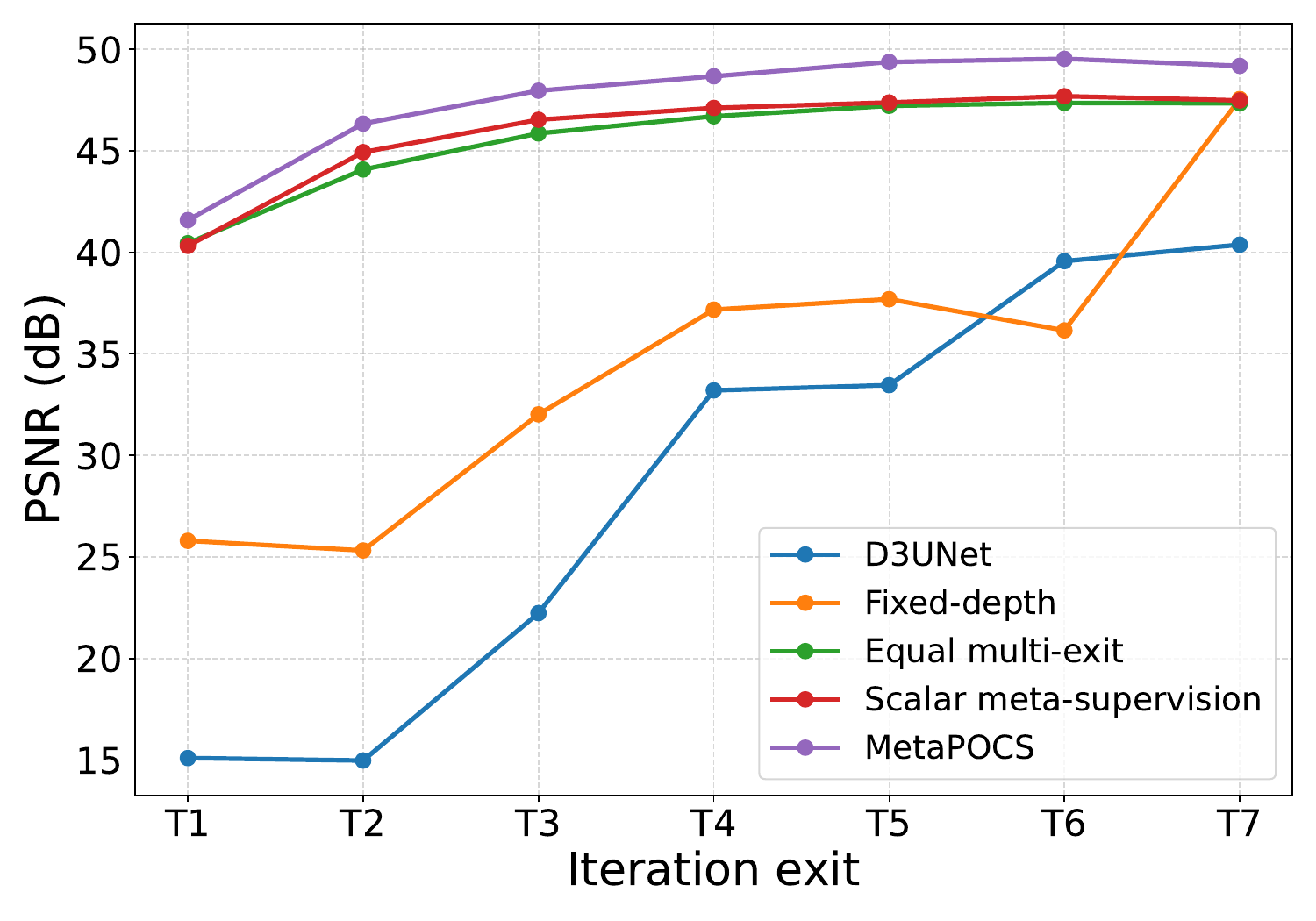}
	\caption{Quantitative stage-wise PSNR evolution trajectories under different supervision mechanisms (averaged across 8 test shots). The proposed MetaPOCS effectively overcomes the gradient conflicts observed in the Equal multi-exit baseline during early stages (T1--T4) and consistently maintains the highest reconstruction accuracy, demonstrating the superiority of the dynamic bi-level meta-supervision strategy.}
	\label{fig:fusion_exit_psnr_curve}
\end{figure}

To quantitatively validate the convergence efficiency and stability of the proposed meta-supervision mechanism, we evaluate the stage-wise reconstruction accuracy. As illustrated in Fig.~\ref{fig:fusion_exit_psnr_curve}, we plot the average PSNR evolution trajectories across 8 test shots at each unrolled iteration exit (T1 to T7). 

It is evident that the \textit{Fixed-depth} baseline exhibits sluggish convergence, as it lacks intermediate constraints. The \textit{Equal multi-exit} strategy, while constrained at every stage, suffers from severe gradient conflicts during the early iterations (T1--T4), leading to suboptimal PSNR gains. Although the \textit{Scalar meta-supervision} mitigates this issue to some extent, it lacks spatial adaptivity. In striking contrast, the proposed \textit{MetaPOCS}, driven by the bi-level dynamic meta-supervision, exhibits a remarkably steep and stable PSNR ascent. It not only resolves the early-stage feature entanglements but also achieves the highest upper bound of reconstruction accuracy at the final stage, corroborating the visual observations and proving the necessity of the proposed adaptive framework.
\begin{figure*}[t]
	\centering
	\hspace{-3mm}
	\subfloat[Reference]{\includegraphics[height=1.5in]{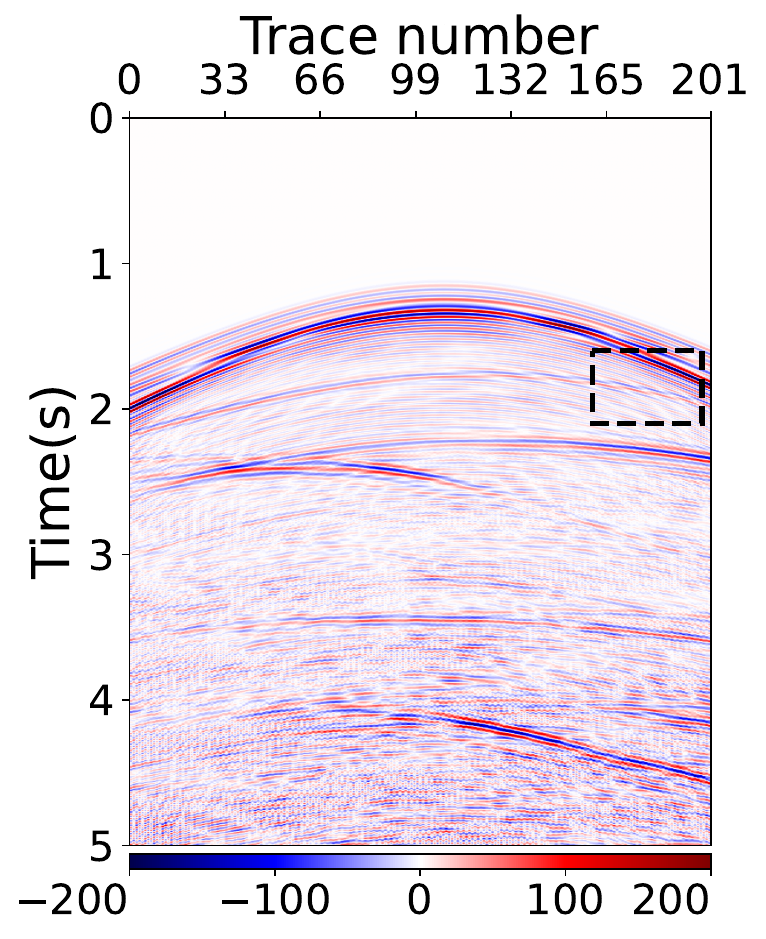}\label{fig:segc3_12_4_Ori}}
	\hspace{0mm}
	\subfloat[Curvelet]{\includegraphics[height=1.5in]{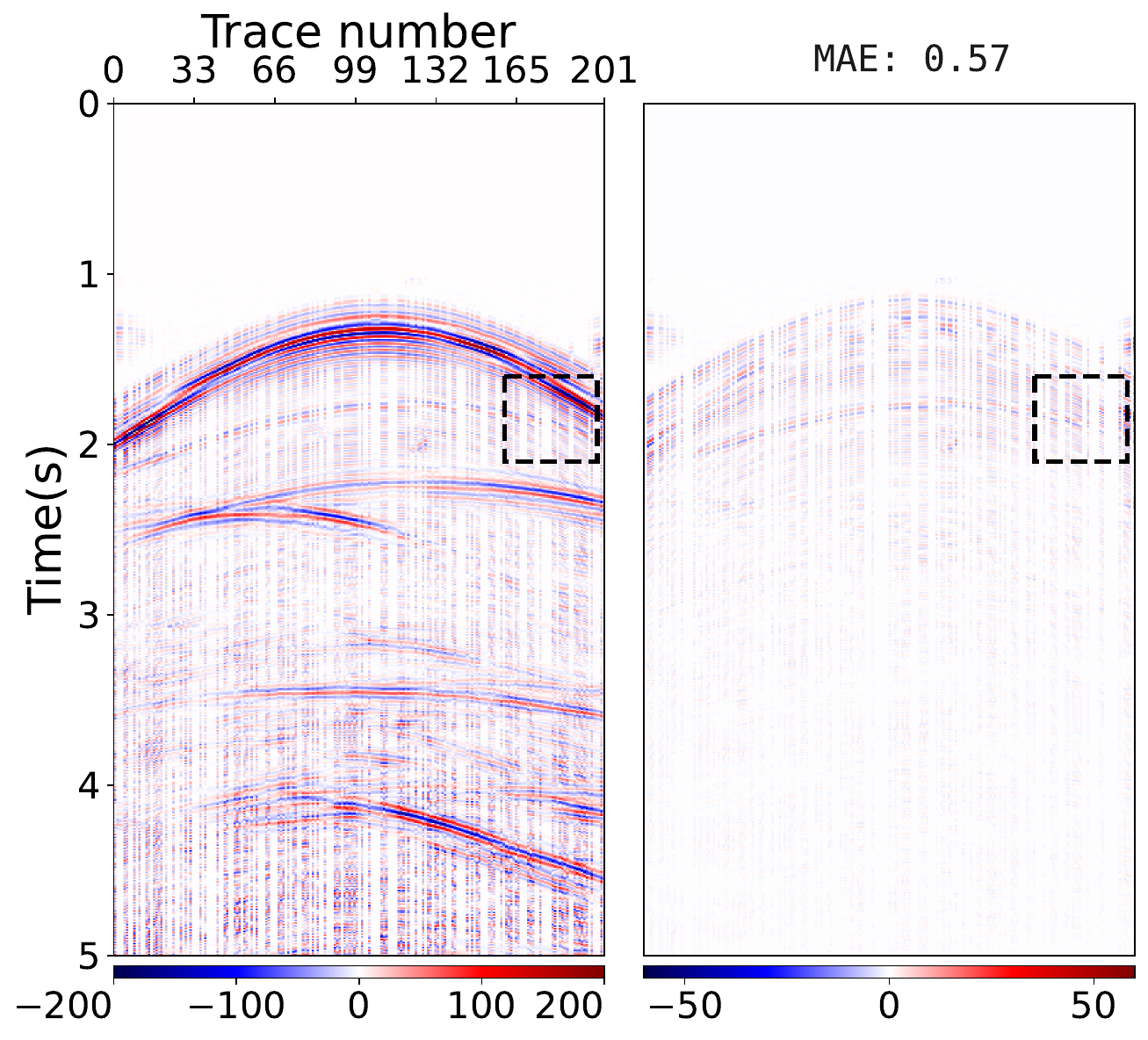}\label{fig:segc3_12_4_Curvelet}}
	\hspace{0mm}
	\subfloat[POCSNet]{\includegraphics[height=1.5in]{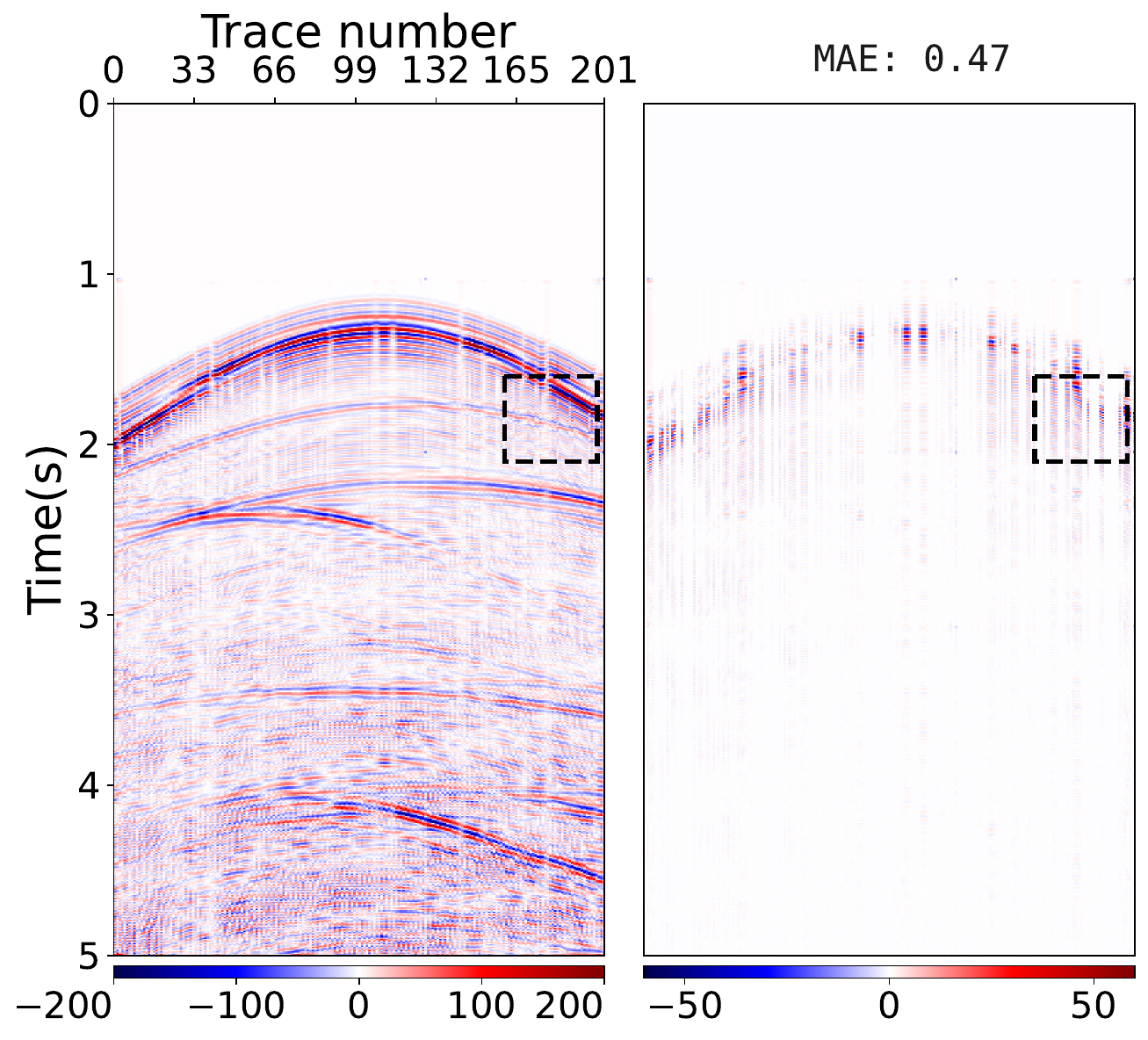}\label{fig:segc3_12_4_POCSNet}}
	
	\hspace{0mm}
	\subfloat[Missing]{\includegraphics[height=1.5in]{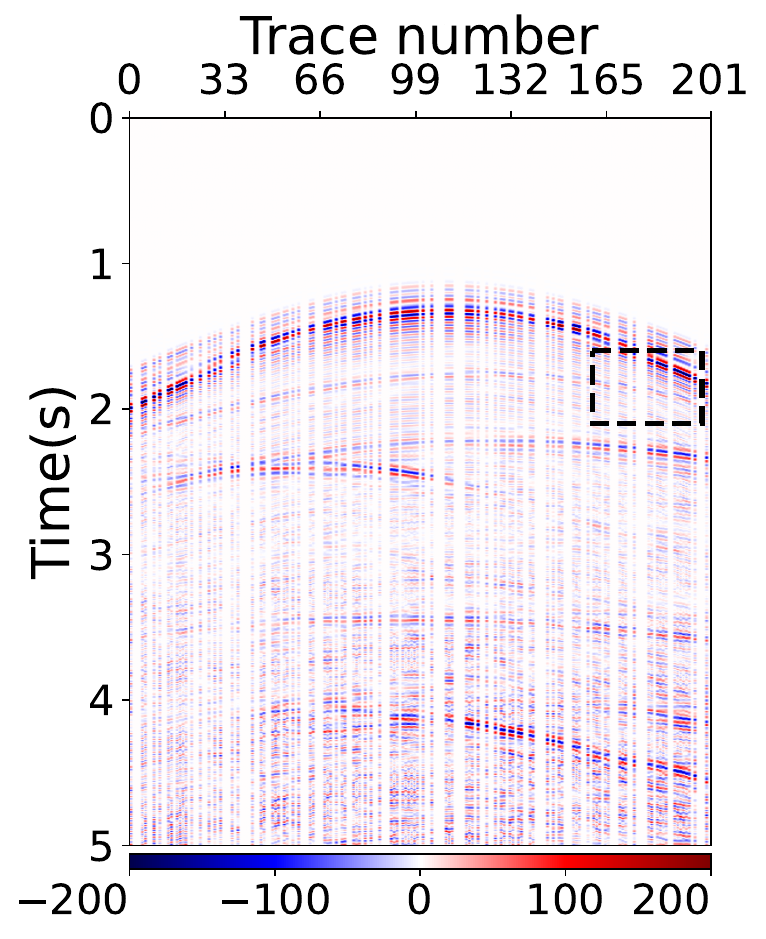}\label{fig:segc3_12_4_Miss}}
	\hspace{0mm}
	\subfloat[SPNet]{\includegraphics[height=1.5in]{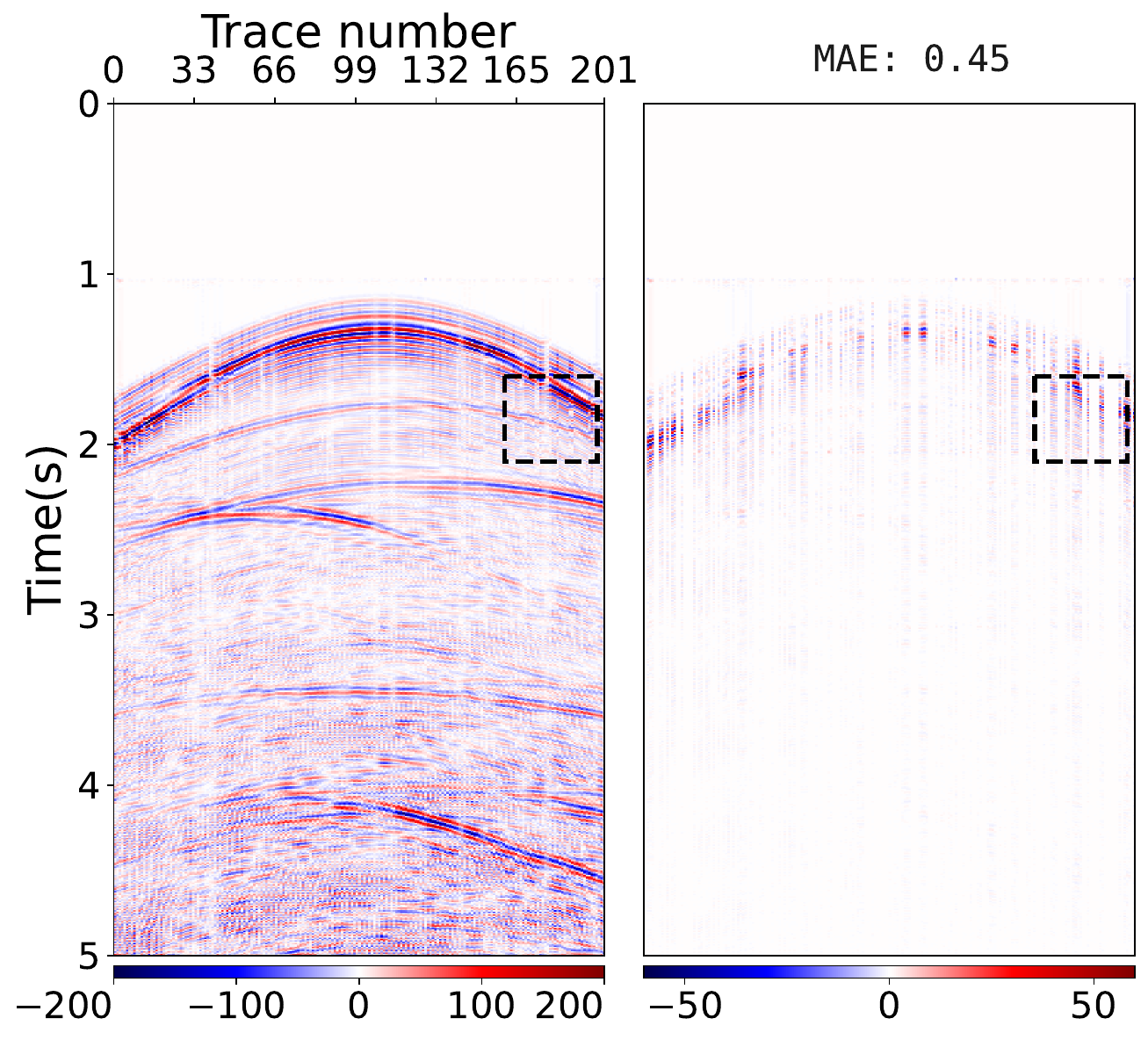}\label{fig:segc3_12_4_SPNet}}
	\hspace{0mm}
	\subfloat[D3UNet]{\includegraphics[height=1.5in]{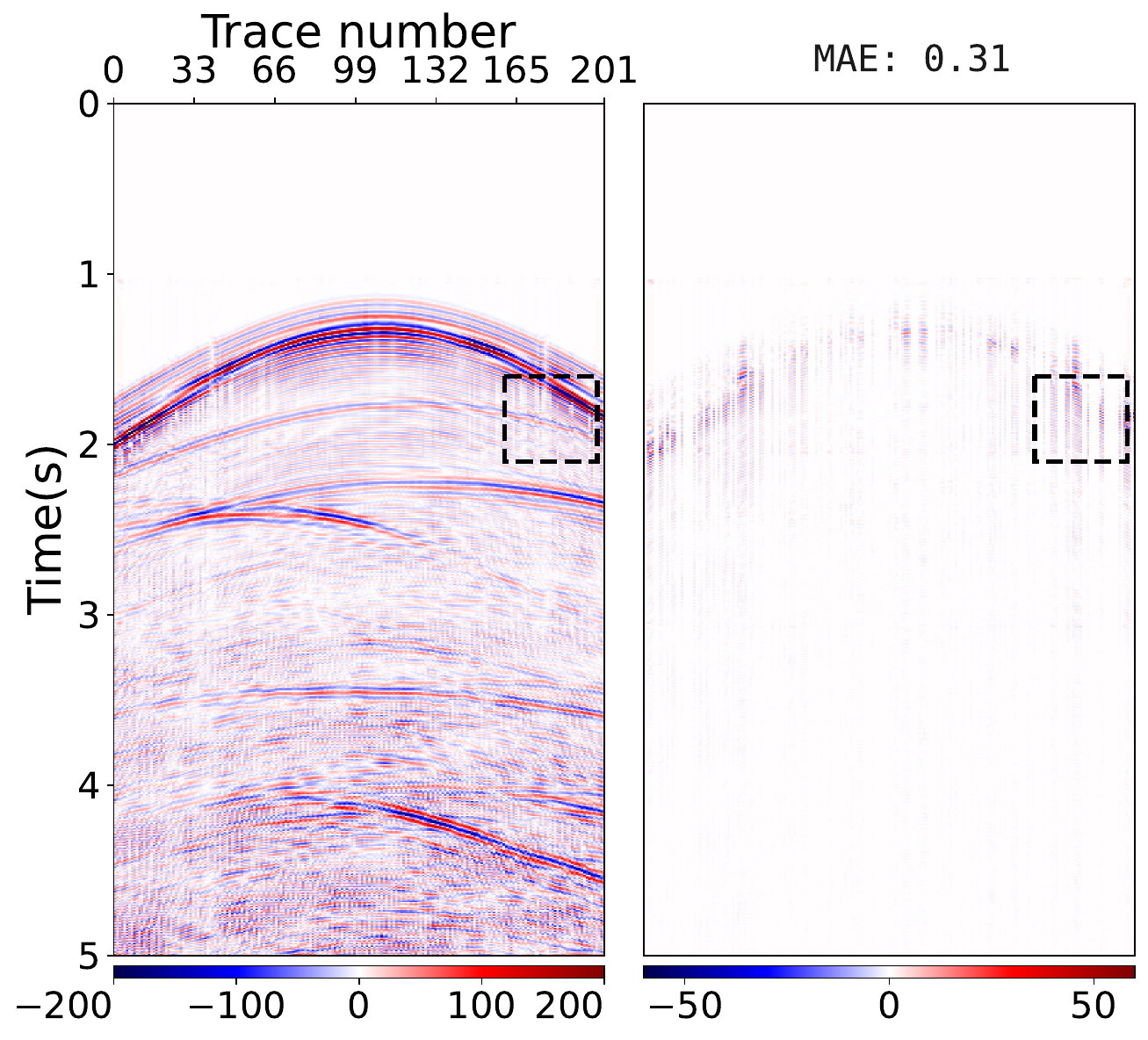}\label{fig:segc3_12_4_D3UNet}}
	\hspace{0mm}
	\subfloat[Ours]{\includegraphics[height=1.5in]{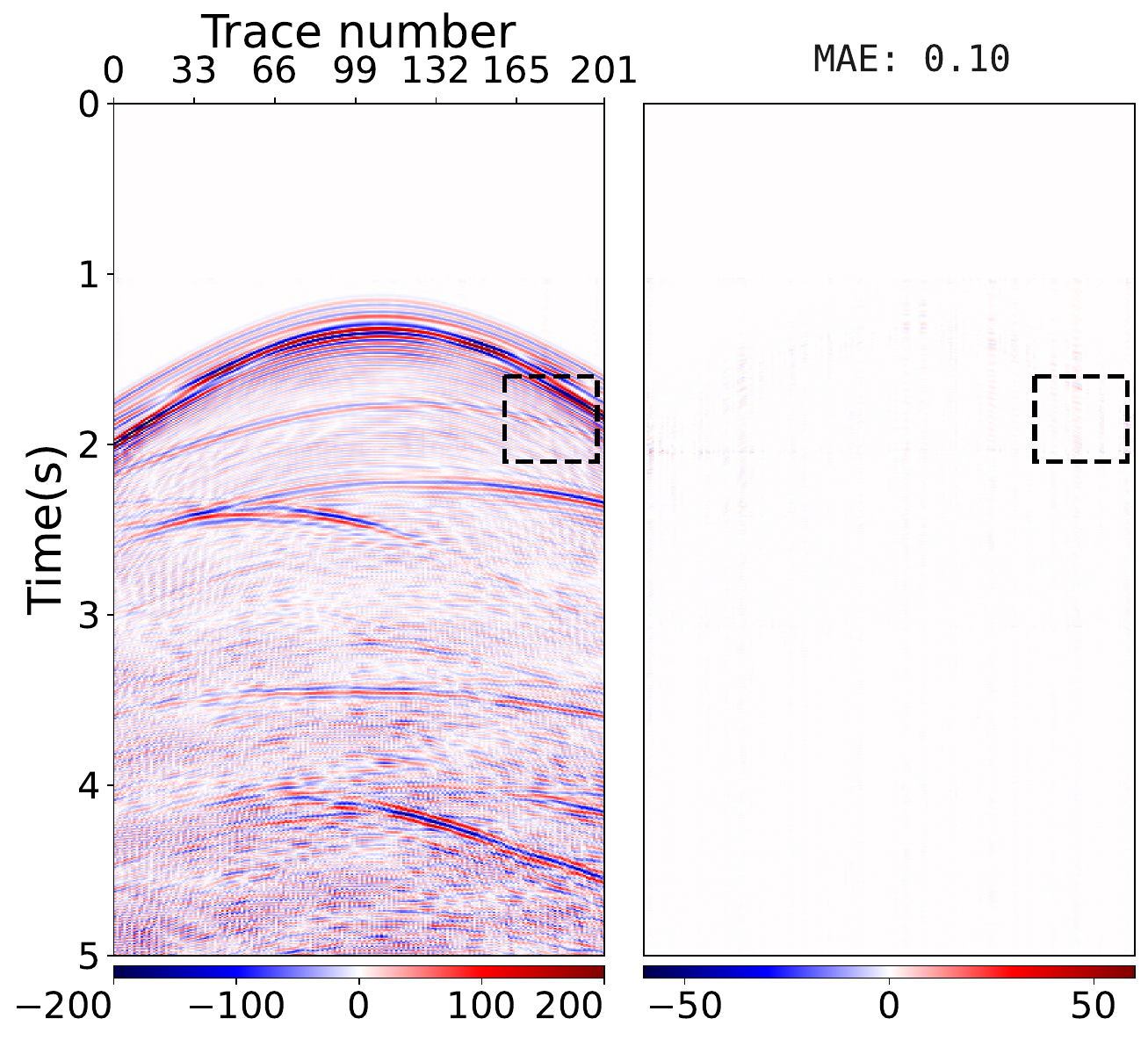}\label{fig:segc3_12_4_Ours}}
	\par
	
	\caption{Qualitative comparison of reconstruction results on the SEG C3 dataset under 50\% random trace removal (shot 12\_4). We restore the seismic data to its original amplitude range.}
	\label{fig:comparison_segc3_12_4}
\end{figure*}

\begin{figure*}[t]
	\centering
	\hspace{-3mm}
	\subfloat[Reference]{\includegraphics[height=1.5in]{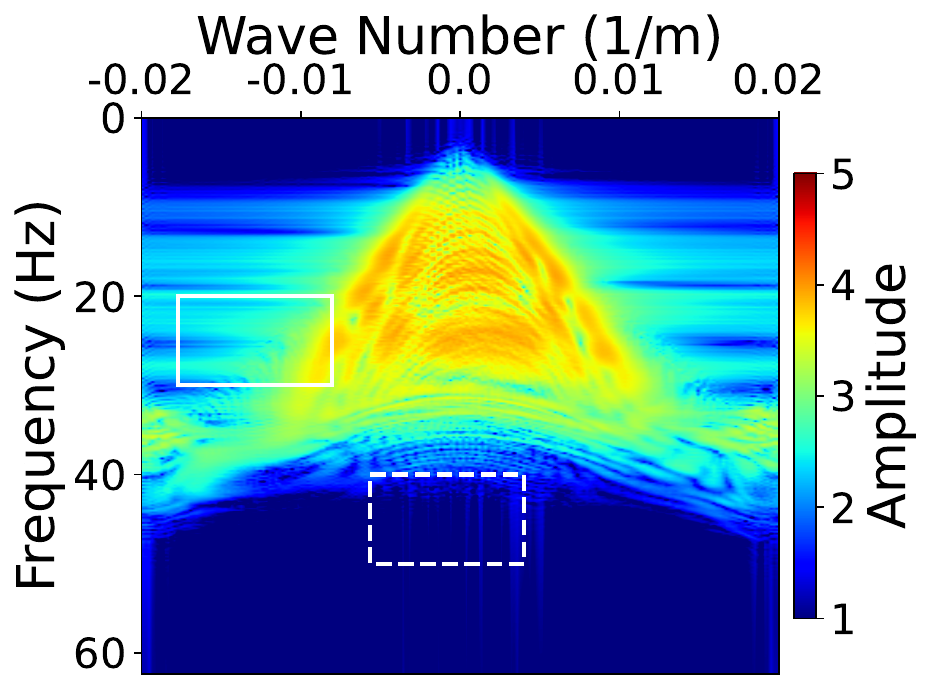}\label{fig:fk_segc3_12_4_Ori}}
	\hspace{0mm}
	\subfloat[Curvelet]{\includegraphics[height=1.5in]{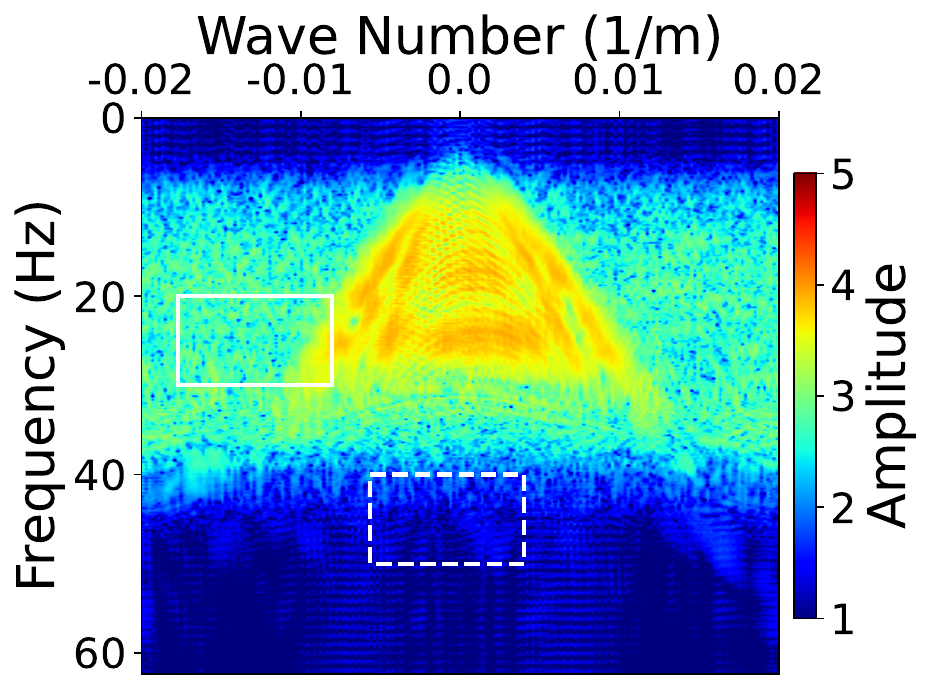}\label{fig:fk_segc3_12_4_Curvelet}}
	\hspace{0mm}
	\subfloat[POCSNet]{\includegraphics[height=1.5in]{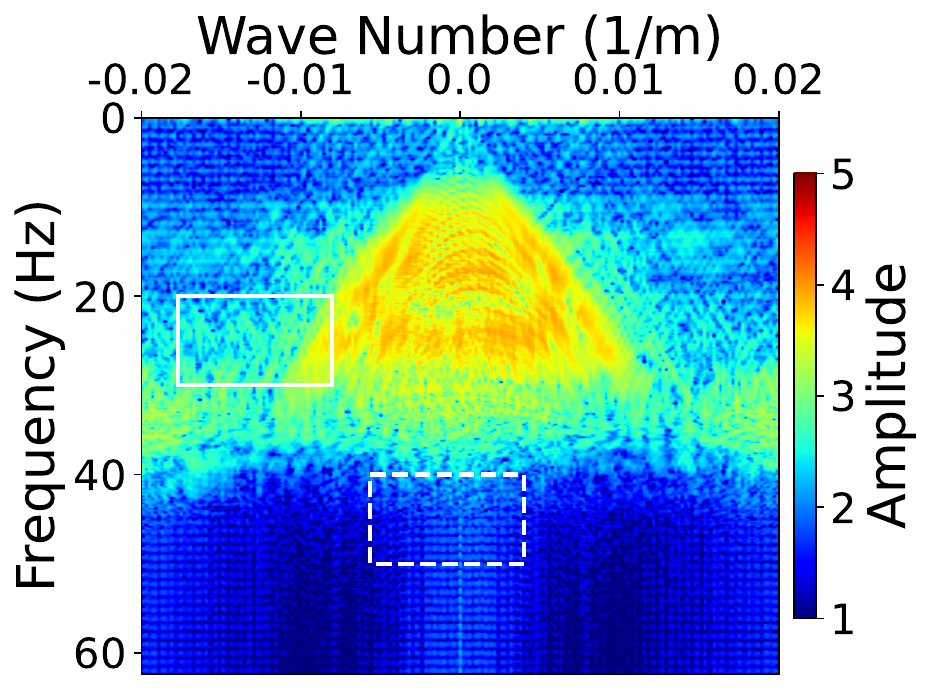}\label{fig:fk_segc3_12_4_POCSNet}}
	
	\hspace{0mm}
	\subfloat[Missing]{\includegraphics[height=1.5in]{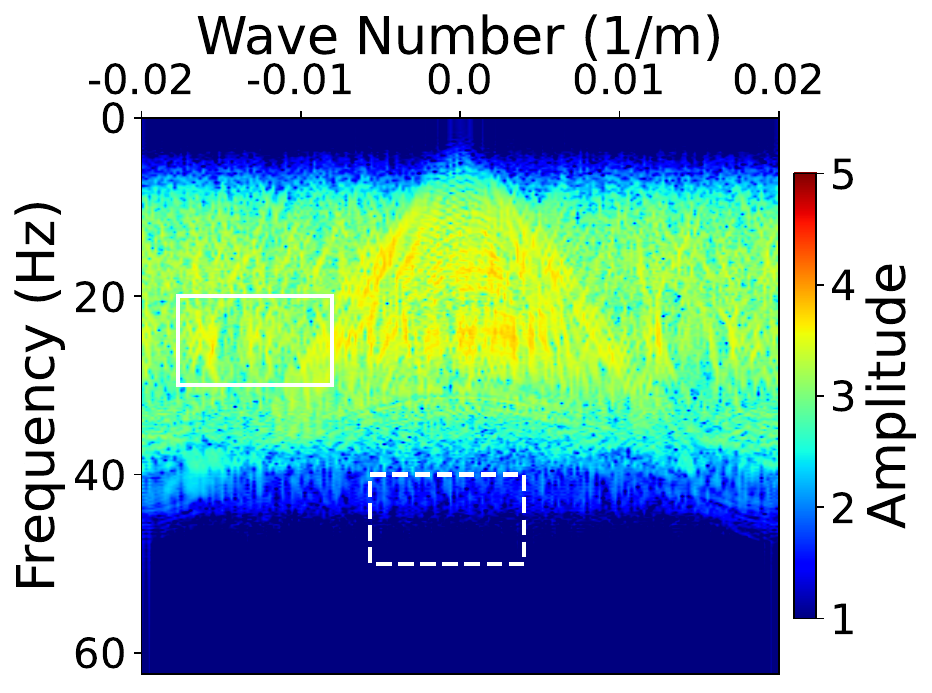}\label{fig:fk_segc3_12_4_Miss}}
	\hspace{0mm}
	\subfloat[SPNet]{\includegraphics[height=1.5in]{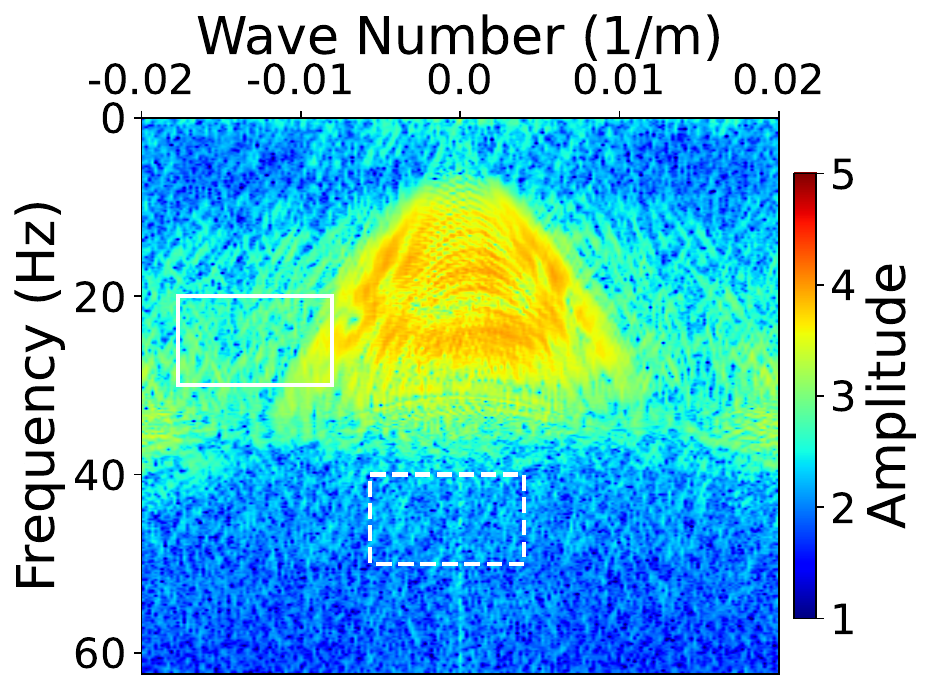}\label{fig:fk_segc3_12_4_SPNet}}
	\hspace{0mm}
	\subfloat[D3UNet]{\includegraphics[height=1.5in]{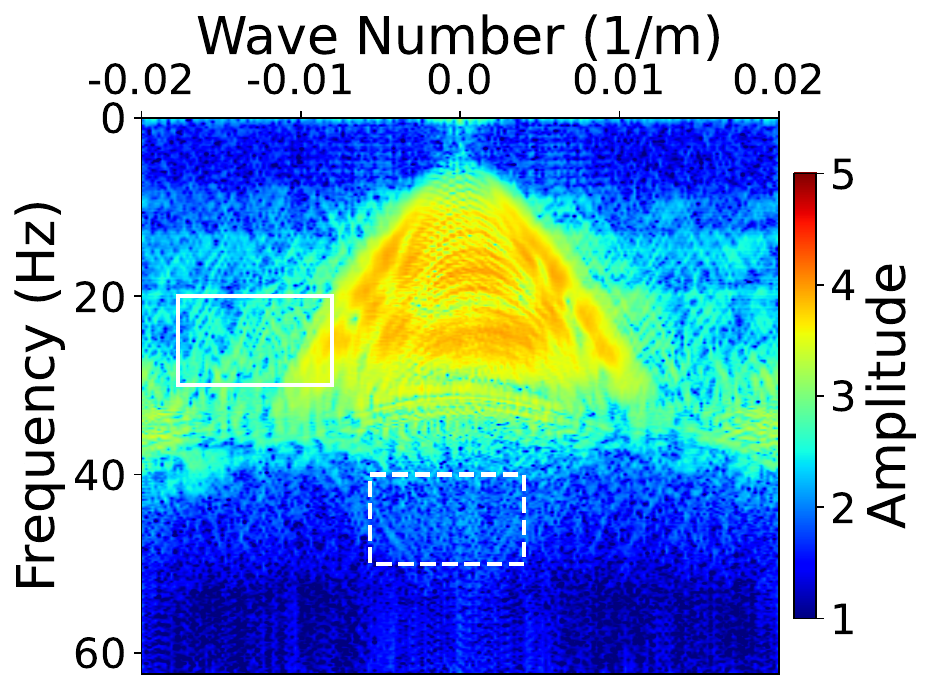}\label{fig:fk_segc3_12_4_D3UNet}}
	\hspace{0mm}
	\subfloat[Ours]{\includegraphics[height=1.5in]{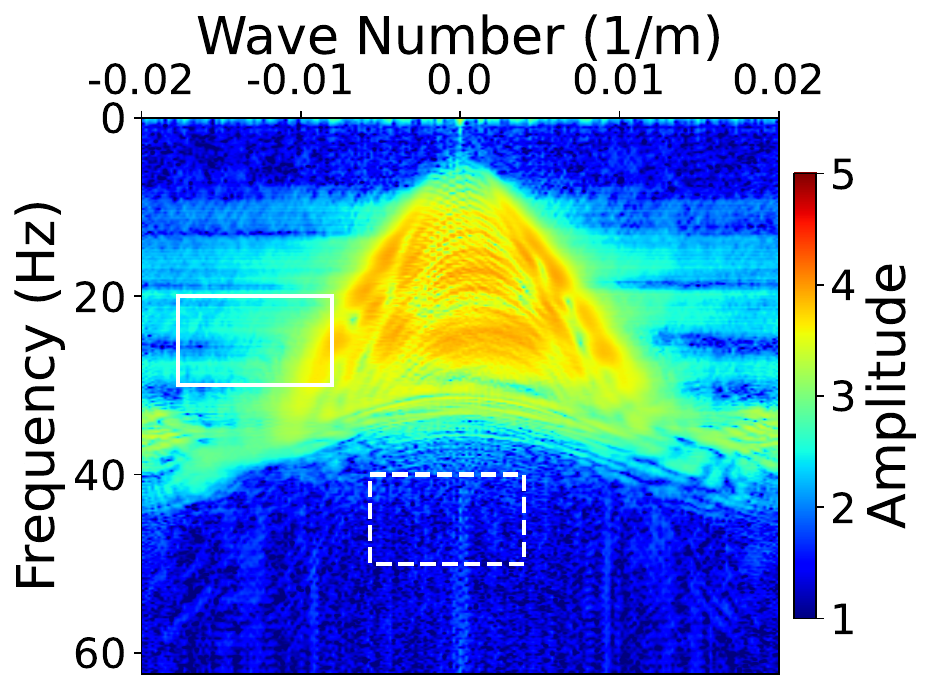}\label{fig:fk_segc3_12_4_Ours}}
	\par
	
	\caption{Qualitative comparison of F-K spectrum reconstruction results on the SEG C3 dataset under 50\% random trace removal (shot 12\_4). We restore the seismic data to its original amplitude range.}
	\label{fig:fk_comparison_segc3_12_4}
\end{figure*}

\begin{figure*}[t]
	\centering
	\hspace{-3mm}
	\subfloat[Reference]{\includegraphics[height=1.5in]{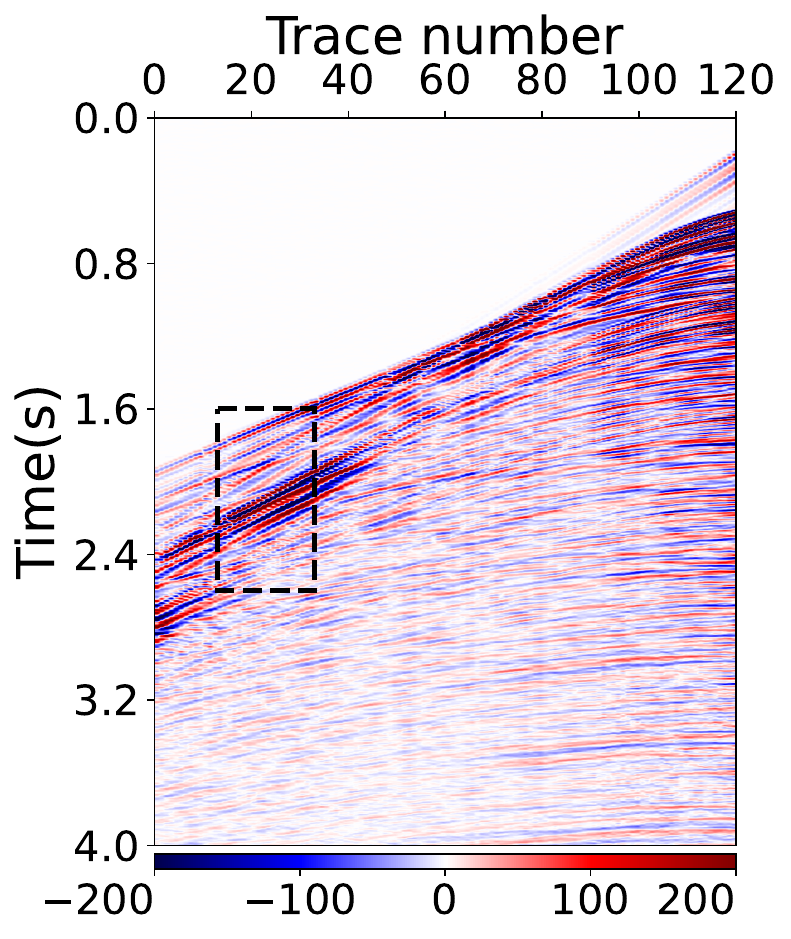}\label{fig:mavo_697_Ori}}
	\hspace{0mm}
	\subfloat[Curvelet]{\includegraphics[height=1.5in]{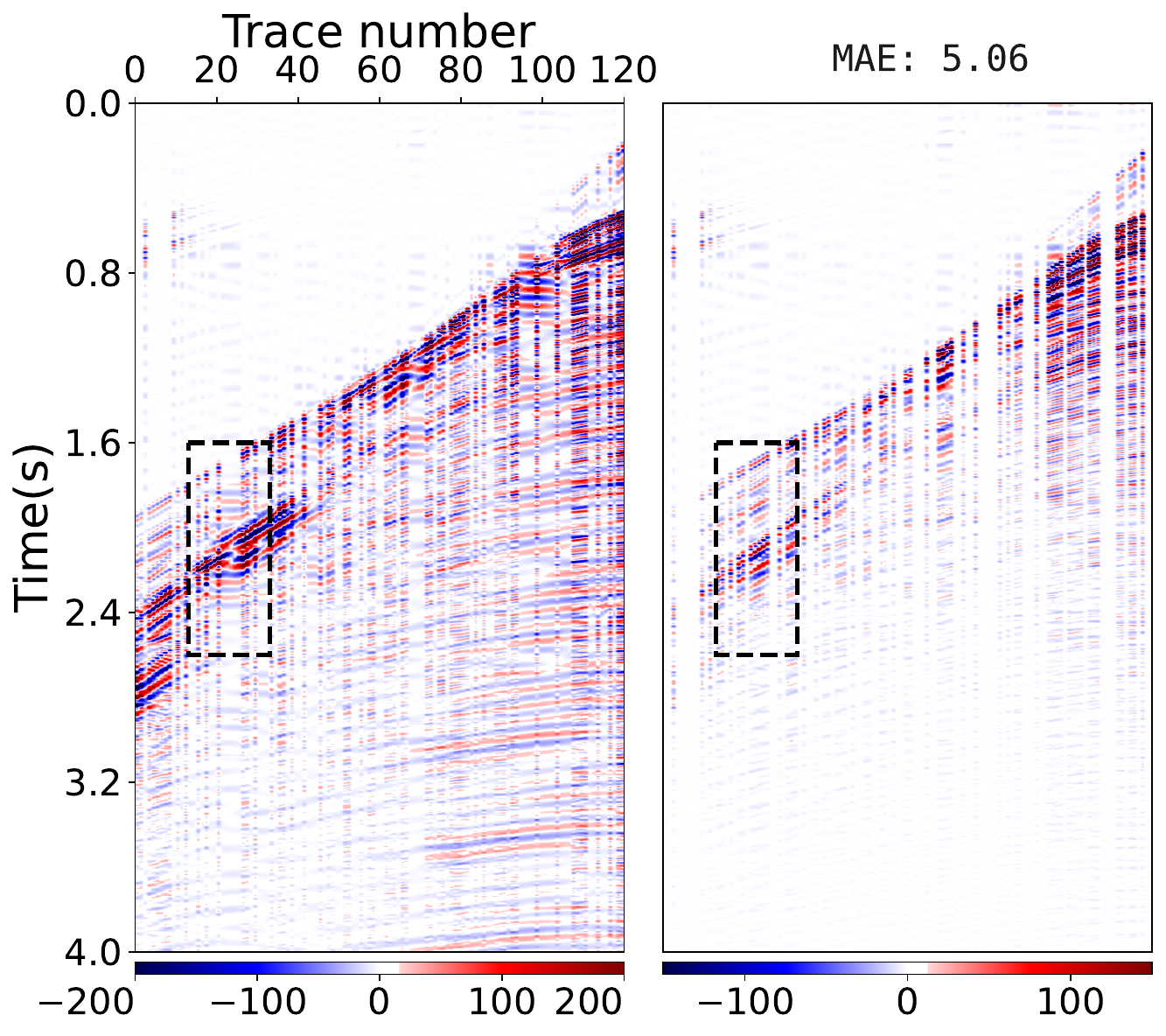}\label{fig:mavo_697_Curvelet}}
	\hspace{0mm}
	\subfloat[POCSNet]{\includegraphics[height=1.5in]{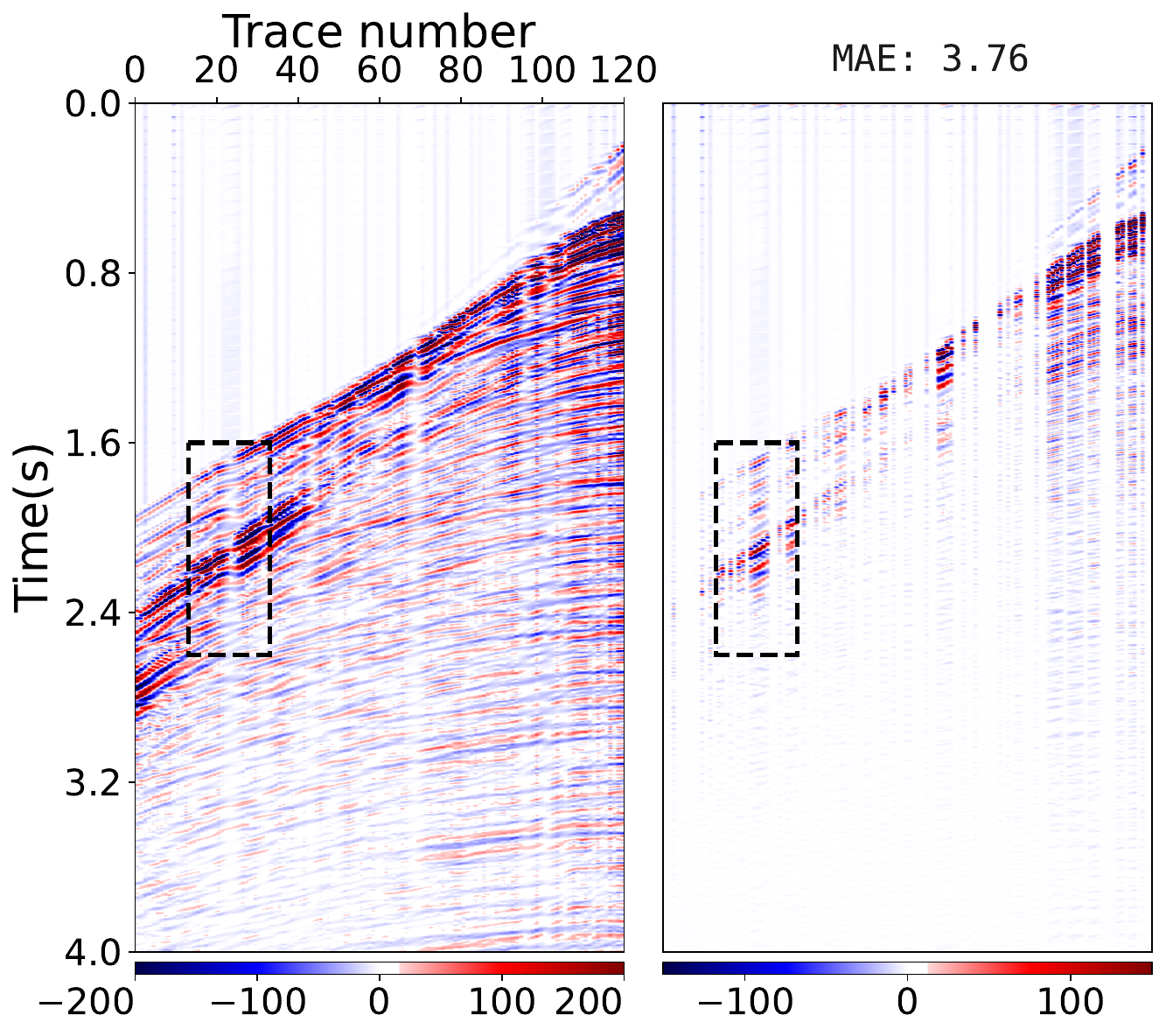}\label{fig:mavo_697_POCSNet}}
	
	\hspace{0mm}
	\subfloat[Missing]{\includegraphics[height=1.5in]{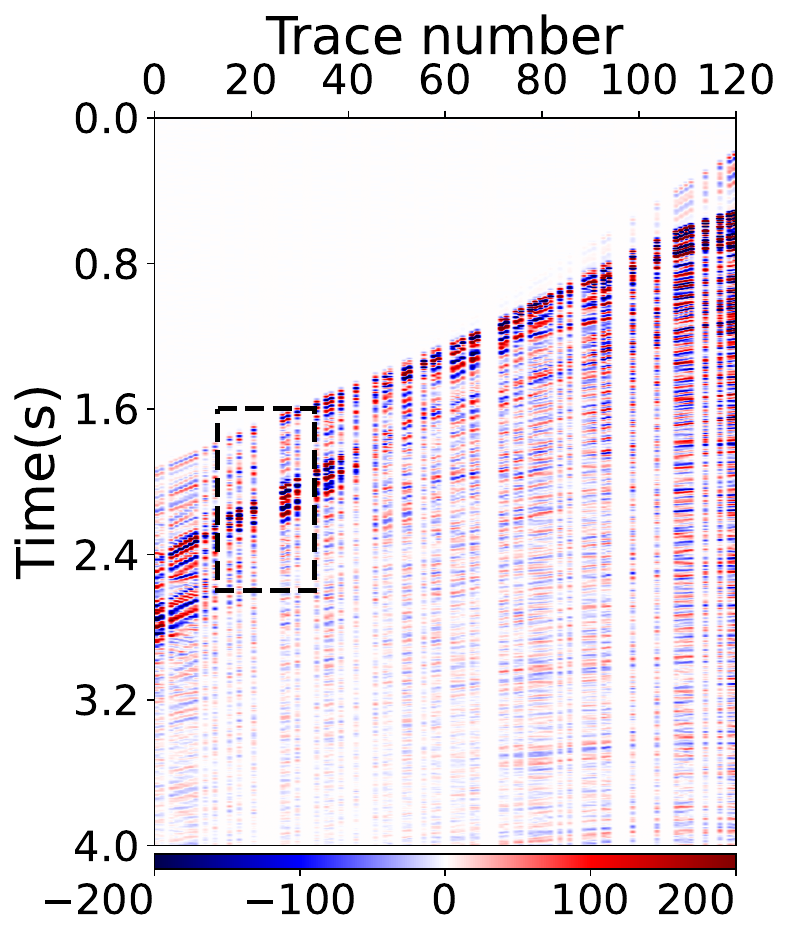}\label{fig:mavo_697_Miss}}
	\hspace{0mm}
	\subfloat[SPNet]{\includegraphics[height=1.5in]{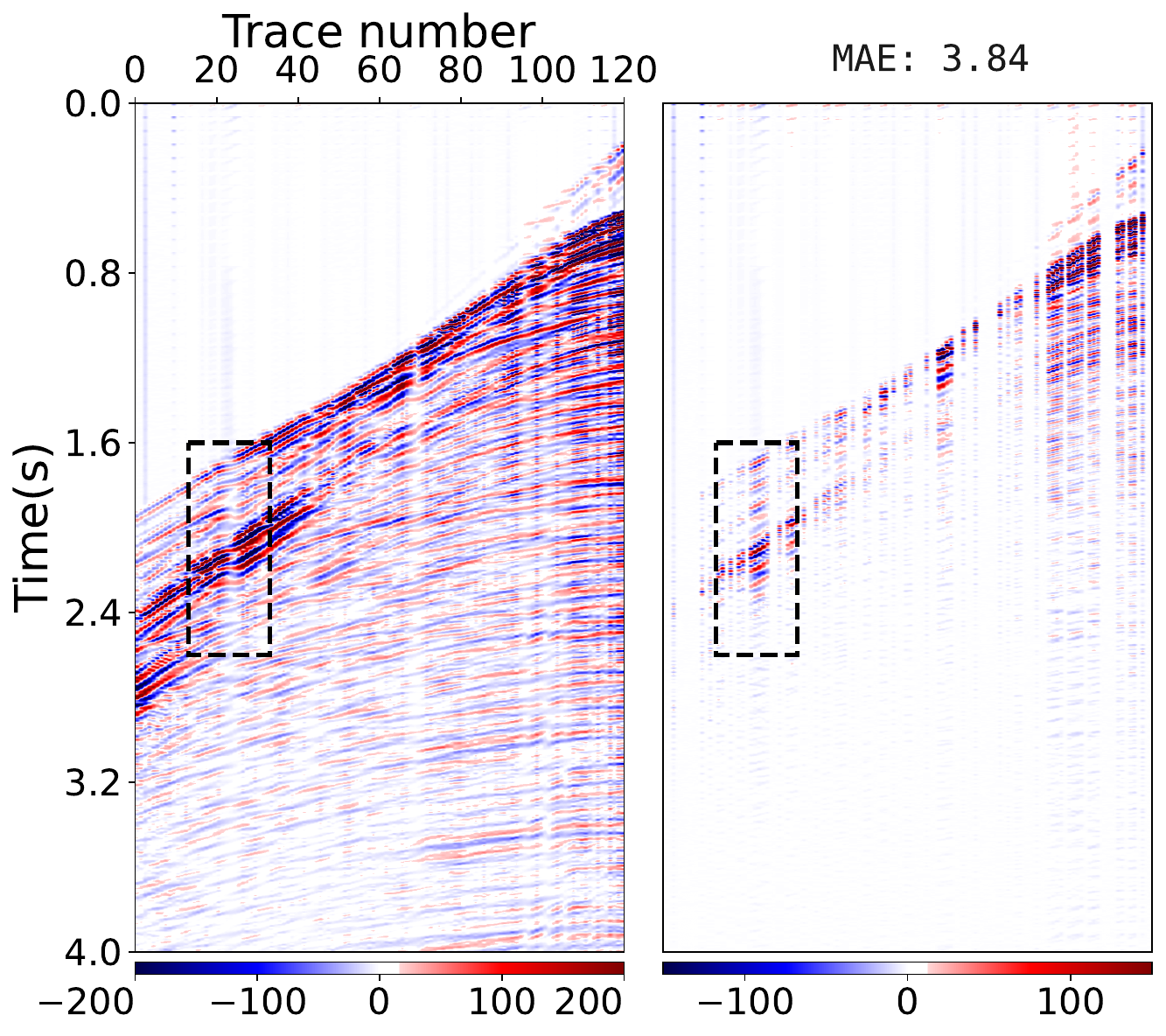}\label{fig:mavo_697_SPNet}}
	\hspace{0mm}
	\subfloat[D3UNet]{\includegraphics[height=1.5in]{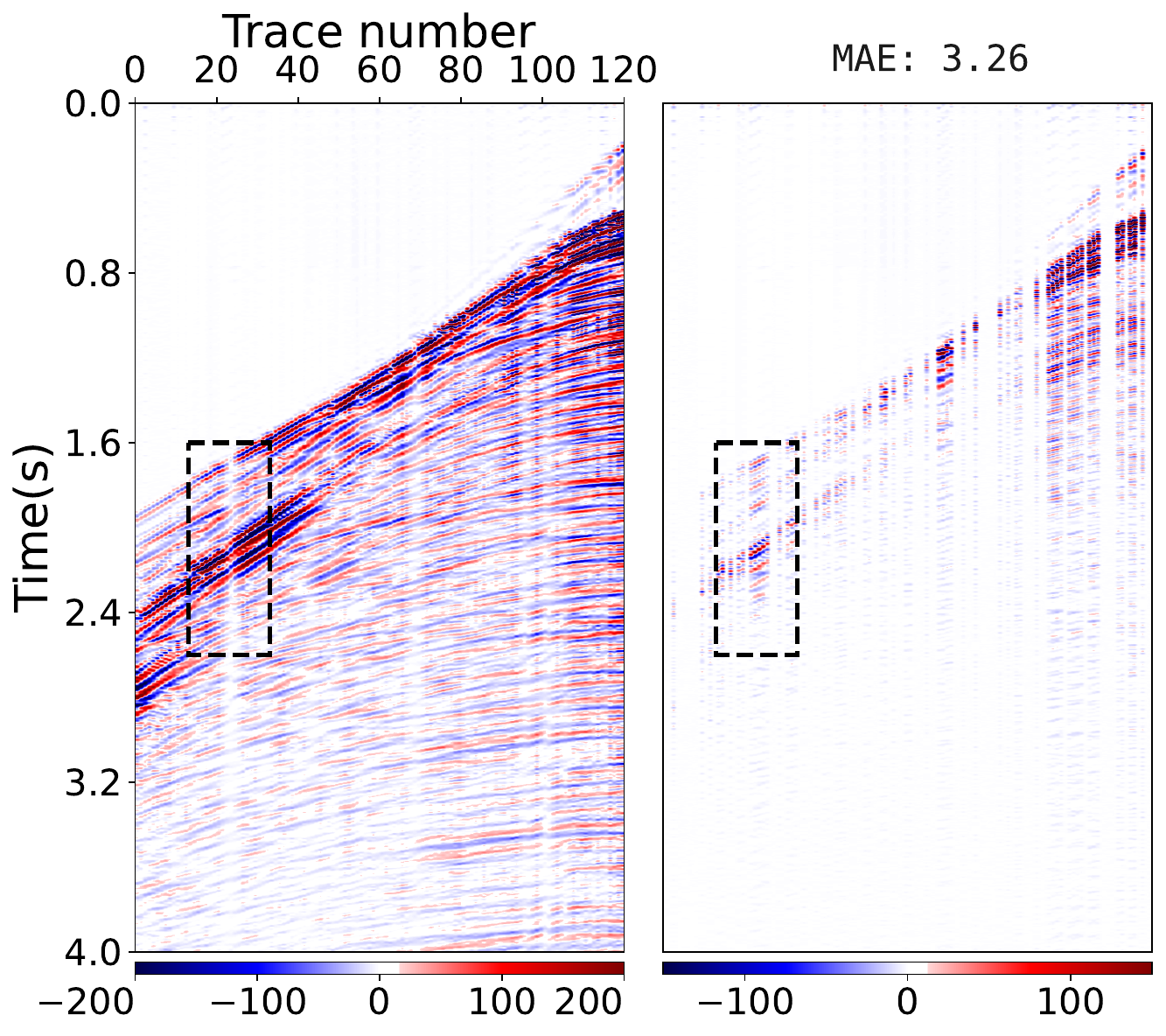}\label{fig:mavo_697_D3UNet}}
	\hspace{0mm}
	\subfloat[Ours]{\includegraphics[height=1.5in]{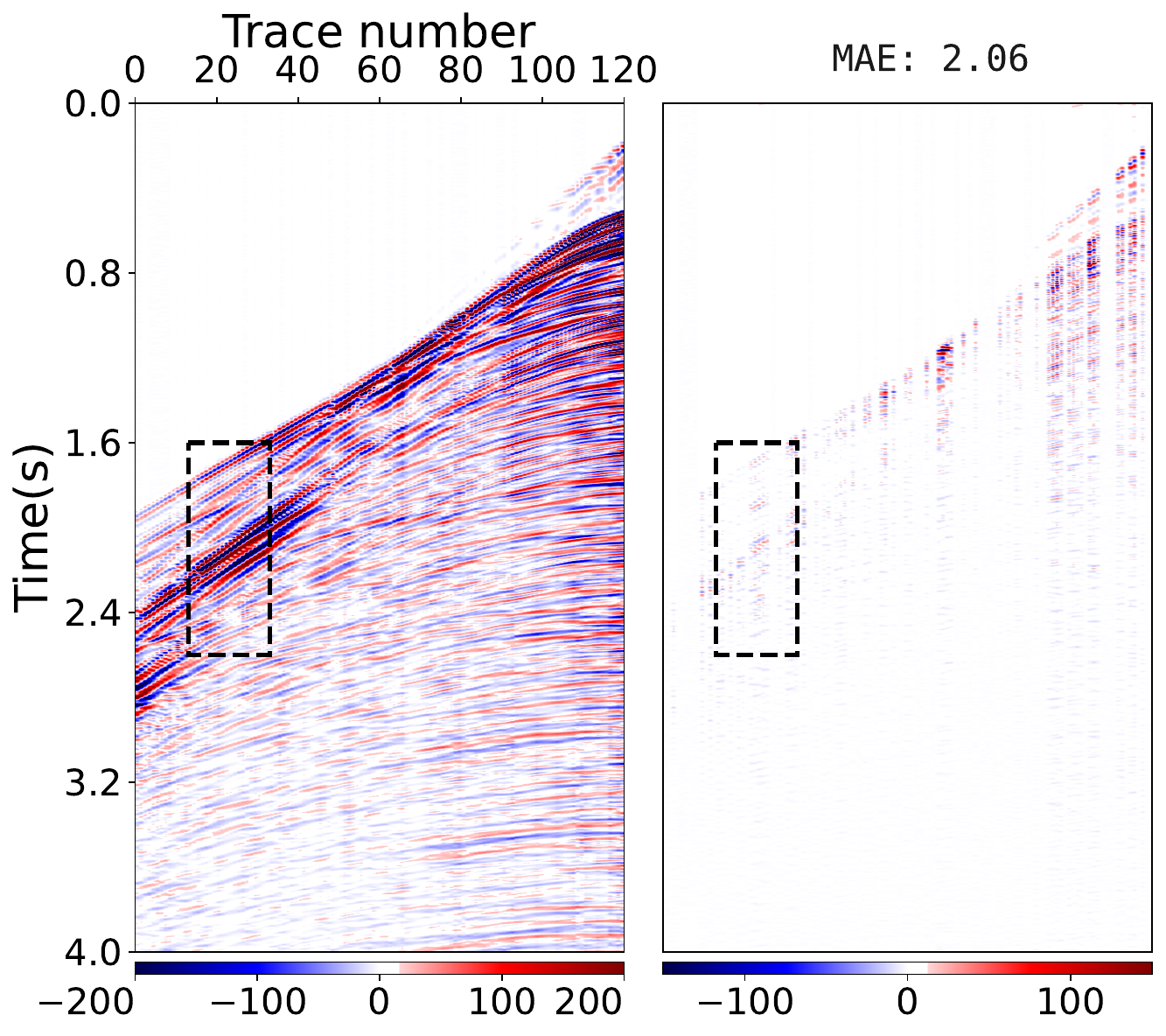}\label{fig:mavo_697_Ours}}
	\par

	\caption{Qualitative comparison of reconstruction results on the MAVO dataset under 50\% trace removal. (a) Random undersampling (shot 180). (b) Consecutive large gaps (shot 697). We restore the seismic data to its original amplitude range. }
	\label{fig:comparison}
\end{figure*}

\begin{figure*}[t]
	\centering
	\hspace{-10mm}
	\subfloat[Reference]{\includegraphics[height=1.3in]{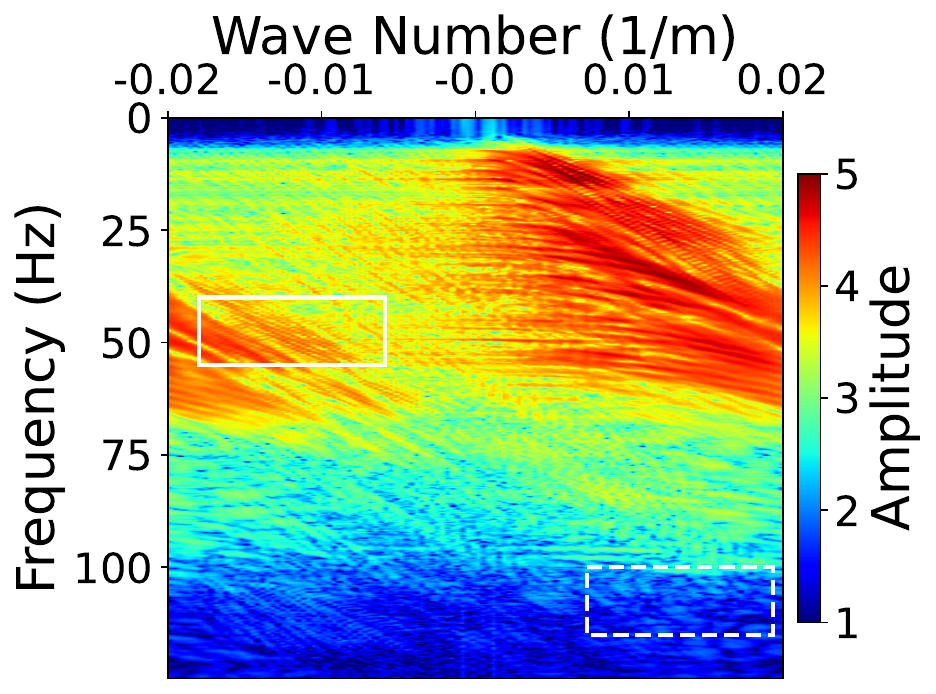}\label{fig:fk_mavo_697_Ori}}
	\hspace{0mm}
	\subfloat[Curvelet]{\includegraphics[height=1.3in]{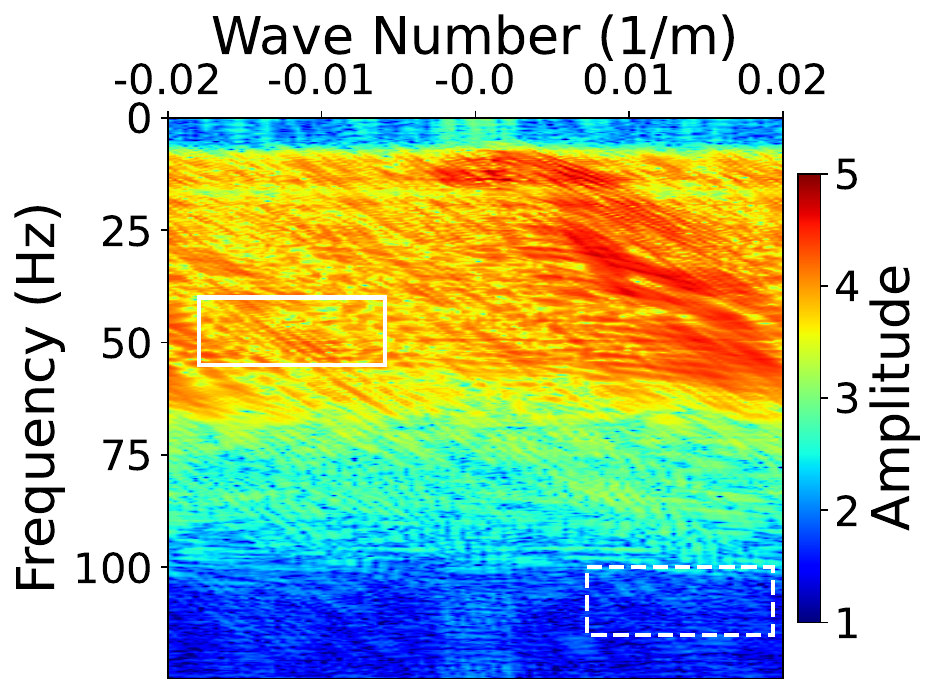}\label{fig:fk_mavo_697_Curvelet}}
	\hspace{0mm}
	\subfloat[POCSNet]{\includegraphics[height=1.3in]{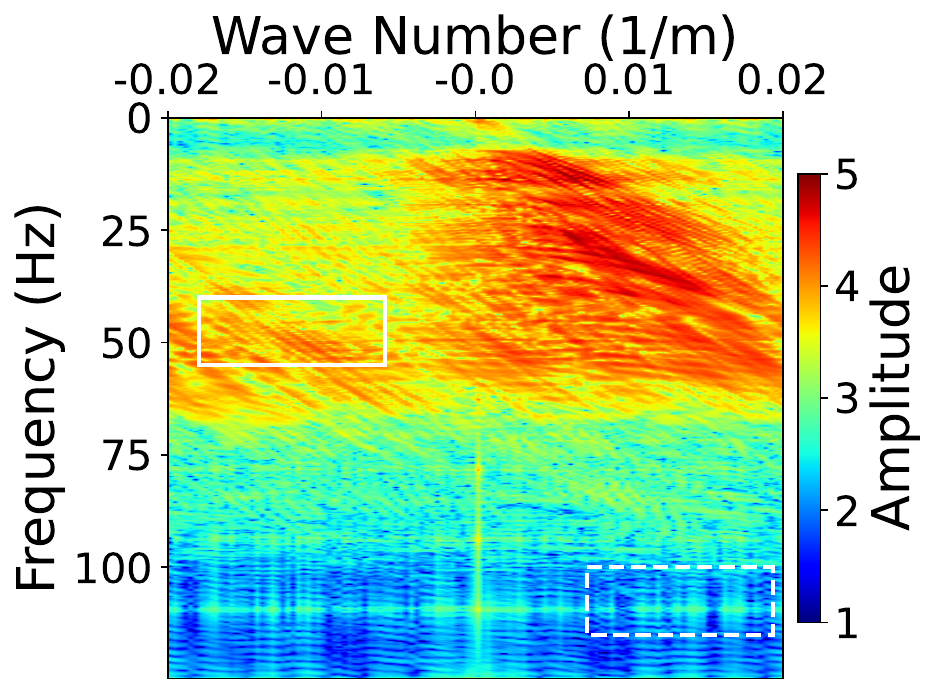}\label{fig:fk_mavo_697_POCSNet}}
	
	\hspace{-10mm}
	\subfloat[Missing]{\includegraphics[height=1.3in]{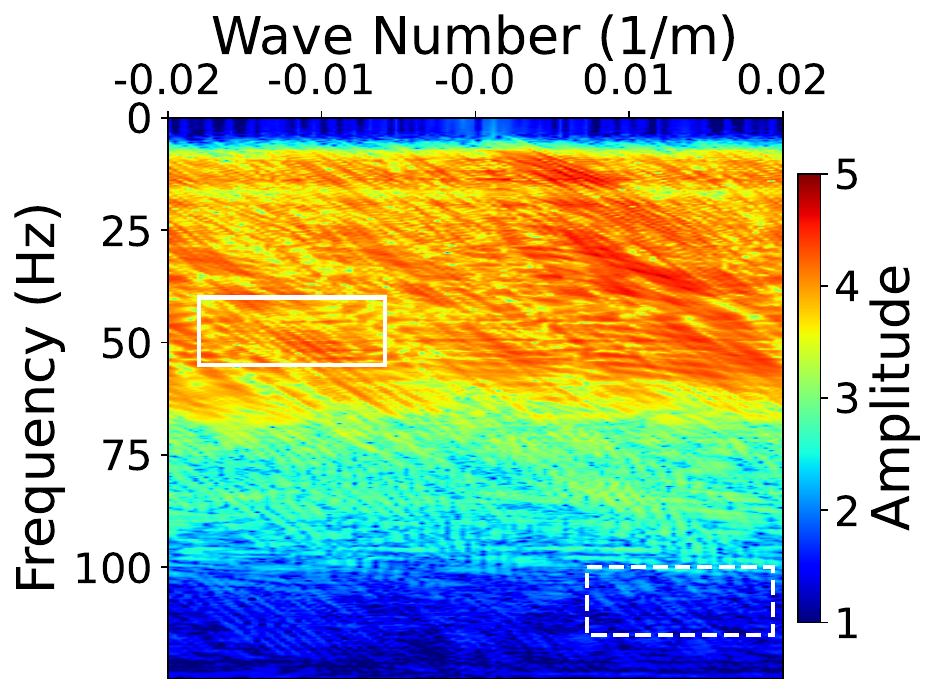}\label{fig:fk_mavo_697_Miss}}
	\hspace{0mm}
	\subfloat[SPNet]{\includegraphics[height=1.3in]{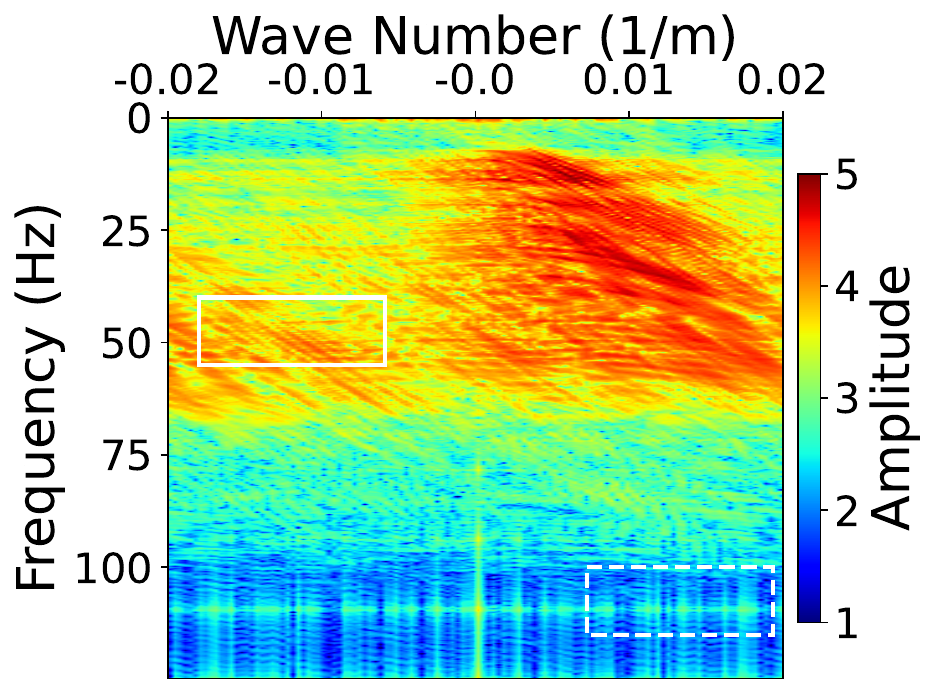}\label{fig:fk_mavo_697_SPNet}}
	\hspace{0mm}
	\subfloat[D3UNet]{\includegraphics[height=1.3in]{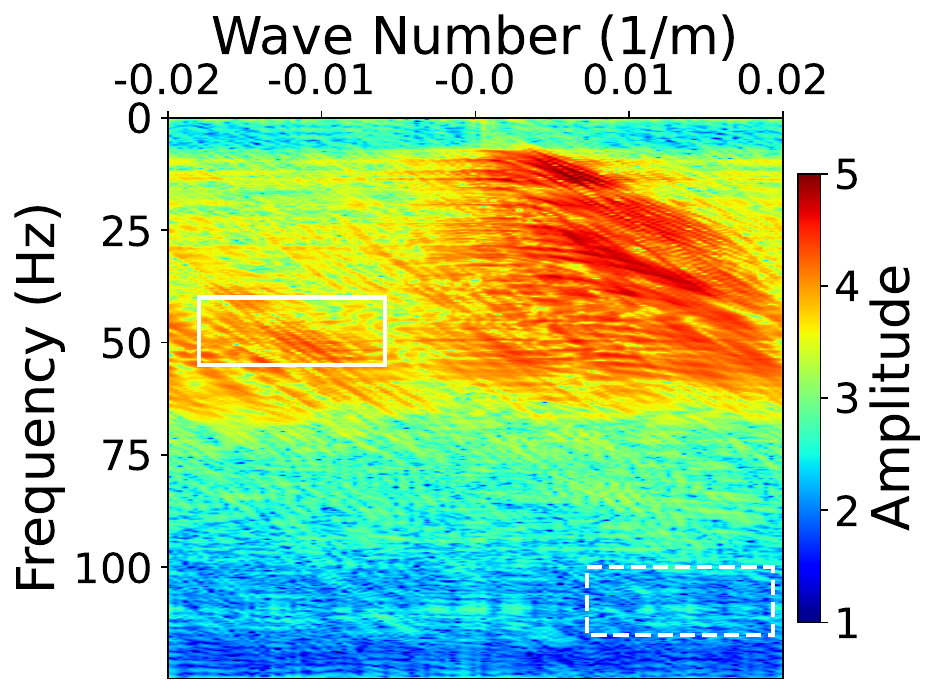}\label{fig:fk_mavo_697_D3UNet}}
	\hspace{0mm}
	\subfloat[Ours]{\includegraphics[height=1.3in]{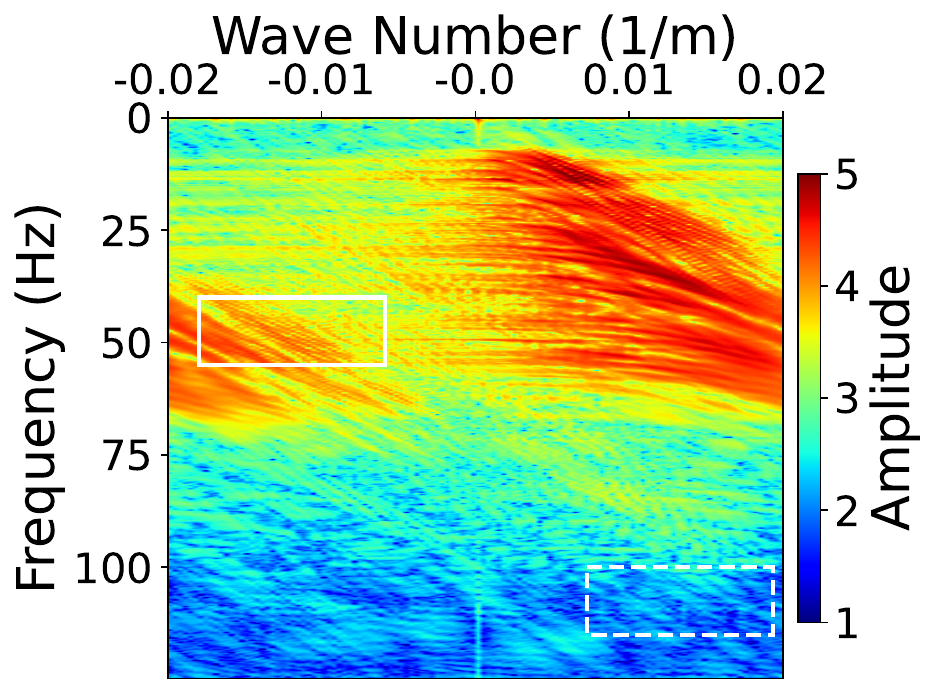}\label{fig:fk_mavo_697_Ours}}
	\par
	
	\caption{Qualitative comparison of F-K spectrum reconstruction results on the MAVO dataset under 50\% trace removal. (a) Random undersampling (shot 180). (b) Consecutive large gaps (shot 697). We restore the seismic data to its original amplitude range.}
	\label{fig:fk_comparison}
\end{figure*}

Relative to equal supervision, the paired full-depth changes for random and consecutive removal are $(\pending,\pending)$~dB on SEG C3 and $(\pending,\pending)$~dB on MAVO; their respective confidence intervals are $(\pending,\pending)$ and $(\pending,\pending)$. The original-recipe SSI-POCS comparison retains the relationship to the preceding method, while the matched fixed-8 comparison controls the loss normalization and reconstruction architecture.

\begin{figure*}[t]
	\centering
	\hspace{-3mm}
	\subfloat[Reference]{\includegraphics[height=1.5in]{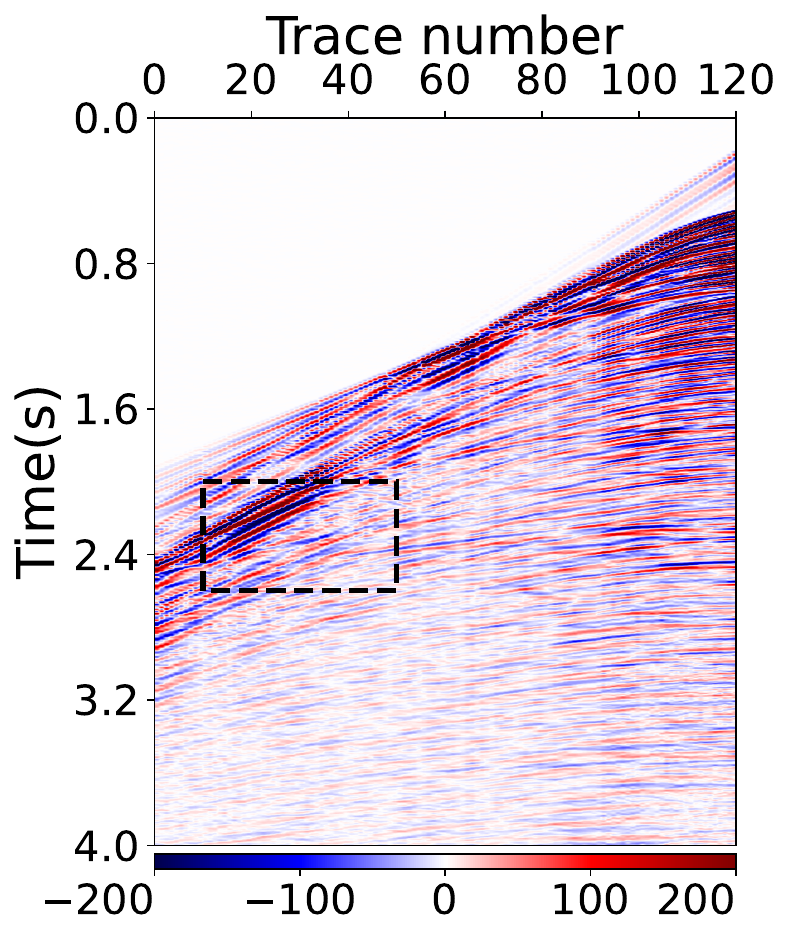}\label{fig:mavo_488_Ori}}
	\hspace{0mm}
	\subfloat[Curvelet]{\includegraphics[height=1.5in]{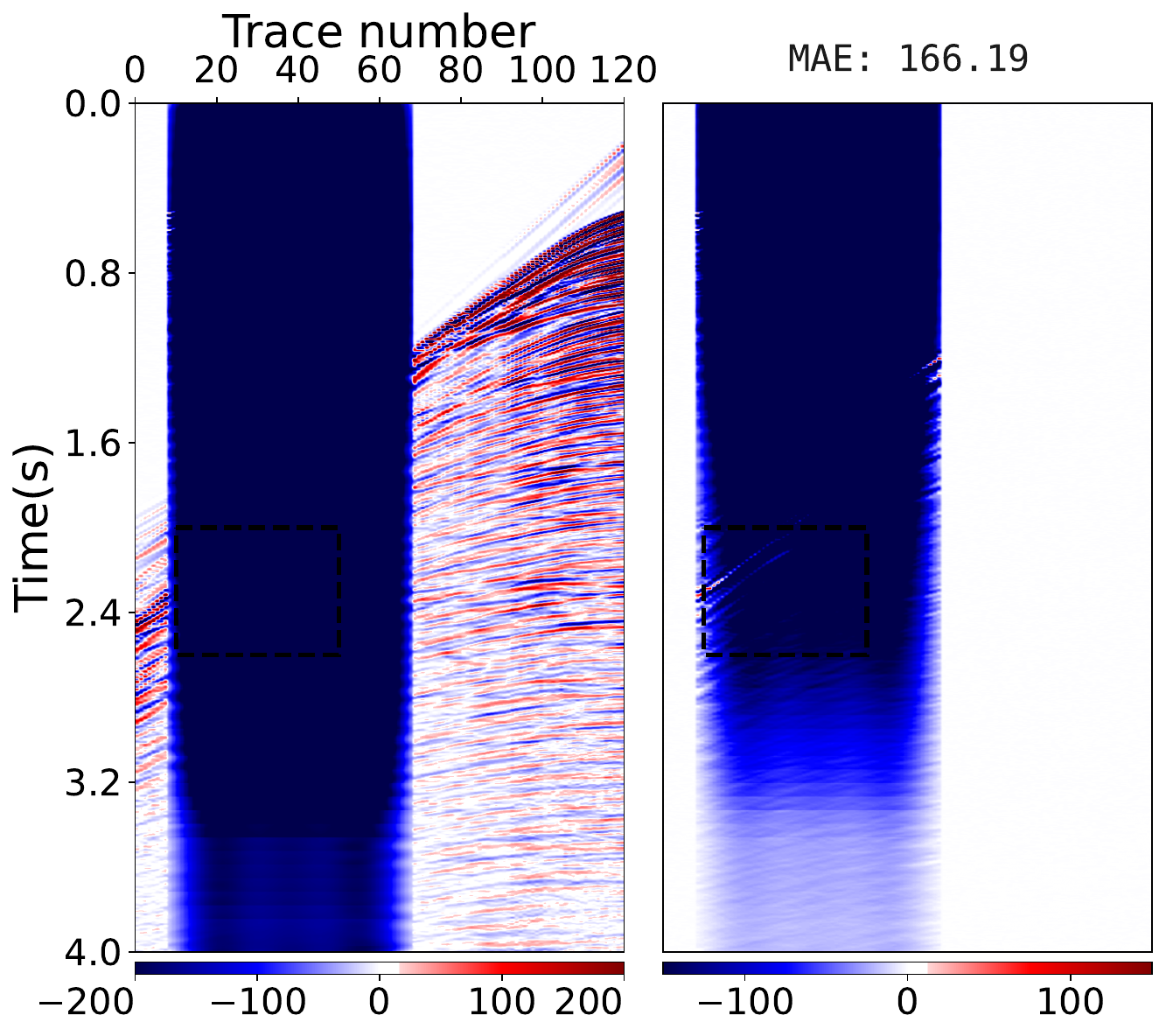}\label{fig:mavo_488_Curvelet}}
	\hspace{0mm}
	\subfloat[POCSNet]{\includegraphics[height=1.5in]{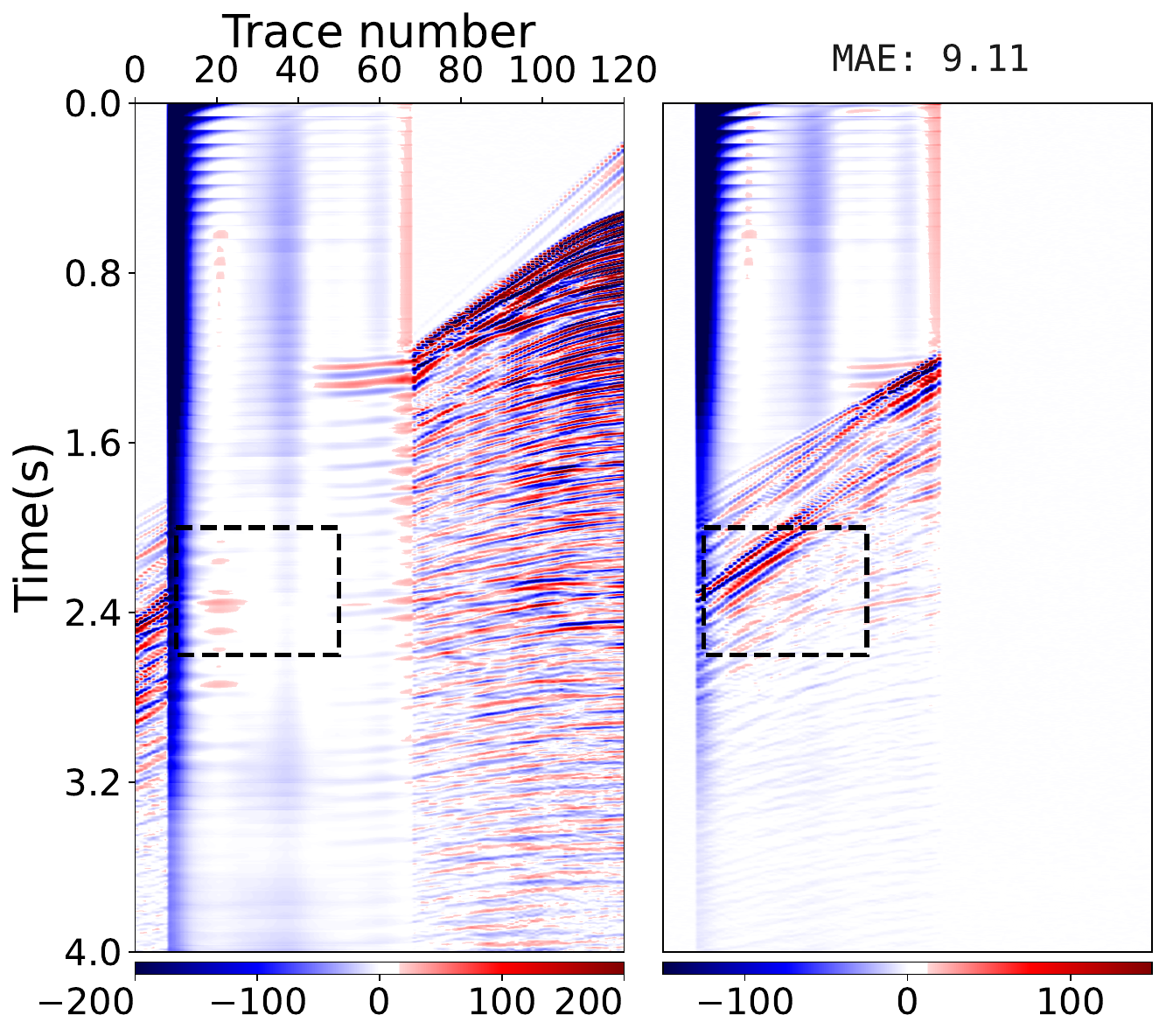}\label{fig:mavo_488_POCSNet}}
	
	\hspace{0mm}
	\subfloat[Missing]{\includegraphics[height=1.5in]{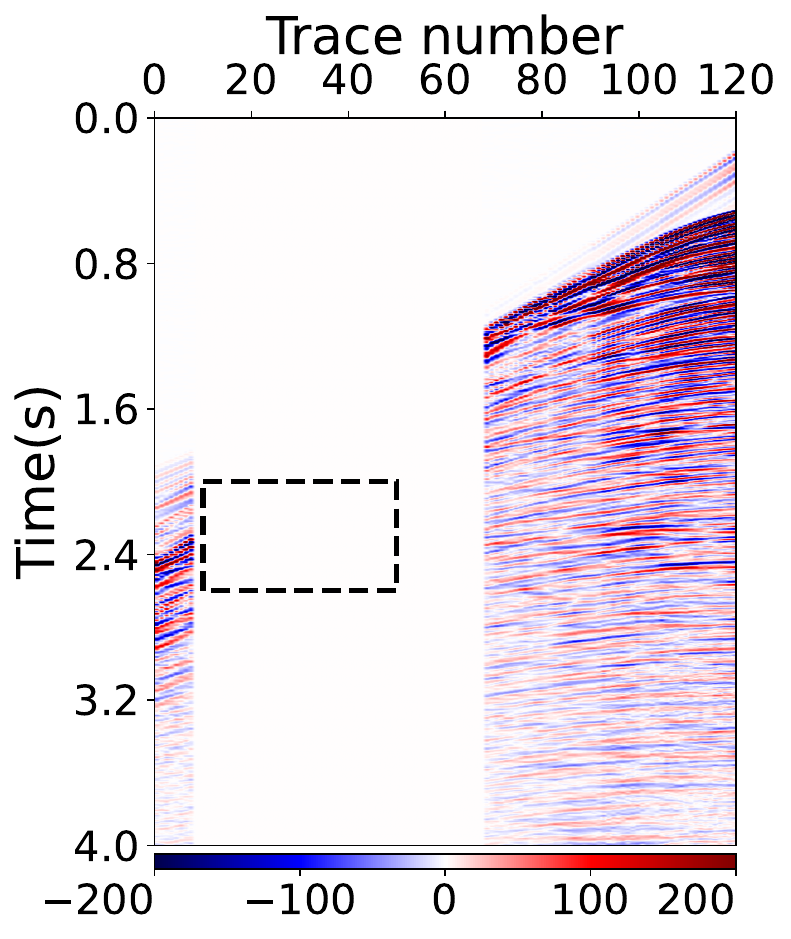}\label{fig:mavo_488_Miss}}
	\hspace{0mm}
	\subfloat[SPNet]{\includegraphics[height=1.5in]{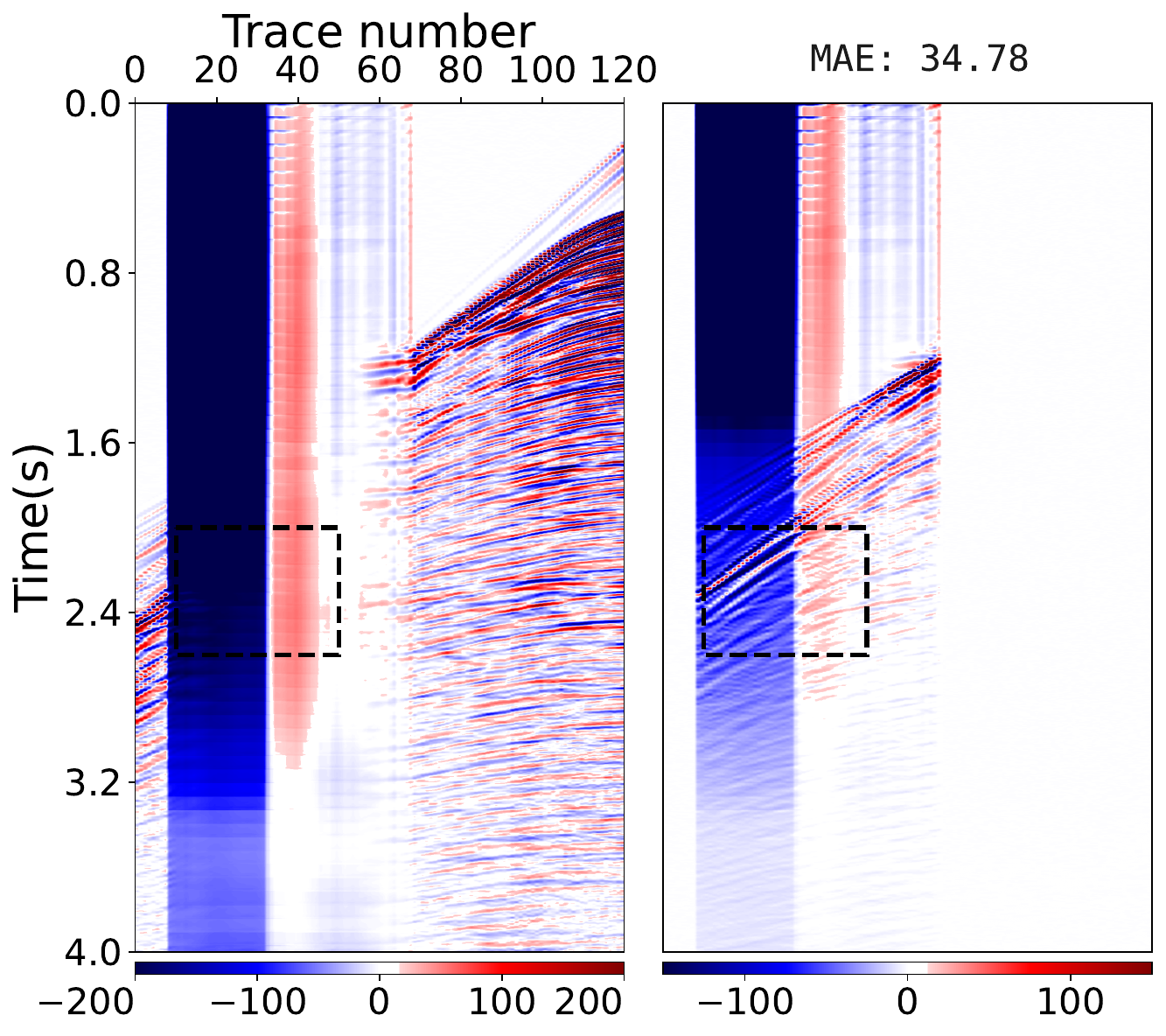}\label{fig:mavo_488_SPNet}}
	\hspace{0mm}
	\subfloat[D3UNet]{\includegraphics[height=1.5in]{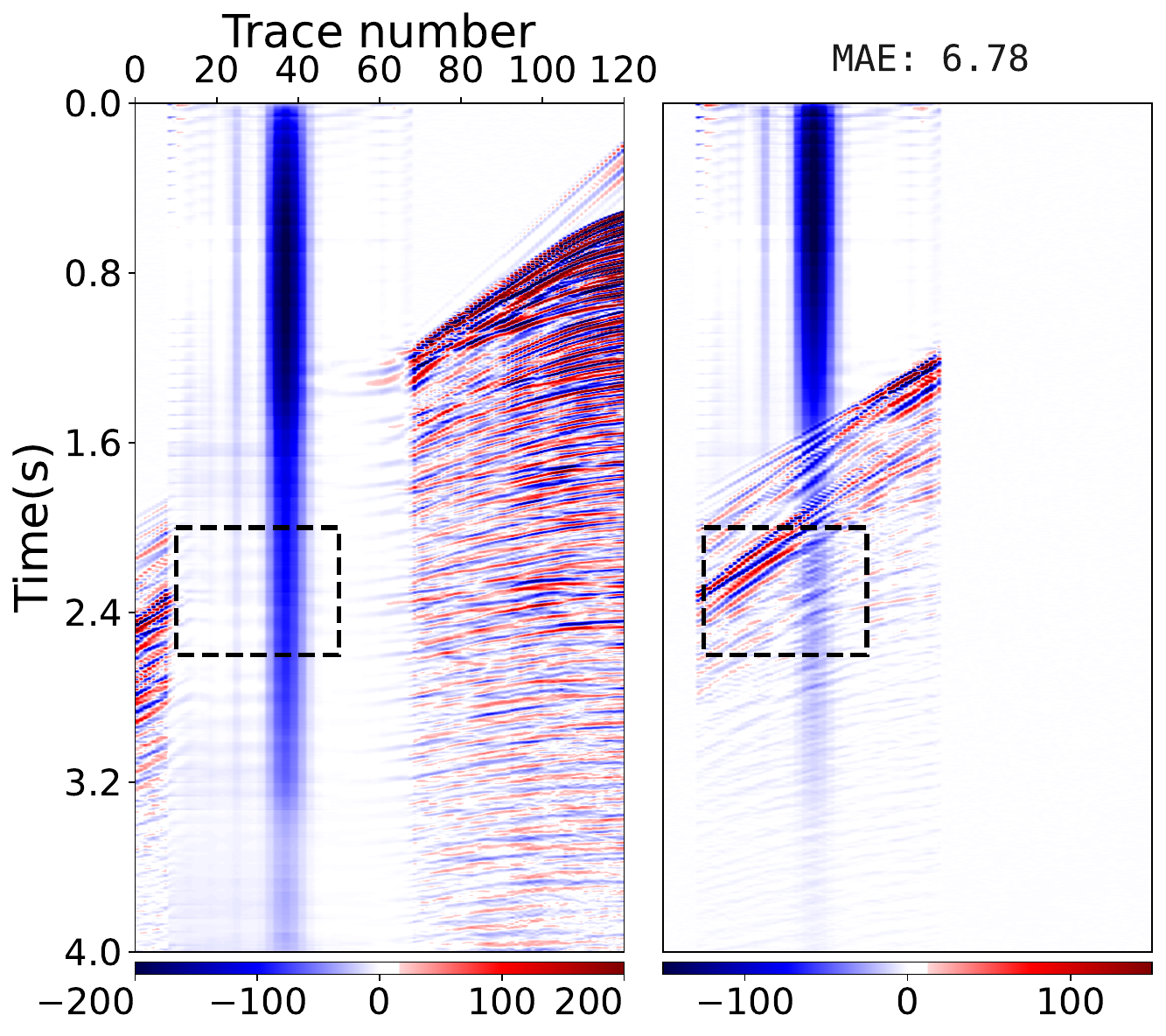}\label{fig:mavo_488_D3UNet}}
	\hspace{0mm}
	\subfloat[Ours]{\includegraphics[height=1.5in]{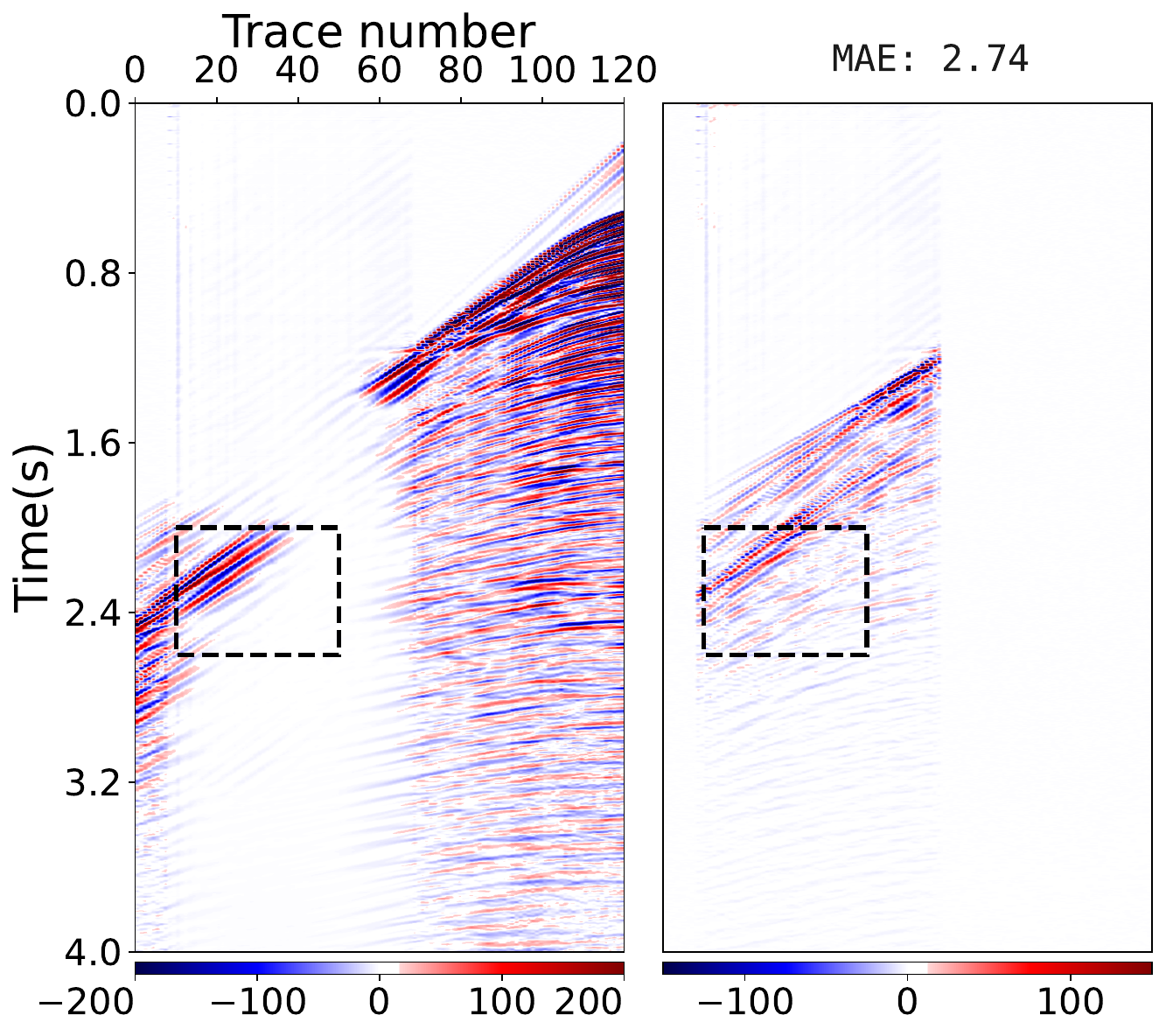}\label{fig:mavo_488_Ours}}
	\par
	
	\caption{Qualitative comparison of reconstruction results on the MAVO dataset under 50\% trace removal. (a) Random undersampling (shot 180). (b) Consecutive large gaps (shot 488). We restore the seismic data to its original amplitude range.}
	\label{fig:comparison_488}
\end{figure*}

%
%

\begin{figure}[t]
	\centering
	\includegraphics[width=\columnwidth]{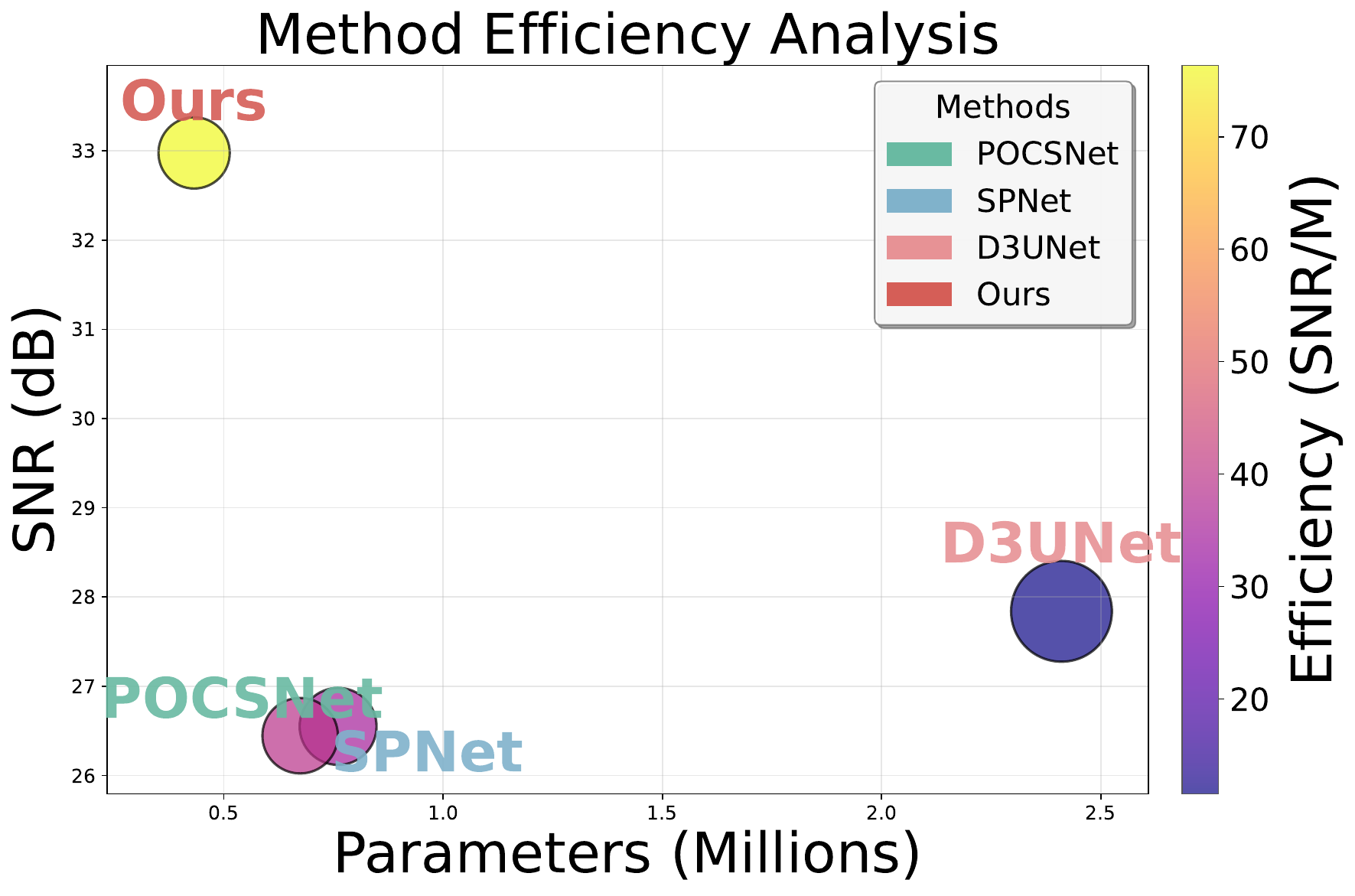}
	
	\caption{Pareto efficiency analysis (PSNR vs. Parameters). The proposed method optimally occupies the top-left Pareto frontier, achieving the highest reconstruction fidelity with minimal parameter overhead.}
	\label{fig:psnr_parameter_efficiency}
\end{figure}

\begin{table*}[t]
\centering\small
\caption{Reconstruction comparison at 50\% trace removal. Each entry is missing-region SNR (dB) / MAE; larger SNR and smaller MAE are preferable.}\label{tab:main}
\begin{tabular}{lcccc}
\toprule
Method & SEG C3 random & SEG C3 consecutive & MAVO random & MAVO consecutive\\
\midrule
Curvelet & 20.131/2.366e-02 & 1.0578/2.220e-01 & 24.600/1.292e-02 & 1.0598/2.333e-01\\
POCS-Net & 22.486/1.664e-02  & 17.810 /3.174e-02 & 26.550/9.237e-03 & 23.178/1.660e-02\\
SP-Net & 22.281/1.236e-02 & 15.578/3.875e-02 & 26.446/9.444e-03 & 19.110 /2.298e-02\\
D3UNet & 24.304/1.366e-02 & 18.527/2.825e-02 & 27.840/8.138e-03 & 24.241/1.317e-02\\
MetaPOCS & \textbf{28.302/8.600e-03} & \textbf{20.114/2.470e-02} & \textbf{32.977/5.657e-03} & \textbf{26.533/1.336e-02}\\
\bottomrule
\end{tabular}
\end{table*}


The waveform comparison in Fig.~\ref{fig:evo_meta_ori} complements the aggregate scores. Event continuity, arrival-time alignment, and signed amplitude are evaluated on fixed windows selected independently of method performance. Signed residuals distinguish coherent reconstruction errors from diffuse artifacts, while the $f$--$k$ representation characterizes spectral leakage and event slopes. These views also distinguish visually smooth interpolation from recovery of the measured wavefield structure.

\subsection{One Checkpoint Across Computational Budgets}
Table~\ref{tab:budget} evaluates the same checkpoint at depths 2, 4, 6, and 8. Independently trained fixed-depth models define budget-specialized references. Truncation of the deep-only model measures the quality of intermediate outputs without direct supervision. Equal and meta-trained exits share the same inference architecture, so differences between them reflect training rather than additional deployment modules.

The latency--quality frontier consists of operating points for which no alternative is both faster and more accurate. Equation~\eqref{eq:budget} selects among these profiled outputs for a specified budget. The comparison considers both shallow-exit performance and the deepest-exit change; a gain at low cost can coexist with a loss at full depth. Training time and checkpoint storage separately quantify the cost of learning one multi-exit model or retaining a collection of specialized models.


\begin{table*}[t]
	\centering\small
	\caption{Reconstruction quality and computational cost across exit depths. $Q_d$ denotes missing-region SNR (dB). Memory and latency are measured on the same device and input geometry. Inference latency is reported as the minimum and maximum time corresponding to $T=1$ and $T=7$, respectively. Training time includes meta optimization.}\label{tab:budget}
	\begin{tabular}{lcccccccc}
		\toprule
		\multirow{2}{*}{Method} & \multicolumn{4}{c}{Reconstruction Quality (dB)} & \multicolumn{3}{c}{Memory (MB)} & \multirow{2}{*}{Inf. Time (ms)} \\
		\cmidrule(lr){2-5} \cmidrule(lr){6-8}
		& \(Q_1\) & \(Q_3\) & \(Q_5\) & \(Q_7\) & Allocated & Peak & Checkpoint & (\(T_1\) / \(T_7\)) \\
		\midrule
		D3UNet & 3.008 & 10.226& 20.672 & 26.531 & 37.81 & 6630 & 9.27  & 3.783 \\
		Fixed-\(T\) & \textbf{26.370} & 30.206 & \underline{32.323} & \underline{32.767} & 81.3 & 19124 & 5.90 & 1.567 / 8.033 \\
		Deep-only trunc. & 14.275 & 19.420 & 23.066 & 32.202 & 81.3 & 19124 & 5.90 & 1.567 / 8.033 \\
		Equal-exit & 24.937 & 30.765 & 31.630 & 31.919 & 72.60 & 17908 & 5.91 & 1.617 / 8.033 \\
		Exit-weight BLO & 24.977 & \underline{31.073} & 31.803 & 31.936 & 73.58 & 23676 & 5.91 & 1.617 / 8.033 \\
		MetaPOCS & \underline{26.034} & \textbf{31.968} & \textbf{32.818} & \textbf{32.977} & 68.45 & 17922 & 2.08 & 1.450 / 8.033 \\
		\bottomrule
	\end{tabular}
\end{table*}

\section{Optimization Analysis and Ablation Studies}\label{sec:mechanism}
\subsection{Gradient Interference and Finite-Update Effects}
On the same minibatch and at the same parameter state, let $g_k=\nabla_{\theta_S}\mathcal L_k$, where $\theta_S$ is the common parameter support of the compared exits. For nonzero gradients,
\begin{equation}
c_{k\ell}=\frac{g_k^\top g_\ell}{\|g_k\|_2\|g_\ell\|_2},\quad
C=\operatorname{mean}_{k<\ell}\mathbf1[c_{k\ell}<0]. \label{eq:cos}
\end{equation}
The common support excludes parameters absent from either exit's computation graph. Layerwise values localize interference within the shared prefix. The gradient norm ratio
\begin{equation}
R_g=\frac{\max_k\|g_k\|_2}{\min_k\|g_k\|_2+\epsilon}
\end{equation}
characterizes scale imbalance independently of directional conflict. Gradients below a fixed numerical tolerance are omitted from cosine averages and counted separately, since their directions are poorly defined.

For $\theta'=\theta-\eta d$, the first-order loss change is $-\eta g_k^\top d$. A negative $g_k^\top d$ therefore predicts a locally harmful update. Positive scalar reweighting preserves the pairwise cosine at a fixed state: $\cos(a g_k,b g_\ell)=\cos(g_k,g_\ell)$ for $a,b>0$. It changes the contribution of each gradient to the aggregate direction, whereas tensor-dependent weighting also changes the relative allocation across parameter groups. Training can subsequently move the model to states with different raw gradients. The analysis separates these immediate and trajectory-dependent effects.

Full fusion rounds operate on multi-step displacements. Their first-order diagnostic uses $d=-\Delta_{\rm fused}$, while the finite-update effect on a fixed probe batch is
\begin{equation}
D_k=\mathcal L_k(\theta+\Delta_{\rm fused};\mathcal D_{\rm probe})
-\mathcal L_k(\theta;\mathcal D_{\rm probe}). \label{eq:probe}
\end{equation}
The intervention applies learned, uniform, frozen, and shuffled weights to the same base parameters and cached candidate displacements. Shuffling exchanges weights only among supported exits of a tensor, preserving the support constraint. Stochastic layers and normalization statistics remain fixed during the probe. Consequently, changes in $D_k$ arise from the fusion rule rather than different inner trajectories. The actual-update comparison is complemented by a common-norm comparison to distinguish direction from step magnitude.


Table~\ref{tab:dataset_params_en} relates optimization diagnostics to exit quality. The learned-minus-uniform change in harmful-update frequency is $\pending$, and the paired shallow/deep reconstruction changes are $\pending/\pending$~dB. The uniform-fusion comparison holds the expensive inner trajectories fixed, isolating whether the outer objective identifies a more favorable mixture. Gradient-scale balance, finite-update harm, and reconstruction quality remain separate response variables: an improvement in one does not imply an improvement in the others.

Learning curves use both task-update count and elapsed GPU time as their horizontal axes. Time to a fixed validation-quality threshold characterizes convergence efficiency; the proportion of runs reaching that threshold characterizes reliability. Similar training error with lower held-out error is instead consistent with a generalization effect. These distinctions connect the hypergradient in Eq.~\eqref{eq:hypergradient} to observable training behavior without requiring every pair of raw gradients to become aligned.


\subsection{Interaction With Reconstruction Priors and Losses}
The factorial comparison in Table~\ref{tab:interaction} separates the training rule from the reconstruction prior. Each backbone/loss configuration appears under both equal supervision and meta fusion. For quality $Q$, the interaction
\begin{equation}
I=[Q_{\rm meta,dual}-Q_{\rm equal,dual}]
-[Q_{\rm meta,spatial}-Q_{\rm equal,spatial}]
\end{equation}
measures how the meta-learning effect changes between a dual-domain and spatial-only prior. The paired confidence interval for $I$ preserves the matched acquisition units and seeds of the four cells. Concatenation and DFM distinguish feature exchange from update fusion. Crossing spatial-only and dual-domain priors with amplitude-only and amplitude--frequency supervision isolates the roles of implicit representation and explicit reconstruction constraints; transform regularization remains fixed. The absolute scores identify the effect of adding FFL to each prior, and the paired meta-minus-equal differences identify its interaction with the training rule. A plain POCS-Net backbone supplies an additional transfer condition. The effect of a prior alone and its interaction with meta training are consequently estimated separately.

\begin{table*}[t]
\centering\small
\caption{Interaction between implicit reconstruction priors, explicit reconstruction objectives, and training. Equal and meta columns report $Q_2/Q_8$ in dB; the final column gives their paired difference. Transform regularization is retained in every matched configuration.}\label{tab:interaction}
\begin{tabular}{llcccc}
\toprule
Prior & Loss recipe & MAE$\downarrow$ & SSIM$\uparrow$ & SNR$\uparrow$ &Improvement \\
\midrule
Spa--Spa, concat & $L_1$ & 8.063e-03 & 0.941 & 28.587 &\ding{51}\\
FFC--FFC, concat & $L_1$ & 8.267e-03 & 0.943 & 28.588 &\ding{51}\\
Spa--FFC, concat & $L_1$ & 8.239e-03 & 0.946 & 29.549 &\ding{51}\\
Spa--FFC, DFM & $L_1$ & 6.862e-03 & 0.956 & 30.831 &\ding{51}\\
Spa--FFC, DFM & $L_1+\mathrm{FFL}$ & 6.904e-03 & 0.957 & 31.049 &\ding{51}\\
Spa--FFC, DFM & $L_1+\mathrm{sym}+\mathrm{FFL}$ & \textbf{6.427e-03} & \textbf{0.959} & \textbf{31.463} &\ding{51}\\
\bottomrule
\end{tabular}
\end{table*}

\subsection{Meta-Optimization Sensitivity}
The sensitivity analysis varies one factor around the reference configuration while retaining the same data partition and reconstruction objective. The auxiliary coefficient $\beta$ controls the distinction among exit-oriented tasks; $\mu$ controls the temporal smoothing of fusion logits. The refresh period changes both optimization dynamics and training cost. Fusion granularity and support are compared using common candidate trajectories, while meta-data-source comparisons retain a separate monitoring role. Scalar meta-supervision has its own floor parameter $\rho$ and is reported as a distinct training family.

\begin{table}
	\centering
	\vspace{-1mm}
	\caption{Generalization comparison of different models on the test set of the Model94 dataset (missing rate: 50\%).}
	\label{tab:general_different_dataset}
	\resizebox{1\linewidth}{!}{
		\begin{tabular}{llcccc}
			\toprule
			Type & Model & MAE$\downarrow$ & SNR$\uparrow$ & SSIM$\uparrow$ & PSNR$\uparrow$ \\
			\midrule
			\multirow{3}{*}{\begin{tabular}{@{}l@{}}SEGC3 to Model94: \\Random missing\end{tabular}}
			& D3UNet & 0.0105 & 25.144 & 0.9501 & 31.712 \\
			& Ours (T=7) & \textbf{0.0081} & 26.788 & 0.9645 & 33.499 \\
			& Ours (Best T=6) & 0.0083 & \textbf{26.875} & \textbf{0.9645} & \textbf{33.586} \\
			\midrule
			\multirow{3}{*}{\begin{tabular}{@{}l@{}}MAVG to Model94: \\Random missing\end{tabular}}
			& D3UNet & 0.0231 & 17.934 & 0.8136 & 24.502 \\
			& Ours (T=7)  & 0.0149 & 21.057 & 0.9073 & 27.768 \\
			& Ours (Best T=6)  & \textbf{0.0141} & \textbf{21.227} & \textbf{0.9073} & \textbf{27.938} \\
			\midrule
			\multirow{3}{*}{\begin{tabular}{@{}l@{}}MAVG to Model94: \\Consecutive missing\end{tabular}}
			& D3UNet  & 0.0593 & 13.331 & 0.7845 & 19.899 \\
			& Ours (T=7) & 0.0492 & 14.717 & 0.7936 & 21.428 \\
			& Ours (Best T=2) & \textbf{0.0480} & \textbf{14.891} & \textbf{0.7936} & \textbf{21.602} \\
			\midrule
			\multirow{3}{*}{\begin{tabular}{@{}l@{}}SEGC3 to Model94: \\Consecutive missing\end{tabular}}
			& D3UNet  & 0.0766 & 11.269 & 0.7638 & 17.837 \\
			& Ours (T=7) & 0.1306 & 6.545 & 0.7621 & 13.257 \\
			& Ours (Best T=2) & \textbf{0.0715} & \textbf{12.027} & \textbf{0.7621} & \textbf{18.739} \\
			\bottomrule
		\end{tabular}
	}
\end{table}


\subsection{Acquisition Shift and Failure Cases}
Table~\ref{tab:dataset_params_en} separates in-distribution interpolation from acquisition and geological transfer. In mask-transfer experiments, a model trained with random removal is evaluated on consecutive or periodic removal without fine-tuning. Density transfer changes the missing fraction while retaining the same checkpoint. Wavelet and velocity-model transfer use independent synthetic conditions, and survey transfer changes the acquisition source. These tests concern missing traces on the documented grid; off-grid coordinate regularization is a different sampling problem.

Failure analysis focuses on gap centers, weak reflections, steep dips, and crossing events. At a fixed missing fraction, gap width changes the distance between an unobserved event and its available boundary information. Weak-event windows further separate amplitude recovery from scores dominated by strong reflectors. In noisy-input experiments, the hard consistency operator retains the actual noisy measurements; the resulting acquired-sample error is therefore evaluated independently from interpolation in missing traces.


\section{Downstream Reverse-Time-Migration Validation}\label{sec:rtm}

\subsection{Controlled Imaging Experiment}
RTM provides a downstream test of whether interpolation preserves wavefield information relevant to subsurface imaging. The controlled setting uses two-dimensional acoustic data with known source and receiver coordinates. In a constant-density medium, the source wavefield satisfies
\begin{equation}
\frac{1}{v^2(\mathbf r)}\frac{\partial^2u_s}{\partial t^2}
-\nabla^2u_s=q_s(\mathbf r,t), \label{eq:wave}
\end{equation}
where $\mathbf r$ is subsurface position, $v$ is velocity, and $q_s$ is the source term. The numerical formulation follows finite-difference wave propagation and reverse-time imaging, for which Devito provides a reproducible implementation framework~\cite{Louboutin2019Devito}.

Layered, faulted, and laterally heterogeneous models define distinct structural conditions. Entire velocity models form the independent test units. Complete gathers are generated with the true model, whereas migration uses a smooth background velocity $v_b$ shared by all interpolation methods. Grid spacing, source wavelet, boundary treatment, and temporal sampling determine the modeled bandwidth and numerical dispersion. The imaging comparison fixes these quantities, with discretization and background-velocity sensitivity treated as separate conditions.

The input groups in Table~\ref{tab:dataset_params_en} are complete data, zero-filled incomplete data, conventional interpolation, SSI-POCS, equal multi-exit reconstruction, and MetaPOCS. Multi-exit inputs cover depths 2, 4, 6, and 8. Direct-arrival treatment, bandpass, migration aperture, and illumination compensation are common to all groups. Interpolated data enter migration after reversal of the reconstruction normalization, preserving their relative amplitudes.

Let $\mathcal M_{v_b}$ be the fixed migration operator using background velocity $v_b$:
\begin{equation}
I_{\rm ref}=\mathcal M_{v_b}(d_{\rm full}),\qquad
I_a=\mathcal M_{v_b}(\hat d_a). \label{eq:rtm}
\end{equation}
For example, a correlation imaging condition has the form
\begin{equation}
I_a(\mathbf r)=\sum_s\int_0^{t_{\max}}
u_s(\mathbf r,t;v_b)\,b_{s,a}(\mathbf r,t;v_b)\,dt,
\end{equation}
where $b_{s,a}$ is the receiver wavefield back-propagated from the reconstructed data. With a fixed source wavefield and linear receiver propagation, the image depends linearly on its data input. Accordingly, fixed linear preprocessing and migration give
\begin{equation}
I_a-I_{\rm ref}=\mathcal M_{v_b}(\hat d_a-d_{\rm full}).
\end{equation}
The complete-data image is the band-limited reference associated with the same imaging operator. Data-space error and image-space error weight interpolation residuals differently: two residual fields with similar energy can produce different coherent artifacts after migration. Therefore, the downstream analysis evaluates imaging error directly rather than treating data SNR as a surrogate.

\subsection{Image-Domain Accuracy}
The relative image error over a fixed subsurface window outside the absorbing boundaries is
\begin{equation}
E_I=\|I_a-I_{\rm ref}\|_F/\|I_{\rm ref}\|_F.
\end{equation}
This measure is complemented by image correlation, reflector-position error in meters, and artifact energy within prescribed nonreflector regions. The synthetic model supplies structural interface locations, while $I_{\rm ref}$ supplies the corresponding band-limited image amplitudes. Paired uncertainty is computed across independent velocity models, retaining the shots and masks of each model within the same resampled unit.

At each reconstruction budget, Table~\ref{tab:dataset_params_en} relates missing-trace SNR to $E_I$, reflector localization, and artifact energy. The paired MetaPOCS-minus-equal change in image error is $\pending$, and the corresponding reflector-position change is $\pending$~m. Figure~\ref{fig:evo_meta_ori} provides the associated image residuals and fixed reflector profiles. Agreement between data- and image-domain changes supports downstream utility within this acoustic setting; disagreement identifies interpolation errors to which the imaging operator is particularly sensitive.

\FloatBarrier
\section{Discussion and Limitations}
MetaPOCS separates reconstruction representation, training allocation, and deployment cost. The spatial--spectral prior supplies the representation inherited from SSI-POCS. Supervised exits make several intermediate wavefields explicit optimization targets. Meta-GF-based fusion then allocates exit-oriented updates across the shared parameter tensors, while the latency policy selects a deployment budget. This separation explains why a feature-fusion improvement, an optimization improvement, and a computational saving require different comparisons.

The fusion derivative also clarifies the role of the outer objective. A candidate update is assessed relative to the current mixture, using the sensitivity of the equal-exit reconstruction loss. This permits tensor-dependent preferences without requiring a globally preferred exit. Nevertheless, local hypergradient information does not imply global convergence or simultaneous improvement of every exit. The probe analysis quantifies finite-step effects, and the full learning curves establish whether those effects persist over optimization.

Several limitations remain intrinsic to the formulation. Full fusion rounds require additional inner trajectories, which may offset the savings obtained during deployment when few gathers are processed. Budget selection is device and input dependent, and it does not enforce monotonic reconstruction quality with depth. Hard reinsertion preserves measured amplitudes but also retains measurement noise; wide gaps remain underdetermined by acquisition constraints alone. Finally, a controlled acoustic imaging experiment characterizes one downstream operator and does not establish elastic or AVO fidelity. These limitations motivate more economical fusion schedules, uncertainty-aware budget selection, and extension to richer acquisition geometries.

\section{Conclusion}
We have presented MetaPOCS, a multi-exit unfolding formulation for seismic interpolation under varying computational budgets. The framework combines the SSI-POCS reconstruction prior with supervised intermediate outputs and Meta-GF-based fusion of exit-oriented parameter displacements. Exact acquisition consistency is maintained at every exit, and a calibrated prefix policy supports several deployment operating points from one checkpoint. The derived fusion hypergradient explains how tensor-wise weights respond to candidate updates relative to their current mixture. The evaluation formulation connects this training mechanism to matched-budget waveform reconstruction and fixed-operator migration error. Together, these components establish a coherent basis for assessing the accuracy, optimization cost, and downstream value of budget-aware seismic unfolding.

\section*{Data and Code Availability}
The implementation, configurations, acquisition partitions, and evaluation scripts are associated with the DAP project. The public repository or archival identifier is $\pending$. Dataset access identifiers and licenses are $\pending$; checkpoint identifiers and the reproducible environment specification are $\pending$.

\section*{Acknowledgment}
The authors acknowledge the seismic data providers. Dataset-specific attribution will be finalized after source and license verification. OpenAI Codex assisted with manuscript restructuring, language drafting, and experimental-plan preparation; the authors are responsible for verifying scientific content, references, and final results.
\bibliographystyle{IEEEtran}
\bibliography{main}
\end{document}